\documentclass{article}
\usepackage[T1]{fontenc}
\usepackage{iclr27/iclr2027_conference,times}

\usepackage{amsmath,amsfonts,bm}

\def\eqref#1{equation~\ref{#1}}

\def\1{\bm{1}}

\DeclareMathAlphabet{\mathsfit}{\encodingdefault}{\sfdefault}{m}{sl}
\SetMathAlphabet{\mathsfit}{bold}{\encodingdefault}{\sfdefault}{bx}{n}

\usepackage{hyperref}
\usepackage{url}

\usepackage{fix-cm}
\usepackage{amssymb}
\usepackage{mathrsfs}
\usepackage{amsthm}
\usepackage{enumitem}%
\usepackage{graphicx}
\usepackage{wrapfig}%
\usepackage{lipsum}%
\usepackage{float}
\usepackage{placeins}
\usepackage{booktabs}
\usepackage{amsmath}
\usepackage{amsfonts}
\usepackage{nicefrac}
\usepackage{microtype}
\usepackage{xspace}
\usepackage{multirow}
\usepackage{subcaption}%
\usepackage{cleveref}
\crefname{section}{Section}{Sections}
\Crefname{section}{Section}{Sections}
\crefname{subsection}{Section}{Sections}
\Crefname{subsection}{Section}{Sections}
\crefname{subsubsection}{Section}{Sections}
\Crefname{subsubsection}{Section}{Sections}
\crefname{figure}{Figure}{Figures}
\Crefname{figure}{Figure}{Figures}
\crefname{table}{Table}{Tables}
\Crefname{table}{Table}{Tables}
\crefname{equation}{Eq.}{Eqs.}
\Crefname{equation}{Eq.}{Eqs.}
\creflabelformat{equation}{#2#1#3}
\usepackage{bm}
\usepackage{times}
\usepackage{epsfig}
\usepackage{adjustbox}
\usepackage{fontawesome}
\usepackage{xcolor,colortbl}
\usepackage{cuted}
\usepackage{fancyvrb}%
\usepackage[most]{tcolorbox}%

\definecolor{agdesign}{HTML}{7E6BB5}
\definecolor{agcode}{HTML}{3E7CB1}
\definecolor{agreview}{HTML}{4F8A63}
\definecolor{agsim}{HTML}{6E7B87}

\definecolor{agbad}{HTML}{B4483C}
\definecolor{agwarn}{HTML}{C08A2E}
\definecolor{aggood}{HTML}{3F7F53}

\newcommand{\traceround}[1]{\par\vspace{5pt}%
  {\scriptsize\bfseries\color{black!55}#1}\hspace{5pt}%
  {\color{black!25}\leaders\hrule height 0.5pt\hfill\kern0pt}\par\vspace{1pt}}

\newcommand{\tchip}[2]{{\setlength{\fboxsep}{1.4pt}\colorbox{#1}{\color{white}\bfseries\tiny #2}}}

\newtcolorbox{tracestep}[4]{%
  enhanced, breakable, sharp corners,
  colback=#1!6!white, colframe=#1!35!white, boxrule=0.3pt,
  borderline west={3.2pt}{0pt}{#1},
  left=5pt, right=5pt, top=3pt, bottom=3.5pt,
  toptitle=2pt, bottomtitle=2pt, lefttitle=5pt, righttitle=5pt,
  colbacktitle=#1, coltitle=white,
  fonttitle=\bfseries\scriptsize,
  title={#2\hspace{0.9em}\normalfont#3\hfill#4},
  before upper={\def\tracelane{#1}},
  after skip=2.5pt, before skip=2.5pt,
}
\providecommand{\tracelane}{black!50}

\newcommand{\methodname}{Artisan\xspace}
\newcommand{\ours}{\textsc{\methodname}\xspace}

\newcommand{\heading}[1]{\vspace{2pt}\noindent\textbf{#1.}}

\newcommand{\best}[1]{\textbf{#1}}
\newcommand\rurl[1]{%
  \href{https://#1}{\nolinkurl{#1}}%
}

\usepackage{silence}
\newif\ifshortdraft
\shortdraftfalse

\title{Towards Physically Grounded Articulated 3D Assets: A Multi-Agent Pipeline and Benchmark}

\iclrfinalcopy
\usepackage{etoolbox}
\makeatletter
\patchcmd{\@maketitle}
  {\begin{tabular}[t]{l}\bf\rule{\z@}{24pt}\@author\end{tabular}}
  {\makebox[\textwidth][c]{\begin{tabular}[t]{c}\bf\rule{\z@}{24pt}\@author\end{tabular}}}
  {}{\PackageError{author-layout}{Could not center author block}{}}
\makeatother
\author{%
Xudong Cai$^{1\#*}$ \quad
Ruijie Zhu$^{1*}$ \quad
Haoliang Shang$^{1,2*}$ \quad
Ruining Li$^{3*}$ \quad \\[0.2em]
\textbf{Marc Pollefeys}$^{2,4}$ \quad
\textbf{Shangzhe Wu}$^{5}$ \quad
\textbf{Andrea Vedaldi}$^{3}$\quad
\textbf{Chuanxia Zheng}$^{1}$\\[0.5em]
\normalfont $^{1}$PVG, Nanyang Technological University \quad $^{2}$ETH Zurich\\
\normalfont $^{3}$VGG, University of Oxford \quad $^{4}$Microsoft \quad $^{5}$University of Cambridge
\\[0.3em]
\small\rurl{physicalvision.github.io/artisan}
}
\begin{document}
\maketitle

\begingroup
\renewcommand{\thefootnote}{}
\footnotetext{$\#$ Project lead. \quad $*$ Core technical contribution.}
\endgroup

\begin{figure}[h]
\centering
\vspace{-2.5em}
\includegraphics[width=\linewidth]{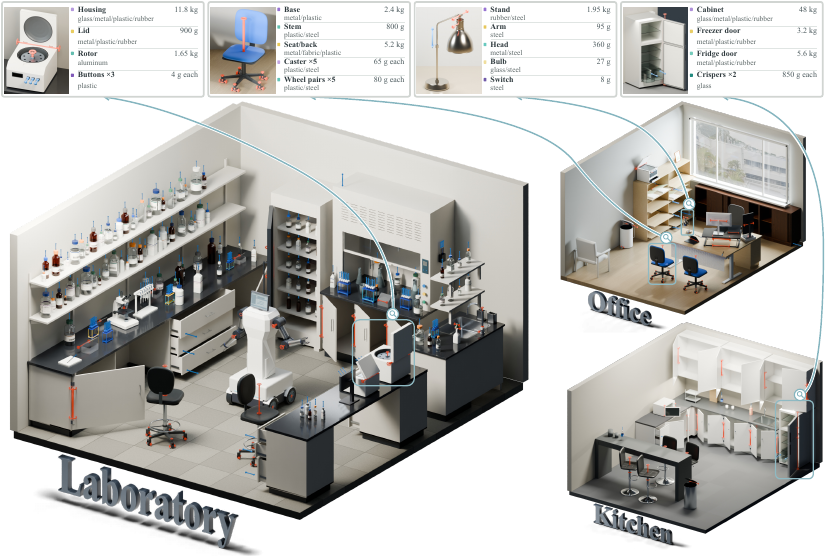}
\vspace{-2em}
\caption{
\textbf{Physically grounded articulated 3D objects generated by \ours} (assembled into scenes).
\textcolor{orange}{Orange} and \textcolor{blue}{blue} annotations indicate revolute and prismatic axes, respectively.
Insets list the physical attributes composition of individual parts for exampled objects. 
\textcolor{red}{Zoom in for more details}.
}
\vspace{-0.5em}
\label{fig:teaser}
\end{figure}

\begin{abstract}

In this paper,
we study \emph{physically grounded articulated 3D asset generation}---moving the goal from reproducing how objects look to capturing how they move and function.
Inspired by the literature on procedural and agentic generation,
we present \ours,
a system that extends the latest method Articraft by introducing a \emph{multi-agent} pipeline with \emph{physical simulation feedback},
enabling comprehensive attributes beyond geometry, appearance, and physically plausible articulation,
such as collision-free motion, stability under gravity, and accurate mass and material properties.
Furthermore,
we construct a unified benchmark spanning four key aspects:
spatial fidelity,
physical plausibility,
downstream task performance,
and overall quality.
The benchmark encompasses 100 reference assets from ArtVIP and LightWheel,
50 stability prompts,
and 50 ABO objects with reference masses,
with protocols evaluating various metrics.
Through extensive evaluation of more than 10 representative models,
\ours achieves competitive spatial fidelity while attaining the lowest collision rates and highest stability rates among the evaluated methods.
Together, our system and benchmark establish insights for advancing 3D generation towards physically plausible, functionally usable assets.

\end{abstract}

\section{Introduction}%
\label{sec:intro}

We consider the problem of generating physically grounded articulated 3D assets from natural language and, optionally, reference images.
Specifically, we aim to reconstruct the objects' appearance and geometry, as well as structural, functional, and physical properties such as part hierarchies, joints, and mass.
The intent is to capture the configuration space of the object and to support simulation of contact and gravity.
By solving this problem, we can reduce the cost of creating assets for physics simulation~\citep{Genesis,genesis2026genesisworld}, product design~\citep{willis2021fusion}, and robotics~\citep{li2023behavior,nasiriany26robocasa365:}.

Appearance, shape, articulation, and physical attributes are interdependent and should be modeled \emph{holistically}.
However, most existing methods address these properties separately, focusing on shape and appearance~\citep{xiang25structured,hunyuan3d25hunyuan3d,wu2025amodal3r}, articulation~\citep{wu2026urdfanything,li2025particulate,li2026instruct}, or specific physical properties, such as materials~\citep{wu2025pbr,youwang2026extracting}, mass~\citep{zhai2024nerf2physics,cao2025physxanything,cao2026physxomni,li2026uniphysgen}, stability~\citep{chen2024atlas3d,li2025dso}, and collision-free motion~\citep{luo2025physpart,kreber2025physnap}.
Addressing each aspect in isolation makes it difficult to satisfy all of them, resulting in artifacts and defects in the reconstructions 
and ultimately limiting the usability of the generated assets.

As alternatives to these learned models, procedural
generators~\citep{lian2025infinite,joshi2025infinigenarticulated} assemble assets programmatically, based on predefined rules that enforce desirable constraints and make at least some physical properties explicit and controllable.
However, such rule-based systems limit the types of objects that can be generated and are difficult to use in tasks like image-based reconstruction.
This limitation is addressed by several recently proposed agentic methods.
For instance, Articraft~\citep{zhou2026articraft} uses a Large Language Model (LLM) to write, validate, and revise programs that construct 3D objects, which is much more flexible than using canned rules of traditional procedural generators.
However, Articraft and other similar approaches use a single agent to both construct and verify the `3D program', which can lead to \emph{self-evaluation bias}, where the agent may satisfy its own checks without genuinely ensuring the quality of the generated assets.
Moreover, while their checks are meant to increase the likelihood that the generated assets would work in a physics-simulated environment, the latter is not directly tested and may still fail.

In this paper, we address these gaps by introducing \ours, a new pipeline that improves on the blueprint of Articraft~\citep{zhou2026articraft} by switching to a \emph{multi-agent} design and introducing \emph{physical simulator feedback}.
The new pipeline (\cref{fig:pipeline}) consists of:
\textbf{(i)} the \emph{Design Agent}, which produces a blueprint of the object specifying its parts, joints, and articulations;
\textbf{(ii)} the \emph{Coding Agent}, which translates the blueprint into a program that outputs the geometry, articulations, and physical parameters of the object; and
\textbf{(iii)} the \emph{Review Agent}, which validates the built asset and provides feedback to the Coding Agent to revise the program.
The agents work \emph{independently}, exchanging only the blueprint, 3D asset, and feedback as explained.
Their independence reduces the self-evaluation bias mentioned above.
Furthermore, the validation step is augmented with feedback obtained from a physical simulator, which verifies physical properties like collisions and stability within the target simulator environment.

Existing benchmarks offer limited assessment of whether visual fidelity translates into physical plausibility
and even functional usability to evaluate the assets produced by such pipelines. 
Our second key contribution is therefore to introduce a new comprehensive benchmark that jointly assesses all of these properties.
It evaluates:
\emph{spatial fidelity}, comparing the object's appearance, geometry, and kinematics to the reference inputs;
\emph{physical attributes}, measuring mass, material, stability under gravity, and collision under motion;
\emph{downstream tasks}, assessing the usability of the generated assets in tasks like robotic manipulation; and
\emph{overall quality}, based on human judgments after manipulating objects in the simulator.
Each aspect is evaluated with multiple metrics.

Using the new benchmark, we conduct extensive experiments to compare our method with state-of-the-art models.
Empirically, we find that the assets constructed by \ours{}:
\textbf{(i)} are faithful to the input references;
\textbf{(ii)} are physically valid and articulate correctly; and
\textbf{(iii)} work well in downstream tasks such as simulated robotic manipulation.
We also show that \ours~outperforms the recent and strong models in most metrics and in human evaluation,
especially improving physical properties like collision and stability.
In most cases, our general-purpose pipeline is more accurate than task-specific feed-forward 3D reconstruction models.
We also show that the constructed objects are functional and support downstream robotic manipulation tasks.

\section{Related Work}%
\label{sec:related}

\heading{Creating Functional 3D Assets}
Early methods build articulated objects through per-instance optimization~\citep{mu2021sdf,wu2022d,tseng2022cla,liu2023paris,weng2024neural,wu2025reartgs},
but are limited to specific instances.
To improve generalizability,
learning-based methods create articulated objects from
pure noise~\citep{lei2023nap,gao2025meshart},
text~\citep{liu2024cage,Su_2025_CVPR},
images~\citep{liusingapo,jiang2022ditto,chen2024urdformer,he2025spark,chen2025ArtiLatent,huang2026react3d,wu2026urdfanything},
videos~\citep{delitzas2026funrec}
or 3D models~\citep{li2025particulate,zhang2026simart,li2026instruct},
 learning from large-scale datasets~\citep{xiang2020sapien,li2026instruct}.
Recent methods focus on inferring independent physical properties from visual representations~\citep{zhai2024nerf2physics,shuai2025pugs}, jointly predicting these alongside joint motion characteristics~\citep{cao2025physxanything,cao2026physxomni,li2026uniphysgen,huang2026physxcot}, and leveraging simulation feedback to refine aspects such as support, clearances, or mobility~\citep{chen2024atlas3d,li2025dso,luo2025physpart,kreber2025physnap,xu2026motionanymesh}.
In contrast,
\ours{} generates articulated objects that take diverse physical properties into account.

\heading{Procedural and Agentic 3D Asset Creation}
Procedural systems construct editable and simulation-ready variations within the authored rules~\citep{lian2025infinite,joshi2025infinigenarticulated}.
The latest agent-based pipelines extend this coverage through executable articulation,
retrieval with visual critique,
or semantic search over kinematic structures~\citep{mandi2025realcode,le2024articulate,qiu2025articulate,wang2025kinematify}.
Closer to our approach,
Articraft~\citep{zhou2026articraft} builds articulated objects through an agent with Software Development Kit (SDK)s.
Recent systems improve reliability through geometric and visual checks, and revisions to earlier construction stages when checks fail~\citep{wang2026adsl,shui2026articad,lin2026procedura}.
In contrast,
\ours{} introduces a multi-agent pipeline with physical simulator feedback,
enabling physically grounded articulated 3D asset generation.

\heading{Datasets and Benchmarks for Physically Grounded 3D Assets}
Articulated datasets provide part geometry and motion annotations~\citep{xiang2020sapien,Geng_2023_CVPR,liu2022akb,delitzas2024scenefun3d,jin2026artvip,Engelbracht_2026_CVPR}.
Program-generated collections further expand the available assets~\citep{zhou2026articraft,cao2026physxomni,li2026uniphysgen}.
Existing benchmarks separately assess geometric and semantic fidelity~\citep{li2025particulate,duggal2025eval3d},
programmatic modelling quality~\citep{yang2026p3dbench,gao2026codebench},
material properties~\citep{cao2026physxomni,li2026uniphysgen},
and motion behavior~\citep{luo2025physpart,pham2026asset2sim}.
Our benchmark comprehensively evaluates the spatial fidelity, physical plausibility, and downstream task utility of generated articulated assets using a set of distinct metrics.

\section{Method: A Multi-Agentic Pipeline with Physical Validation}%
\label{sec:method}

Our goal is to construct a system $\mathcal{P}$ that, given as input a natural language prompt $\bm{p}$ and, optionally, one or more reference images $\mathcal{I}$, generates as output an articulated 3D asset $\mathcal{O}$:
\begin{equation}
\label{eq:goal}
\mathcal{P} :
(\bm{p}, \mathcal{I}) 
\mapsto \mathcal{O} 
= 
\{
\{\mathcal{G}_i, \rho_i, \mathcal{M}_i\}_{i=1}^P, (\mathcal{K}, \mathcal{J})
\}.
\end{equation}
The object $\mathcal{O}$ consists of $P$ rigid parts defined by their geometry $\mathcal{G}_i$, mass $\rho_i$, and material
$
\mathcal{M}_i
=
\{
\bm{c}_i, m_i, r_i, \alpha_i
\}
$
specified by its
base color $\bm{c}_i$,
metalness $m_i$,
roughness $r_i$
and opacity $\alpha_i$ (i.e., standard Physically Based Rendering (PBR) parameters).%
\footnote{The SDK supports distinct PBR materials with optional textures for components within each part.}
$\mathcal{K}$ specifies the \emph{kinematic structure}, defining the parent-child relationships among the $P$ parts, while $\mathcal{J}$ specifies the joint types, axes, motion limits, and any coupling constraints between dependent joints.
The pipeline aims to output an object $\mathcal{O}$ that faithfully reproduces the inputs $\bm{p}$ and $\mathcal{I}$ and is physically sound:
parts should not collide, freestanding objects should remain stable under gravity, and it must be possible to grasp and manipulate it.
The pipeline consists of three independent LLM agents equipped with tools and execution harnesses, $\mathcal{P}=\{\mathcal{A}_i\}_{i=1}^3$, described below, with implementation details given in \cref{sec:impl}.

\begin{figure}[tb!]
    \centering
    \includegraphics[width=\linewidth]{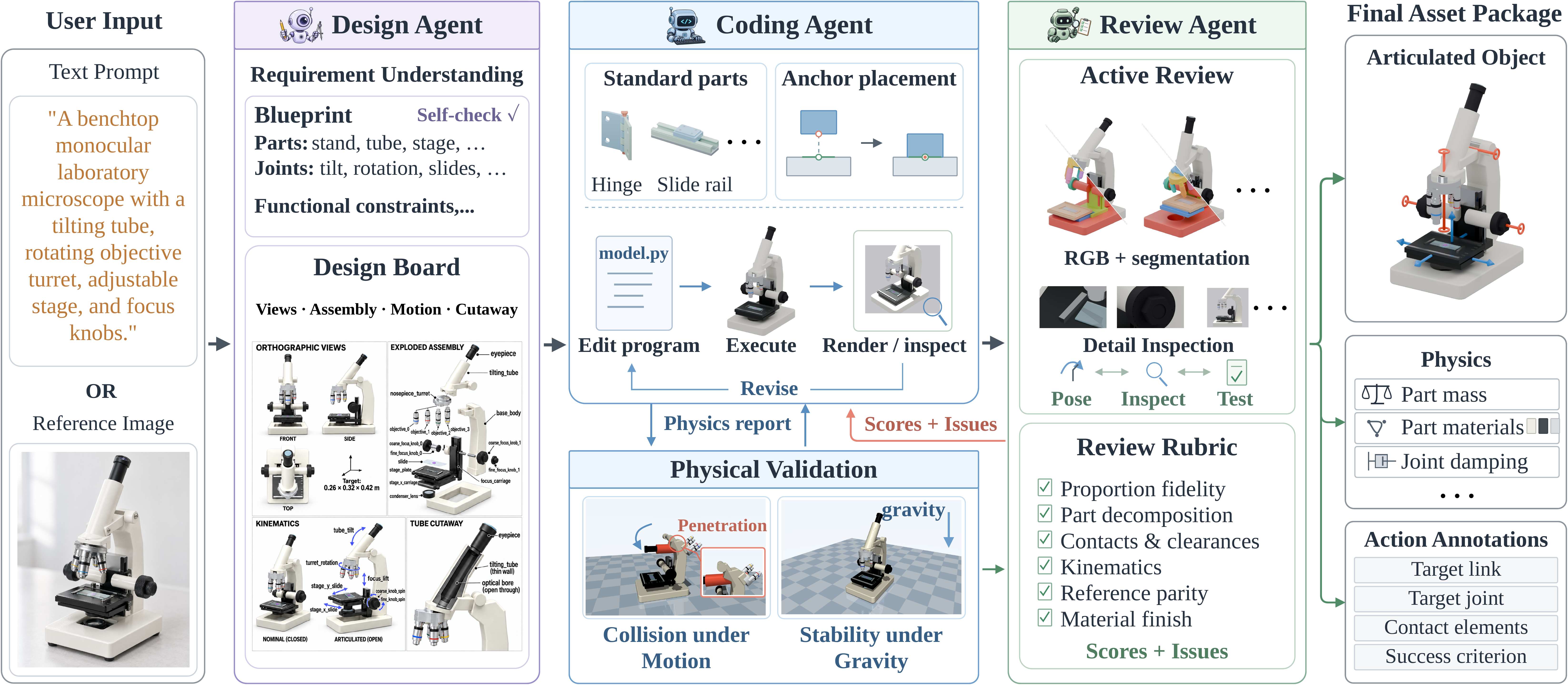}
    \caption{The staged pipeline of the three agents: Design Agent, Coding Agent, and Review Agent.
    }
\label{fig:pipeline}
\end{figure}

\subsection{Design Agent}%
\label{sec:design_agent}

The \emph{Design Agent}
$\mathcal{A}_{\mathrm{design}}$
takes the prompt and image as input, and outputs a structured textual blueprint $\mathcal{B}$ and a design board $\mathcal{D}$:
\begin{equation}
  \label{eq:design_flow}
  \begin{gathered}
  \mathcal{A}_{\mathrm{design}}(\bm{p}, \mathcal{I}) \rightarrow (\mathcal{B},\mathcal{D}),
  \text{where}
  ~\mathcal{B} = \bigl\{\{\tilde{\mathcal{G}}^{\text{text}}_i,\tilde{\mathcal{M}}^{\text{text}}_i\}_{i=1}^P,\ (\tilde{\mathcal{K}}^{\text{text}}, \tilde{\mathcal{J}}^{\text{text}}),\ \mathcal{Q}^\text{text}\bigr\}.
  \end{gathered}
\end{equation}
The superscript $*^\text{text}$ indicates that \emph{these are all textual descriptions} rather than the final actual representations.
Text
$\tilde{\mathcal{G}}^{\text{text}}_i$ and $\tilde{\mathcal{M}}^{\text{text}}_i$ describe the geometry and material of part $i$ and
$\tilde{\mathcal{K}}^{\text{text}}$ and $\tilde{\mathcal{J}}^{\text{text}}$ describe the kinematic structure and joints between these parts of the asset.

Beyond geometry, materials, and articulation considered in prior work~\citep{zhou2026articraft}, $\mathcal{Q}^\text{text}$ describes novel \emph{functional} properties.
First, it specifies coupled motions, such as a bin lid opening when its pedal is pressed.
It also defines interaction tasks in plain language, such as ``\emph{raise the mixing head and lift out the mixing bowl}''.
These functional constraints guide asset verification in \cref{sec:review_agent} and the generation of downstream task annotations in \cref{sec:downstream_task}, helping ensure that the asset is both physically plausible and functionally usable in downstream applications.

\heading{Blueprint Verification and Visual Reference}
$\mathcal{A}_{\mathrm{design}}$ checks the blueprint against the input requirements and corrects inconsistencies in parts, motions, and functions.
An image-generation model~\citep{openai2026gptimage2} converts the blueprint into a \emph{design board}
$\mathcal{D}$ (\cref{fig:pipeline}).
It lays out orthographic views of the articulated object,
an exploded view of the parts,
a kinematic diagram of the moving parts in their closed and open states,
and cutaway sections of the parts whose interiors matter.

\subsection{Coding Agent}%
\label{sec:coding_agent}

The \emph{Coding Agent}
$\mathcal{A}_{\mathrm{coding}}$
takes as input the user prompt and the optional reference images,
the blueprint $\mathcal{B}$ and design board $\mathcal{D}$ from the Design Agent,
the program $\Pi_{t-1}$ and the corresponding 3D asset $\mathcal{O}_{t-1}$ from its previous iteration,
the physics report $\mathcal{E}_{t-1}$,
and the review scores $\mathcal{S}_{t-1}$ and list of issues $\mathcal{R}_{t-1}$ from the Review Agent,
and produces the next program $\Pi_{t}$ and asset $\mathcal{O}_{t}$:
\begin{equation}
  \mathcal{A}_{\mathrm{coding}}\Bigl(
    \underbrace{(\bm{p},\mathcal{I})}_{\text{Input}},
    \underbrace{(\mathcal{B},\mathcal{D})}_{\mathcal{A}_{\mathrm{design}}},
    \underbrace{(\Pi_{t-1},\mathcal{O}_{t-1})}_{\mathcal{A}_{\mathrm{coding}}},
    \mathcal{E}_{t-1},
    \underbrace{(\mathcal{S}_{t-1},\mathcal{R}_{t-1})}_{\mathcal{A}_{\mathrm{review}}}
  \Bigr)
  \rightarrow (\Pi_t,\mathcal{O}_t),
  \label{eq:coding}
\end{equation}
Here, $t$ indexes a coding--review iteration.
The previous program $\Pi_0$, asset $\mathcal{O}_0$, physics report $\mathcal{E}_0$ (from the Physics Simulator in~\cref{sec:physics}), and review feedback $\mathcal{S}_0$ and $\mathcal{R}_0$ (returned by the Review Agent $\mathcal{A}_{\mathrm{review}}$, described in~\cref{sec:review_agent}) are absent on the first iteration.

Building on Articraft's SDK and execution harness~\citep{zhou2026articraft},
the agent iteratively \emph{edits} the program $\Pi_{t}$ (written in \textsc{Python}),
\emph{executes} it to obtain a corresponding asset $\mathcal{O}_{t}$,
and then \emph{estimates} its quality for iterative improvement.
We refer readers to~\cite{zhou2026articraft} for more details.
We extend the workflow for physical plausibility with three components:
\textbf{(i)} \emph{tools} for geometric and visual inspection (detailed in \cref{sec:app_tools}),
\textbf{(ii)} reusable \emph{standard connectors} for joint connections,
and \textbf{(iii)} \emph{physical simulation feedback} (\cref{sec:physics}) that guides repairs to physical attributes.

\begin{figure}[tb!]
    \centering
    \includegraphics[width=\linewidth]{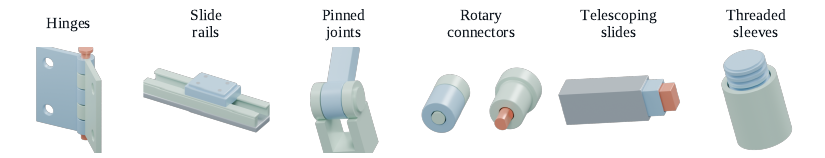}
    \caption{Illustrations of the six families of parameterized standard parts.}
    \label{fig:standard_parts}
\end{figure}

\heading{Standard Connectors for Joint Connections}
Inspired by standard connectors in mechanical design~\citep{poli2001design},
we extend the Articraft SDK with reusable, parameterized connectors, such as hinges, slide rails, and rotary connectors (\cref{fig:standard_parts}).
$\mathcal{A}_{\mathrm{coding}}$ only needs to select a connector and adjust its dimensions and mounting positions,
instead of designing a geometrically accurate connector from scratch.
This reduces floating connections and unintended interpenetration,
making the generated asset more collision-free as measured in \cref{sec:physics}.

\heading{Action Annotation}
Let $\tau$ denote a task described in natural language in $\mathcal{Q}^\text{text}$.
For each task $\tau$,
the Coding Agent first interprets the task to determine the required actions, and then
the Coding Agent records the target rigid part $\ell_\tau$, the geometric components forming its grasp or press region $V_\tau$, the joint to actuate $j_\tau$, and the fraction of joint travel required for success $\sigma_\tau$.
These are written to a manifest beside the exported meshes so that a downstream pipeline needs no manual labelling.

\subsection{Review Agent}%
\label{sec:review_agent}

The \emph{Review Agent}
$\mathcal{A}_{\mathrm{review}}$
evaluates the asset against the intended design and physical constraints.
It takes as input the user prompt and optional images,
the blueprint $\mathcal{B}$ and design board $\mathcal{D}$,
the current program $\Pi_t$ and asset $\mathcal{O}_t$,
and the corresponding physics report $\mathcal{E}_t$ from \cref{sec:physics}
and outputs a set of scores $\mathcal{S}_t$, one for each evaluation criterion, as well as a structured list of issues $\mathcal{R}_t$:
\begin{equation}
  \mathcal{A}_{\mathrm{review}}\Bigl(
    \underbrace{(\bm{p},\mathcal{I})}_{\text{Input}},
    \underbrace{(\mathcal{B},\mathcal{D})}_{\mathcal{A}_{\mathrm{design}}},
    \underbrace{(\Pi_{t},\mathcal{O}_{t})}_{\mathcal{A}_{\mathrm{coding}}},
    \mathcal{E}_{t}
  \Bigr)
  \rightarrow (\mathcal{S}_t,\mathcal{R}_t).
  \label{eq:review}
\end{equation}

The Review Agent critically reviews repairs against the blueprint and visual references to check that intended functions are preserved.
For example,
$\mathcal{A}_{\mathrm{coding}}$ may shorten a door's travel to pass collision checks,
while $\mathcal{A}_{\mathrm{review}}$ flags the reduced opening range as inconsistent with the blueprint.

Each issue identifies the affected component, supporting evidence, severity, and what the Coding Agent needs to fix.
The pipeline uses the scores and issue severity to request repairs when necessary; critical structural or functional failures cannot be offset by a high overall score.
Note that
the feedback $(\mathcal{S}_t,\mathcal{R}_t)$ guides the next coding iteration,
but the Review Agent itself does not alter the asset program.

\subsection{Physical Validation}%
\label{sec:physics}

We validate each candidate asset $\mathcal{O}_t$ through collision and stability checks in MuJoCo~\citep{todorov2012mujoco}.
The resulting physics report $\mathcal{E}_t$ records colliding part pairs and penetration depths at the tested joint configurations, together with tilt measurements under gravity, to guide coding and review.

\heading{Collision under Motion}
The collisions are determined based on whether different parts interpenetrate during the simulated motion sequences,
across sampled configurations of single-joint, double-joint, and multi-joint motions (details in Appendix~\ref{sec:app_sim}).
Specifically,
we first perform convex decomposition~\citep{wei2022coacd} to construct mesh collision geometry for simulation.
Then, we simulate the motion of each joint in isolation and record any collisions between parts.
Collisions that persist during individual joint operation require remediation.
Collision findings across motions include the affected part pairs, poses, and penetration depths, enabling the Review Agent to assess whether these collisions occur during normal operation (\cref{sec:review_agent}).

\heading{Stability under Gravity}
We simulate the assembly under gravity and measure its tilt angle relative to the initial pose to assess whether it remains upright.
We run the MuJoCo simulation for up to 5 seconds under gravity.
The 3D asset is considered unstable if it tilts beyond a certain threshold during the simulation.
The tilt measurements are recorded in $\mathcal{E}_t$ for review against the intended use.

\section{Benchmark}%
\label{sec:benchmark}

A key contribution of our paper is to build a comprehensive benchmark for physically grounded articulated 3D assets.
To this end,
we design a set of evaluation protocols and metrics,
and curate benchmark data with abundant annotations for these metrics.

\subsection{Evaluation Protocol and Metrics}%
\label{sec:evalution_protocol}

\heading{Spatial Fidelity}
The spatial fidelity metrics evaluate how well the 3D shape of the generated asset matches the reference asset in both the rest-pose and the articulated states. 
We first perform point sampling and align the meshes using ICP registration~\citep{besl1992icp}, then establish correspondences between parts based on their centers using the Hungarian matching algorithm~\citep{kuhn1955hungarian}.
Due to the difficulty of clearly distinguishing between states, outputs from different states may sometimes be confused.
To avoid erroneous penalties, we test both state pairings and retain the one with the lower alignment error across the two states as the final aligned result.

\emph{Appearance:}
After aligning the generated 3D asset to the reference asset, we render both using Blender at $1024^2$ resolution from three viewpoints per state.
We report the Intersection over Union (IoU) of the rendered masks, pixel-level PSNR, patch-level SSIM~\citep{wang2004ssim}, feature-level LPIPS~\citep{zhang2018lpips}, and CLIP score~\citep{radford2021clip} over all six images.

\emph{Rest-Pose and Articulated Geometry:}
We calculate 3D metrics in both states.
We report the object Chamfer Distance (oCD) for the object as a whole.
We also report the part-aware CD (pCD) and mean part IoU (pIoU) by averaging the metrics of all matched parts,
following~\citep{li2025particulate}.

\heading{Physical Properties}
Except for spatial fidelity on appearance, geometric and kinematics, we also assess the physical properties of the generated assets,
including the reconstructed \emph{materials} and \emph{mass}, the absence of \emph{collisions} during articulated motion, and \emph{stability} under gravity.

\emph{Materials:}
Following PBR-NeRF~\citep{wu2025pbr} and LMRM~\citep{youwang2026extracting},
we evaluate the materials of the generated assets by rendering the colour, metallic, and roughness buffers from the same viewpoints used for appearance evaluation,
and then calculating PSNR against corresponding renders of the reference asset.
To mitigate the impact of geometric inconsistencies, we also develop a protocol to transfer predicted materials to the ground-truth geometry so that the material attributes can be rendered and compared on the same geometry (results in Appendix~\cref{tab:material_transfer}).

\emph{Mass:}
We compare the mass $\hat{m}$ of the object against the corresponding ground-truth mass $m$.
Here,
we report three scores following~\citep{standley2017image2mass, zhai2024nerf2physics}: $\mathrm{ADE}=|\hat{m}-m|$ in kilograms, $\mathrm{APE}=|\hat{m}-m|/m$ and $\mathrm{MnRE}=\min(r,1/r)$, where $r=\hat{m}/m$.

\emph{Collision:}
We evaluate the collision of the generated asset at rest and during articulated motion.
We sample the motion in three levels: single-joint, double-joint, and multi-joint.
Specifically,
for single-joint motion,
we uniformly sample 9 states ranging from the rest state to its (predicted) upper-limit state.
For double-joint motion,
we then combine two joints to form a $9\times 9$ grid of motion states.
To efficiently evaluate multi-joint motion,
we randomly sample a state for each joint and perform a total of 200 trials.
To avoid penalizing collisions from joint combinations that violate motion dependencies,
we establish joint correspondences using Hungarian matching~\citep{kuhn1955hungarian},
reproduce each sampled configuration on the ground-truth asset, and exclude colliding part pairs that also contact in that asset.
At each sampling level,
we report the percentage of collisions over all sampled states.

\emph{Stability:}
Prior work~\citep{guo2024physically,li2025dso} evaluates stability under gravity primarily considering the whole object as a single rigid body,
with uniform mass distribution.
We extend this evaluation to a challenging benchmark of 50 articulated objects with high centers of mass.
We simulate each object in MuJoCo~\citep{todorov2012mujoco} for 10 seconds in both its rest and fully articulated states,
and report the percentage of stable assets separately for each state.
An object is considered stable if its maximum tilt angle relative to its initial pose remains below $30^\circ$,
its positional change is below $1$\,mm,
and its rotational change does not exceed $0.01$\,rad.

\heading{Robotic Task Performance}
Beyond spatial fidelity and physical plausibility,
we evaluate whether the generated assets are functionally usable and improve downstream task performance.

\emph{Success Rate:}
We evaluate whether training with generated articulated assets improves the performance of the vision-language-action policy $\pi_{0.5}$~\citep{intelligence2025pi05} on unseen instances.
For each held-out asset, we perform policy rollouts to evaluate the performance of $\pi_{0.5}$, which is fine-tuned on robotic demonstrations generated with predefined interactions on articulated assets.
We report the mean success rate for each task and the aggregate success rate across all tasks.

\emph{Asset Qualification:}
To further assess the simulator-readiness of the assets,
we additionally report how many generated assets can be utilized for robotic tasks.
The qualification consists of two stages: 
\textbf{(i)} \emph{physical stability and task feasibility (Round 1)}, which assesses whether an asset remains in its intended initial configuration after being placed in the simulator and whether it can physically reach the benchmark-defined success state under valid interactions; 
\textbf{(ii)} \emph{demonstration generation success (Round 2)}, which assesses whether the asset supports the generation of successful demonstrations using the benchmark's synthetic data generation pipeline within a predefined trial budget.

\heading{Overall Quality}
We also add a \emph{human} evaluation to assess the overall quality.
Participants are presented with the input text prompt and reference images,
rendered images of the generated assets in rest and open states,
and interactive \emph{MuJoCo} scenes in which each asset can be operated to inspect how its parts move,
with six methods' outputs anonymized and randomly ordered.
Each participant selects and ranks the top three of six results according to overall quality, considering visual fidelity, articulation, and physical interaction (detailed in \cref{sec:app_user}).

\subsection{Datasets}%
\label{sec:datasets}

We evaluate spatial fidelity on ArtVIP~\citep{jin2026artvip} and LightWheel~\citep{lightwheel},
each containing 50 articulated objects with diverse categories, shapes, and kinematics.
We use our system to reconstruct each of the assets from a text description and a reference image.
For physical evaluation, we use ArtVIP for material and collision, 50 articulated objects from ABO-500~\citep{collins2022abo} for mass, and a curated set of 50 objects for stability.
The downstream evaluation covers three tasks of RoboTwin 2.0~\citep{chen2025robotwin2} and eight from RoboCasa365~\citep{nasiriany26robocasa365:}, with disjoint native training and test assets, generated assets that correspond to the training distribution, and 50 demonstrations per asset.
More details of the dataset are provided in \cref{sec:app_corpora,sec:app_vla}.

\subsection{Baselines}%
\label{sec:baselines}

We compare against three groups of baselines:
\textbf{(i)} \emph{task-specific learned articulation models}, including the latest feed-forward method Instruct-Particulate~\citep{li2026instruct}, PhysX-Anything~\citep{cao2025physxanything} and URDF-Anything+~\citep{wu2026urdfanything};
\textbf{(ii)} \emph{specialized agentic methods} for articulated asset construction, Articraft~\citep{zhou2026articraft} and Articulate-Anything~\citep{le2024articulate};
and \textbf{(iii)} \emph{general-purpose coding agents}, Codex~\citep{openai2026codex}, Claude (Fable 5.1 high), and Antigravity (Agy) with Gemini-Flash-3.8.
For Articraft, Codex, and \ours{}, we report \texttt{GPT-5.5}, \texttt{GPT-5.6 Sol}, and \texttt{GPT-6 Astra} variants where available, as indicated in the tables.

\begin{table}[tb!]
\caption{\textbf{Spatial fidelity scores} on ArtVIP~\citep{jin2026artvip}, reported in the rest and the open state.
For simplicity, we only report the up-to-date results of Articraft and Codex, while the full results are available in~\cref{tab:spatial_full}.
\textbf{Bold} marks the best value and \underline{underline} marks the second-best value per metric.}%
\label{tab:spatial_short}
\vspace{-1em}
\centering
\setlength{\tabcolsep}{6pt}
\renewcommand{\arraystretch}{1.1}
\resizebox{\textwidth}{!}{
\begin{tabular}{@{}l ccccc ccc ccc @{}}
\toprule
\multirow{2}{*}{\textbf{Method}} & \multicolumn{5}{c}{\textbf{Appearance}} & \multicolumn{3}{c}{\textbf{Rest-Pose Geometry}} & \multicolumn{3}{c}{\textbf{Articulated Geometry}} \\
\cmidrule(lr){2-6} \cmidrule(lr){7-9} \cmidrule(lr){10-12}
& IoU$\uparrow$ & SSIM$\uparrow$ & PSNR$\uparrow$ & LPIPS$\downarrow$ & CLIP$\uparrow$ & oCD$\downarrow$ & pCD$\downarrow$ & pIoU$\uparrow$ & oCD$\downarrow$ & pCD$\downarrow$ & pIoU$\uparrow$ \\
\midrule
Instruct-Particulate & 0.6843 & 0.8581 & 17.12 & 0.2705 & 0.9167 & 0.0027 & 0.1143 & 0.2436 & \best{0.0010} & 0.1132 & 0.2732 \\
PhysX-Anything & 0.5259 & 0.8457 & 15.26 & 0.3899 & 0.8474 & 0.0098 & 0.2703 & 0.0965 & 0.0123 & 0.2989 & 0.0920 \\
URDF-Anything+ & 0.6294 & 0.8598 & 16.11 & 0.3819 & 0.8319 & 0.0062 & 0.2913 & 0.1262 & 0.0068 & 0.3034 & 0.1236 \\
Articulate-Anything & 0.6054 & 0.8197 & 14.44 & 0.3893 & 0.8444 & 0.0087 & 0.1923 & 0.1223 & 0.0097 & 0.2076 & 0.1254 \\
\midrule
Articraft (GPT-5.5) & 0.7195 & 0.8580 & 17.48 & 0.2781 & 0.9033 & 0.0026 & 0.0211 & 0.2795 & 0.0040 & 0.0235 & 0.2731 \\
Articraft (GPT-6 Astra) & \underline{0.7602} & 0.8754 & \underline{18.62} & \underline{0.2347} & \underline{0.9197} & 0.0022 & 0.0135 & 0.3244 & 0.0034 & \underline{0.0229} & \underline{0.3224} \\
Codex (GPT-6 Astra) & 0.7504 & 0.8700 & 18.16 & 0.2486 & 0.9169 & 0.0032 & 0.0206 & 0.3331 & 0.0039 & 0.0265 & 0.3095 \\
Claude (Fable 5.1) & 0.7316 & 0.8767 & 18.21 & 0.2619 & 0.9048 & 0.0015 & 0.0283 & 0.3309 & 0.0050 & 0.0407 & 0.3024 \\
Agy (Gemini-Flash-3.8) & 0.7542 & \best{0.8771} & 18.01 & 0.2562 & 0.9130 & \best{0.0010} & \underline{0.0129} & \underline{0.3553} & 0.0050 & 0.0235 & 0.3113 \\
\midrule
Ours (GPT-5.5 w/o phys.) & 0.7196 & 0.8669 & 17.64 & 0.2784 & 0.9100 & 0.0021 & 0.0162 & 0.2974 & 0.0052 & 0.0260 & 0.2825 \\
Ours (GPT-5.5) & 0.7374 & 0.8552 & 17.57 & 0.2826 & 0.9086 & 0.0016 & 0.0199 & 0.2956 & 0.0024 & 0.0248 & 0.2937 \\
Ours (GPT-5.6 Sol) & 0.7283 & 0.8701 & 17.63 & 0.2721 & 0.9150 & 0.0017 & 0.0179 & 0.2995 & 0.0026 & 0.0249 & 0.2991 \\
Ours (GPT-6 Astra) & \best{0.7675} & \underline{0.8769} & \best{18.65} & \best{0.2319} & \best{0.9239} & \underline{0.0011} & \best{0.0101} & \best{0.3587} & \underline{0.0021} & \best{0.0160} & \best{0.3417} \\
\bottomrule
\end{tabular}
}
\end{table}

\begin{figure}[tb!]
    \centering
    \includegraphics[width=\linewidth]{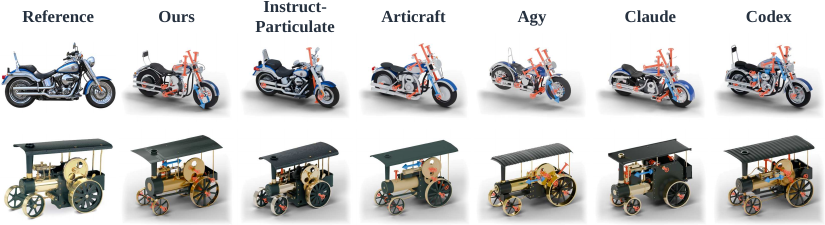}
    \vspace{-1.4em}
    \caption{
    \textbf{Qualitative comparison} of articulated assets generated by different models.
    \textcolor{orange}{Orange} rods indicate revolute axes; \textcolor{blue}{blue} arrows indicate prismatic directions. }
    \label{fig:main_vis}
\end{figure}

\begin{table}[tb!]
\caption{\textbf{Physical attribute scores} on ArtVIP~\citep{jin2026artvip} (\underline{Collision}, \underline{Material}), our own dataset (\underline{Stability}), and ABO-500~\citep{collins2022abo} (\underline{Mass}).
The full results are available in~\cref{tab:physics_full}.
\textbf{Bold} marks the best value and \underline{underline} marks the second-best value per metric.
$-$ indicates an unavailable result due to the problem scope of each method.
}%
\label{tab:physics_short}
\vspace{-1em}
\centering
\setlength{\tabcolsep}{2pt}
\resizebox{\textwidth}{!}{
\begin{tabular}{@{}l ccc cc ccc ccc @{}}
\toprule
\multirow{2}{*}{Method}
& \multicolumn{3}{c}{\textbf{Collision}}
& \multicolumn{2}{c}{\textbf{Stability}}
& \multicolumn{3}{c}{\textbf{Material}}
& \multicolumn{3}{c}{\textbf{Mass}} \\
\cmidrule(lr){2-4} \cmidrule(lr){5-6}
\cmidrule(lr){7-9} \cmidrule(lr){10-12}
& Single $\downarrow$ & Double $\downarrow$ & Multiple $\downarrow$
& ~Rest~ $\uparrow$ & ~Open~ $\uparrow$
& Base color $\uparrow$ & Metallic $\uparrow$ & Roughness $\uparrow$
& MnRE $\uparrow$ & ADE (kg) $\downarrow$ & APE $\downarrow$ \\
\midrule
NeRF2Physics & $-$ & $-$ & $-$ & $-$ & $-$ & $-$ & $-$ & $-$ & 0.601 & 10.78 & 0.509 \\
PUGS & $-$ & $-$ & $-$ & $-$ & $-$ & $-$ & $-$ & $-$ & 0.412 & 44.50 & 3.016 \\
\midrule
Instruct-Particulate & 70.4 & 90.4 & 98.7 & 68\% & 42\% & \best{12.00} & \best{18.02} & 10.40 & $-$ & $-$ & $-$ \\
PhysX-Anything & 58.4 & 77.6 & 81.3 & 74\% & 22\% & 10.06 & $-$ & $-$ & 0.302 & 198.07 & 19.292 \\
URDF-Anything+ & 84.6 & 99.3 & 100.0 & 64\% & 48\% & $-$ & $-$ & $-$ & $-$ & $-$ & $-$ \\
Articulate-Anything & 31.2 & 39.2 & 32.8 & 64\% & 44\% & 7.92 & $-$ & $-$ & $-$ & $-$ & $-$ \\
\midrule
Articraft (GPT-5.5) & 44.5 & 57.5 & 79.4 & 82\% & 78\% & 10.75 & $-$ & $-$ & $-$ & $-$ & $-$ \\
Articraft (GPT-6 Astra) & 45.5 & 60.7 & 78.8 & 82\% & 82\% & 11.75 & $-$ & $-$ & $-$ & $-$ & $-$ \\
Codex (GPT-6 Astra) & 32.4 & 47.0 & 63.7 & 86\% & 82\% & 9.98 & $-$ & $-$ & 0.763 & 6.63 & 0.303 \\
Claude (Fable 5.1) & 39.9 & 51.7 & 64.7 & 84\% & 86\% & 10.50 & $-$ & $-$ & 0.743 & 6.51 & 0.388 \\
Agy (Gemini-Flash-3.8) & 46.3 & 65.6 & 81.5 & 80\% & 78\% & 10.15 & $-$ & $-$ & \best{0.775} & \best{5.49} & \underline{0.296} \\
\midrule
Ours (GPT-5.5 w/o phys.) & 37.5 & 47.3 & 77.6 & 84\% & 82\% & 10.69 & 16.32 & 12.97 & 0.742 & 6.07 & 0.340 \\
Ours (GPT-5.5) & 3.5 & 6.4 & \best{12.3} & \underline{96\%} & \underline{94\%} & 11.48 & 16.28 & \underline{13.05} & 0.720 & 7.22 & 0.385 \\
Ours (GPT-5.6 Sol) & \underline{1.9} & \underline{4.8} & 16.4 & \best{98\%} & \best{96\%} & 10.57 & 15.98 & \best{14.32} & 0.670 & 8.25 & 0.566 \\
Ours (GPT-6 Astra) & \best{1.7} & \best{3.7} & \underline{13.0} & \best{98\%} & \best{96\%} & \underline{11.90} & \underline{16.38} & 12.36 & \underline{0.773} & \underline{5.86} & \best{0.260} \\
\bottomrule
\end{tabular}
}
\end{table}

\begin{figure}[t]
\centering
\begin{minipage}[t]{0.48\linewidth}
\centering
\textbf{\footnotesize (a) Collision}\par\smallskip
\includegraphics[width=\linewidth]{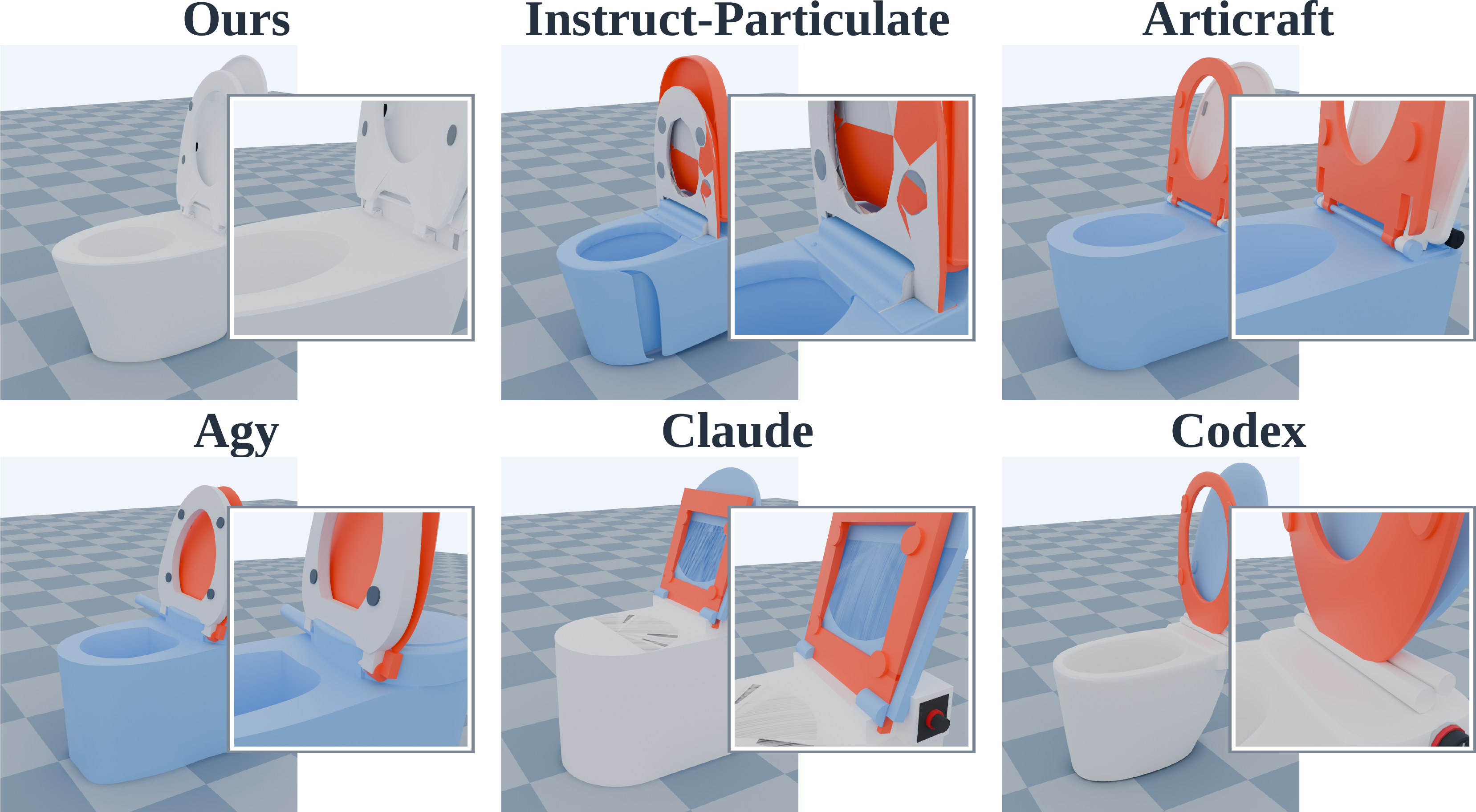}
\end{minipage}\hfill
\begin{minipage}[t]{0.48\linewidth}
\centering
\textbf{\footnotesize (b) Stability}\par\smallskip
\includegraphics[width=\linewidth]{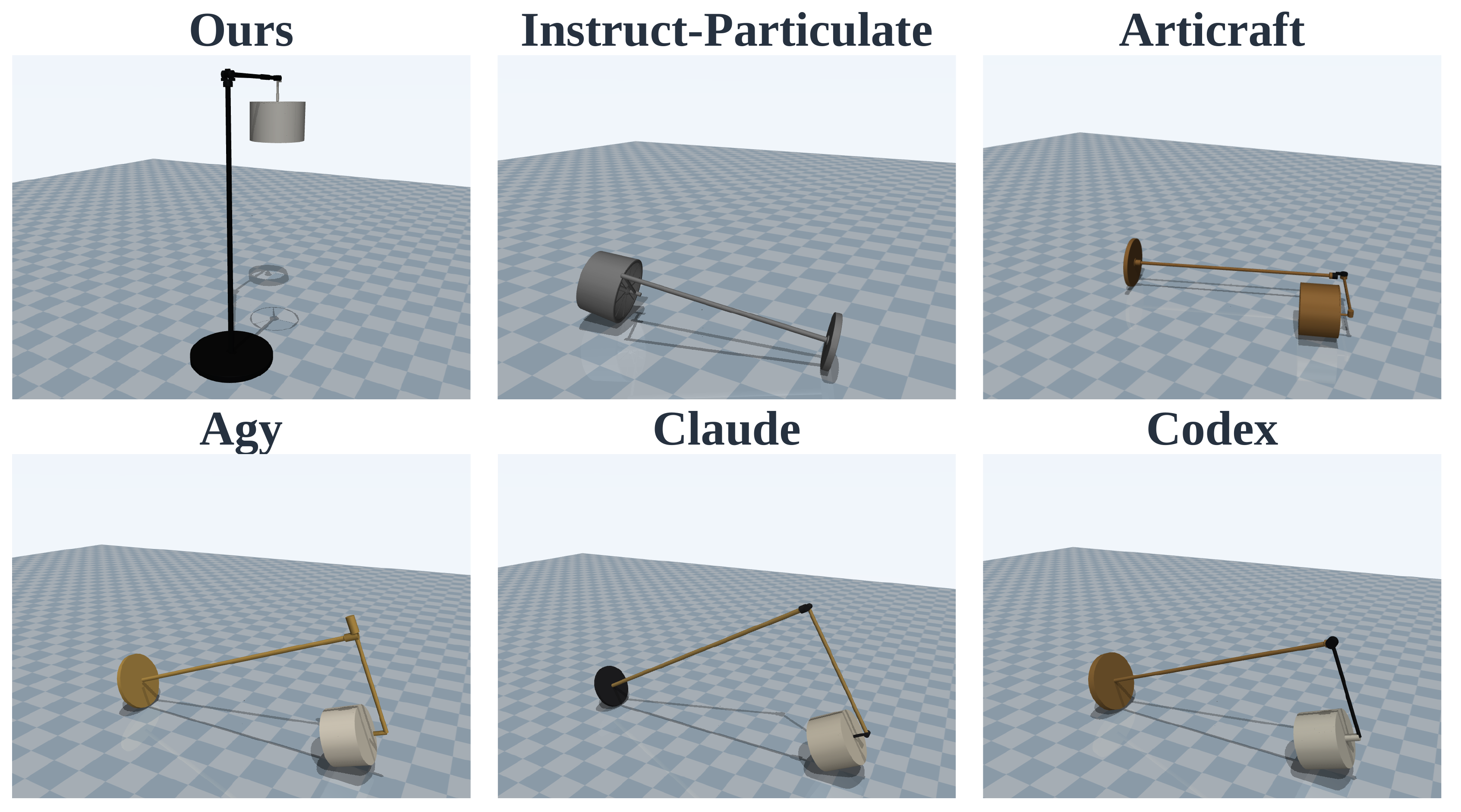}
\end{minipage}
\caption{\textbf{Qualitative collision and stability comparisons.} (a) Visualization of Collision. Red and blue denote the major intersected part pair. \ours{} is clean, whereas the other baselines involve collisions between components.
(b) Visualization of stability under gravity. \ours{} remains upright, whereas all five baselines topple after gravity is applied.
}
\label{fig:collision_stability}
\end{figure}

\begin{table*}[t]
\centering
\setlength{\tabcolsep}{4.5pt}
\caption{
\textbf{Downstream robotic evaluation of articulated assets.}
\textbf{(a)} Task success rate (\%) of $\pi_{0.5}$~\citep{intelligence2025pi05}
fine-tuned with different training-asset compositions, averaged over
3 RoboTwin 2.0~\citep{chen2025robotwin2} tasks and
8 RoboCasa365~\citep{nasiriany26robocasa365:} tasks.
We evaluate all policies on held-out native assets not seen during fine-tuning.
\textbf{(b)} 
Percentage of all assets that pass the qualification test (Round 1) and support successful demonstration generation (Round 2).
}
\label{tab:vla}

\begin{minipage}[t]{0.47\textwidth}
\centering
\textbf{(a) Task Success Rate (\%)} \\[3pt]
\resizebox{\textwidth}{!}{
\begin{tabular}{@{}lcc@{}}
\toprule
\textbf{Assets}
& \textbf{RoboTwin}
& \textbf{RoboCasa} \\
\midrule

Native only
& 80\% & 39\% \\

Articraft only
& 60\% & 5\% \\

\ours{} only
& 63\% & 27\% \\

Native + \ours{}
& \textbf{84\%} & \textbf{43\%} \\

\bottomrule
\end{tabular}
}
\end{minipage}
\hfill
\begin{minipage}[t]{0.51\textwidth}
\centering
\textbf{(b) Asset Qualification Rate (\%)} \\[3pt]
\resizebox{\textwidth}{!}{
\setlength{\tabcolsep}{3.5pt}
\begin{tabular}{@{}lcccc@{}}
\toprule
\multirow{2}{*}{\textbf{Assets}}
& \multicolumn{2}{c}{\textbf{RoboTwin}}
& \multicolumn{2}{c}{\textbf{RoboCasa}} \\
\cmidrule(lr){2-3}
\cmidrule(lr){4-5}
& \shortstack{\textbf{Round 1}}
& \shortstack{\textbf{Round 2}}
& \shortstack{\textbf{Round 1}}
& \shortstack{\textbf{Round 2}} \\
\midrule

Native
& 100\% & 100\%
& 98\% & 76\% \\

Articraft
& 62\% & 62\%
& 39\% & 8\% \\

\ours{}
& 100\% & 100\%
& 95\% & 44\% \\

\bottomrule
\end{tabular}
}
\end{minipage}

\end{table*}

\section{Results}%
\label{sec:exp}

\subsection{Analysis of Benchmark Results}

\heading{Main Results}
We compare \ours{} to the state-of-the-art baselines described in \cref{sec:baselines}.
We show spatial fidelity results in~\Cref{tab:spatial_short,fig:main_vis} and physical attribute results in~\Cref{tab:physics_short,fig:collision_stability}.
Without bells and whistles, ours (GPT-6 Astra) outperforms the other methods in 8 of 11 metrics for spatial fidelity in~\Cref{tab:spatial_short}.
More importantly,
it dramatically improves physical plausibility,
reducing collision occurrences to 1.7\% on single-joint and 13.0\% on multi-joint evaluations,
and achieving 98\% and 96\% stability in the rest and articulated states, respectively.
This highlights the significant advantage of our method in generating physically grounded articulated assets.
In the following, we summarize several key findings. More details are provided in~\Cref{sec:app_exp}.

\heading{Agentic generation methods can rival or even surpass state-of-the-art task-specific methods}
Compared to very recent state-of-the-art task-specific model Instruct-Particulate,
whose input mesh is generated by Hunyuan-3.1~\citep{tencent2026hunyuan3d31} from the reference image,
our method consistently achieves superior results in most metrics in~\Cref{tab:spatial_short,tab:physics_short}.
\Cref{fig:main_vis} further illustrates the visual results in two additional reference-guided examples: a motorcycle and a miniature steam traction engine.
Ours provides better appearance and articulated structure.
This reveals the potential impact and versatility of agentic generation methods.

\heading{Physical validation is the key to generating a ready-to-use asset}
As shown in \cref{tab:spatial_short,tab:physics_short}, we ablate ours (GPT-5.5) by disabling the physical validation (GPT-5.5 w/o phys.). 
Experiments demonstrate that, without physical validation, models can still achieve comparable performance in terms of appearance, geometry, and even kinematics in~\Cref{tab:spatial_short}.
However, their physical behavior degrades drastically with a high rate of collision detection in~\Cref{tab:physics_short}, indicating that such assets cannot execute their intended articulated motion within the simulator.
As shown in \cref{fig:collision_stability}, our model with physical validation is the only method that avoids component intersections and maintains stability within the simulator.
Therefore, compared with existing agentic or feed-forward methods, our method brings the generated assets much closer to being truly ready to use.
This suggests that physical validation in the generation loop is crucial for producing physically grounded assets.

\heading{Multi-agent design and independent review help eliminate the self-evaluation bias}
In~\cref{tab:spatial_short,tab:physics_short},
a fair comparison between ``Ours (GPT-5.5 w/o phys.)'' and ``Articraft (GPT-5.5)'' shows average improvements of 5.4\% in pCD, 4.9\% in pIoU, and 11.9\% in collision rates.
This demonstrates the effectiveness of the multi-agent design.
This further suggests that designing effective and efficient harnessing of agentic systems for physical 3D asset generation is
useful,
but it is a new and emerging area of physical 3D asset generation research.

\subsection{Assessing Robotic Tasks and Human Judgements}%
\label{sec:downstream_task}

\heading{Robotic Tasks}
We report task success and asset qualification rates in~\Cref{tab:vla}.
Based on the experimental results, we have the following key findings:
(i) \emph{Our assets are more suitable for robotic tasks:}
When trained on robotic demonstrations generated with \ours{}, the $\pi_{0.5}$ model achieves a higher success rate than that trained on Articraft demonstrations on both benchmarks (rows 2 \& 3 in~\Cref{tab:vla}(a));
(ii) \emph{Our assets can be a good extension to the current robotic training sets:} After incorporating our assets into the original training set, the newly trained model achieves a higher success rate on the same test set with unseen instances (rows 1 \& 4 in~\Cref{tab:vla}(a));
(iii) \emph{Our assets have better availability:} 
Compared to Articraft assets, our assets achieve higher pass rates (37 percentage points higher on average) across two rounds of qualification reviews on two benchmarks (\Cref{tab:vla}(b)),
suggesting that our assets are more ready for robotic tasks.

\begin{figure}[tb!]
\centering
\includegraphics[width=0.9\linewidth]{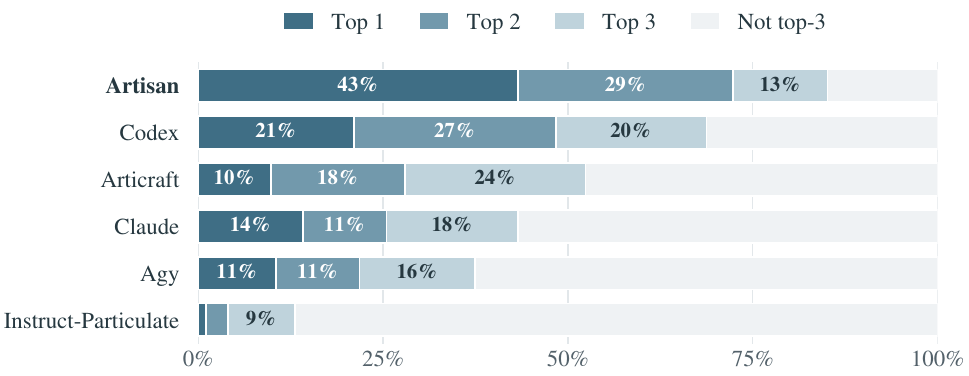}
\caption{\textbf{Human evaluation results.} Each bar stacks the rates at which a method is ranked first, second, and third; labels give each segment's rate.}
\label{fig:user_study_results}
\end{figure}

\heading{Human Judgements} Details of the human evaluation are provided in~\Cref{sec:app_user}.
The results in \Cref{fig:user_study_results} show that \ours achieves 43.3\% of the Top-1 votes and 85.1\% of the Top-3 votes, far surpassing the runner-up Codex (21.1\% in Top-1 and 68.7\% in Top-3).
These results are based on user feedback following object manipulation in the MuJoCo simulation environment, further demonstrating that the assets produced by our method are closer to being simulation-ready.

\ifshortdraft\else
\section{Conclusion}
\label{sec:conclusion}

We presented \ours{},
a pipeline that achieves physically grounded, articulated 3D asset generation by combining multiple agents and physical simulation feedback.
We also introduced a new comprehensive benchmark to evaluate various properties of the generated assets,
including appearance, geometry, kinematics, materials, collision, mass, stability, and even robotic task performance.
Empirically,
we found that \ours{} achieves the best overall performance across a wide range of evaluation metrics,
and even outperforms existing task-specific models in most areas.
We also found that multi-agent harnessing with physical simulation feedback significantly improves the spatial and physical fidelity of the generated assets.

Our findings motivate further research for our community: revisiting the role of feed-forward models, developing efficient agent harnesses across tasks, providing richer and more actionable simulator feedback, and extending simulators to capture a broader range of physical phenomena.
We hope this work will help the community measure and advance progress toward physically grounded 3D worlds.

\heading{Contribution and Acknowledgements}
XC led the project with management support from CZ, implemented the core algorithms, and conducted the experiments.
XC initiated the multi-agent pipeline, and SW and AV proposed the benchmark concept.
RZ contributed to the spatial and physical evaluation and analysis, with additional support from XC on the physical evaluation.
HS supported the downstream robotic evaluation by generating training data and conducting experiments on RoboTwin 2.0, while RL assisted with the RoboCasa365 experiments.
XC, RZ, RL, SW, AV, and CZ participated in weekly project meetings.
XC and RZ led the manuscript writing, with input from all other authors.
XC prepared the video demo and other visualizations.
RZ made the project page, with feedback from SW, XC and CZ.

We thank Matt Zhou for clarifying the technical details of Articraft;
Brandon Smart for proofreading the manuscript;
Tat-Jen Cham, Mengdi Wang, Stanislaw Szymanowicz, Edgar Sucar, Yuedong Chen, Qianyi Wu, and Jiawang Bian for helpful discussions on agentic 3D asset generation and related topics;
Mengdi Wang for suggesting the virtual laboratory concept;
and Jianfei Yang for providing access to the kitchen of their real-world robotic experiments.

This work is supported by NTU SUG-NAP and National Research Foundation, Singapore, under its NRF Fellowship Award NRF-NRFF17-2025-0009.
Ruining Li is supported by a Toshiba Research Studentship.
Andrea Vedaldi is supported by a Royal Society Faraday Discovery Fellowship FDF\textbackslash{}S2\textbackslash{}251003.

\clearpage

\section*{AI Use Statement}
We used generative artificial intelligence tools to assist in manuscript editing, including correcting grammar, refining sentence structure, and improving clarity and conciseness.
The authors reviewed and revised the AI-assisted text and verified its consistency with the methods and experimental results.
The authors assume full responsibility for the final manuscript, including its scientific claims and reported results.

\section*{Ethics Statement}
Our human evaluation involves inspecting rendered assets and interacting with virtual objects in a physics simulator, as described in Appendix~\ref{sec:app_user}.
Participants assess visual quality, articulation, and physical behavior through simulated interactions.
We report aggregate preference statistics, and the supplementary offline demo contains no real participant data.
The generated assets are intended for simulation and evaluation; their use in real-world robotic applications should undergo additional validation appropriate to the deployment setting.

\section*{Reproducibility Statement}
We describe the generation pipeline in Section~\ref{sec:method} and the datasets and evaluation metrics in Section~\ref{sec:benchmark}.
Appendices~\ref{sec:impl} and~\ref{sec:app_bench} provide implementation details, dataset construction procedures, and evaluation protocols, including the human-study setup.
Appendices~\ref{sec:app_exp}--\ref{sec:app_trace} document complete results, component ablations, generation budgets and computational costs, robotic experiment settings, and representative agent traces.
The supplementary material also includes an offline demonstration of the human-evaluation interface, allowing readers to inspect the rendered candidates and interact with the simulated assets.

\fi

\bibliographystyle{iclr27/iclr2027_conference}
\bibliography{refs,chuanxia_general,chuanxia_specific,vedaldi_general}

\ifshortdraft\else
\clearpage
\appendix

\section*{Appendix Contents}

\begingroup
\small
\setlength{\parindent}{0pt}

\noindent
\textbf{\hyperref[sec:impl]{\ref*{sec:impl}\quad Implementation Details}}
\dotfill
\textbf{\pageref*{sec:impl}}
\par\vspace{2pt}

\hspace*{1.8em}
\hyperref[sec:app_sdk]{\ref*{sec:app_sdk}\quad SDK Extensions}
\dotfill
\pageref*{sec:app_sdk}
\par

\hspace*{1.8em}
\hyperref[sec:app_tools]{\ref*{sec:app_tools}\quad Agent Tools}
\dotfill
\pageref*{sec:app_tools}
\par

\hspace*{1.8em}
\hyperref[sec:app_sim]{\ref*{sec:app_sim}\quad Physical Simulation}
\dotfill
\pageref*{sec:app_sim}
\par\vspace{6pt}

\noindent
\textbf{\hyperref[sec:app_bench]{\ref*{sec:app_bench}\quad Benchmark Details}}
\dotfill
\textbf{\pageref*{sec:app_bench}}
\par\vspace{2pt}

\hspace*{1.8em}
\hyperref[sec:app_spatial]{\ref*{sec:app_spatial}\quad Spatial Fidelity Protocol}
\dotfill
\pageref*{sec:app_spatial}
\par

\hspace*{1.8em}
\hyperref[sec:app_material]{\ref*{sec:app_material}\quad Material Transfer Protocol}
\dotfill
\pageref*{sec:app_material}
\par

\hspace*{1.8em}
\hyperref[sec:app_corpora]{\ref*{sec:app_corpora}\quad Evaluation Datasets}
\dotfill
\pageref*{sec:app_corpora}
\par

\hspace*{1.8em}
\hyperref[sec:app_user]{\ref*{sec:app_user}\quad Human Evaluation}
\dotfill
\pageref*{sec:app_user}
\par\vspace{6pt}

\noindent
\textbf{\hyperref[sec:app_exp]{\ref*{sec:app_exp}\quad Additional Results}}
\dotfill
\textbf{\pageref*{sec:app_exp}}
\par\vspace{2pt}

\hspace*{1.8em}
\hyperref[sec:app_results]{\ref*{sec:app_results}\quad Full Metric Tables}
\dotfill
\pageref*{sec:app_results}
\par

\hspace*{1.8em}
\hyperref[sec:ablation]{\ref*{sec:ablation}\quad Ablation}
\dotfill
\pageref*{sec:ablation}
\par

\hspace*{1.8em}
\hyperref[sec:app_gallery]{\ref*{sec:app_gallery}\quad Additional Qualitative Examples}
\dotfill
\pageref*{sec:app_gallery}
\par\vspace{6pt}

\noindent
\textbf{\hyperref[sec:app_cost]{\ref*{sec:app_cost}\quad API and Time Cost}}
\dotfill
\textbf{\pageref*{sec:app_cost}}
\par\vspace{6pt}

\noindent
\textbf{\hyperref[sec:app_vla]{\ref*{sec:app_vla}\quad Robotic Experiments}}
\dotfill
\textbf{\pageref*{sec:app_vla}}
\par\vspace{2pt}

\hspace*{1.8em}
\hyperref[sec:app_robotwin]{\ref*{sec:app_robotwin}\quad RoboTwin 2.0}
\dotfill
\pageref*{sec:app_robotwin}
\par

\hspace*{1.8em}
\hyperref[sec:app_robocasa]{\ref*{sec:app_robocasa}\quad RoboCasa365}
\dotfill
\pageref*{sec:app_robocasa}
\par\vspace{6pt}

\noindent
\textbf{\hyperref[sec:app_trace]{\ref*{sec:app_trace}\quad Agent Trace Examples}}
\dotfill
\textbf{\pageref*{sec:app_trace}}
\par\vspace{2pt}

\hspace*{1.8em}
\hyperref[sec:app_trace_mixer]{\ref*{sec:app_trace_mixer}\quad
Stand Mixer: Control Accessibility}
\dotfill
\pageref*{sec:app_trace_mixer}
\par

\hspace*{1.8em}
\hyperref[sec:app_trace_shoe]{\ref*{sec:app_trace_shoe}\quad
Shoe Cabinet: Drawer Connections}
\dotfill
\pageref*{sec:app_trace_shoe}
\par

\endgroup

\newpage

\section{Implementation Details}%
\label{sec:impl}

\subsection{SDK Extensions}%
\label{sec:app_sdk}
\heading{Physical Property Abstraction}
The SDK exposes three groups of physical attributes: part materials, part masses, and joint damping and friction.
For materials, the Coding Agent selects a named material preset and can directly adjust its PBR values, including albedo, metallicity, roughness, and opacity.
For mass, the agent reasons from the blueprint to estimate each part's mass, considering its dimensions, materials, and construction.
We deliberately avoid computing mass as material density times geometric volume, since real objects often combine multiple materials and contain hollow regions, making this approximation prone to large errors.
The SDK computes each part's default inertia from its geometry and estimated mass.
The agent can also adjust each part's inertial properties to reflect its mass distribution.
Joint damping and friction are configured according to the motions and mechanisms specified in the blueprint.

\heading{Standard Parts for Joint Connections}
We provide six commonly used families of parameterized \emph{standard parts}: hinges, slide rails, pinned joints, rotary connectors, telescoping slides, and threaded sleeves (\cref{fig:standard_parts}).
Each connector is built from matching components with shared dimensions, alignment, and clearance parameters.
For example, hinge halves share a pin axis, while a slide rail pairs a guide with a matching runner.
The Coding Agent adapts their dimensions and mounting positions to the asset, reducing repeated geometric design and mismatches between mating components.

\heading{Anchor-based Placement}
Authoring every pose in explicit coordinates requires the agent to calculate distances and orientations, and to update them whenever geometry changes.
We instead let the agent specify the intended spatial relationship and let the SDK solve the corresponding pose from the built geometry.
An \emph{anchor} names a geometric feature on an existing element: an oriented plane from a face, an axis from an edge or a round feature's center line, or a point from a corner.
Given a subject anchor $a_{\mathrm{sub}}$, reference anchors $\mathcal{A}_{\mathrm{ref}}$, and placement parameters $\theta$, the interface computes $T = \omega(a_{\mathrm{sub}}, \mathcal{A}_{\mathrm{ref}}; \theta)$,
where $T$ is the subject's rigid pose and $\theta$ specifies offsets, clearances, or a translation direction.
Common operations are $\omega \in \{\textsc{attach},\,\textsc{centre},\,\textsc{snap}\}$:
\textsc{attach} aligns mating anchors with the requested orientation and clearance;
\textsc{center} centers the subject between two references along a chosen axis;
and \textsc{snap} translates it along a specified direction until it reaches a support at the prescribed gap.
The SDK resolves the anchors in a shared coordinate frame and computes the rotation and translation required by the selected relation, preserving pose components that the operation does not constrain.
Operations can be composed, for example, to center a shelf between side panels and then seat it against the back panel.
Changes to the underlying geometry automatically propagate through these relationships, keeping dependent placements consistent without manually recomputing coordinates.

\subsection{Agent Tools}%
\label{sec:app_tools}
Following Articraft's tool interface~\citep{zhou2026articraft}, we expose actions for editing, execution, inspection, and reference retrieval (\cref{tab:tools}).
The Coding Agent can modify the asset program, while the Review Agent can inspect the asset and edit its own tests without changing the asset itself.

\begin{table}[tb!]
\centering
\footnotesize
\setlength{\tabcolsep}{3pt}
\renewcommand{\arraystretch}{1.12}
\begin{tabular}{@{}lp{0.28\linewidth}ccp{0.53\linewidth}@{}}
\toprule
Family & Tool(s) & C & R & Purpose \\
\midrule
\multirow[t]{2}{*}{Edit}
 & \texttt{write\_file}, \texttt{replace} & \checkmark & & Create or revise the asset program and geometry scripts. \\
 & \texttt{write\_test}, \texttt{edit\_test} & & \checkmark & Create or revise independent tests against the blueprint. \\
\midrule
\multirow[t]{3}{*}{Execute}
 & \texttt{run\_blender} & \checkmark & & Execute a geometry script and return the mesh and its diagnostics. \\
 & \texttt{compile\_model} & \checkmark & & Build the asset and return structural, geometric, and test feedback. \\
 & \texttt{run\_tests} & & \checkmark & Run the Review Agent's tests and return failures or errors. \\
\midrule
\multirow[t]{4}{*}{Inspect}
 & \texttt{probe\_model} & \checkmark & \checkmark & Query part and joint structure or geometric measurements. \\
 & \texttt{read\_gate\_evidence} & \checkmark & \checkmark & Retrieve motion and physics evidence for the current asset. \\
 & \texttt{request\_render} & \checkmark & \checkmark & Return RGB views or semantic maps with a distinct colour per part. \\
 & \texttt{crop\_region} & & \checkmark & Enlarge a selected region of an existing render. \\
\midrule
\multirow[t]{3}{*}{Reference}
 & \texttt{read\_file} & \checkmark & \checkmark & Read source code, task specifications, or SDK documentation. \\
 & \texttt{find\_examples} & \checkmark & & Retrieve curated examples of geometry and mechanisms. \\
 & \texttt{lookup\_cadquery\_api}, \texttt{lookup\_blender\_api} & \checkmark & & Look up signatures and documentation in the installed APIs. \\
\bottomrule
\end{tabular}
\caption{Tools available to the Coding (C) and Review (R) Agents. The Coding Agent edits the asset program; the Review Agent edits only its own tests.}
\label{tab:tools}
\end{table}

\heading{Editing and execution}
The Coding Agent writes or patches the construction program and can execute Blender scripts to create the corresponding geometry.
Compilation builds the asset and returns diagnostics for invalid structure, disconnected geometry, unintended overlaps, and failed tests.
The Review Agent separately writes and revises tests against the blueprint, then executes them on the asset to obtain failures and supporting details.

\heading{Inspection}
Both agents can query the part hierarchy and joint axes and limits, or run read-only probes to measure dimensions, clearances, overlaps, and connectivity.
These measurements identify the parts and spatial relationships involved in a suspected defect.
They can also retrieve the physics evidence associated with the current asset to examine failures under articulation or gravity.

\heading{Visual Inspection}
\texttt{request\_render} uses Blender to produce RGB images or semantic maps that assign a distinct flat colour to each part, making part boundaries easier to distinguish.
Both agents can request front, side, top, or perspective views in open or closed configurations and focus on named parts or joints.
The Review Agent can additionally specify joint positions and camera angles, or crop an existing image to inspect a local detail.
RGB views reveal appearance and proportions, while semantic maps help locate unexpected part visibility or apparent interpenetration for further geometric inspection.

\subsection{Physical Simulation}%
\label{sec:app_sim}
\heading{Collision under Motion}
We use MuJoCo to check collisions at sampled joint configurations, with mesh collision geometry constructed by CoACD using a decomposition threshold of 0.05, at most 32 convex hulls per mesh, and random seed 0.
The initial collision screen includes the rest pose, seven evenly spaced values along each joint's travel, pairwise $7\times7$ grids, and 200 random multi-joint samples drawn from the same joint-value grids.
We cap this initial set at 1,500 configurations and use sampling seed 0.
The functional constraints in $\mathcal{Q}^{\text{text}}$ guide checks of dependent motions: for example, an internal drawer is tested with its access door opened, so an invalid operation sequence is not mistaken for a geometric defect.
Collisions that persist during individual joint operation are sent to the Coding Agent for repair. For collisions found only when multiple joints move together, the Review Agent determines whether they occur during normal use and need repair.
For each affected part pair, the report records a representative configuration and penetration depth, allowing the agents to reproduce and inspect the finding.

\heading{Stability under Gravity}
For free-standing assets, we test stability under gravity in MuJoCo at rest and with each joint opened in turn. In each test, the assembly is placed on a ground plane with a free root, allowing it to settle or topple.
We simulate for up to 5\,s, stopping early when the root's linear and angular speeds remain at most $0.02$\,m/s and $0.05$\,rad/s, respectively, for 0.4\,s.
Tilt is measured as the angle between the root's initial and current up directions; a final tilt above $3^\circ$ is flagged for review.
The physics report $\mathcal{E}_t$ records, for each pose, the final and maximum tilt, whether the assembly settled, and the simulated duration, so the Review Agent can assess the result against the asset's intended use.

\section{Benchmark Details}%
\label{sec:app_bench}

\subsection{Spatial Fidelity Protocol}%
\label{sec:app_spatial}

\heading{Alignment and Correspondence}
We sample 100,000 points uniformly over each asset's surface and record their positions in both endpoint poses.
To resolve ambiguity in joint-axis motion direction, we test both ways of pairing the predicted poses with the ground-truth closed and open states.
For each pairing, we align both states with ICP from several initial orientations.
We select the pairing with the lower alignment error across both states, retaining the original ordering when the difference is negligible.
After alignment, we normalize both assets so that the longest side of the ground-truth asset's closed-state bounding box has unit length.
We compute each part's centroid as the mean of its sampled surface points.
Hungarian matching minimizes the total Euclidean distance between predicted and reference centroids in the aligned closed state, yielding one-to-one correspondences that are reused in the open state.

\subsection{Material Transfer Protocol}%
\label{sec:app_material}
To separate material errors from geometric incoherence, we transfer predicted materials onto the reference geometry.
We use Hungarian assignment to match parts by their relative positions, shapes, and motion between the rest and open states.
Within each matched part pair, component matching minimizes a weighted cost combining differences in normalized centers, bounding-box dimensions and aspect ratios, and relative surface area and bounding-box volume.
We transfer materials between the resulting component matches.
Numeric material parameters are copied directly, while texture inputs are reduced to surface-area-weighted means because the meshes do not share UV layouts.

Keeping the ground-truth geometry and joint configurations fixed, we render base-color, metallic, and roughness buffers in Blender at $512^2$ resolution from three viewpoints in each state.
We compute PSNR over visible object interiors, excluding boundary pixels.
Scores are averaged over the six images and then over objects.
\Cref{tab:material_transfer} reports the transferred-material scores.

\subsection{Evaluation Datasets}%
\label{sec:app_corpora}

\heading{Spatial Fidelity Evaluation Datasets}
LightWheel~\citep{lightwheel} is a library of simulation-ready OpenUSD assets modelled by artists and released under a CC-BY-NC license.
Recent articulation work adopts 243 of these assets, spanning 14 categories, as an evaluation set~\citep{li2025particulate,li2026instruct}.
It contains fourteen kitchen-appliance categories, including blenders, electric kettles, range hoods, and stovetops.
We sample 50 objects evenly across the fourteen categories (\cref{fig:bench_fidelity}a).

ArtVIP~\citep{jin2026artvip} comprises 992 digital-twin articulated objects over 9 categories and 37 subcategories.
Each asset is built by professional 3D modellers to a unified standard and ships with tuned dynamic parameters, embedded modular interaction behaviours, and pixel-level affordance annotations.
Its categories extend beyond kitchen appliances to furniture, fixtures, cleaning tools, stationery, storage, and mechanical equipment.
This diversity challenges methods to construct a wider range of geometries and mechanisms, motivating our use of ArtVIP as the main benchmark.
We select 50 objects through category-stratified sampling to span diverse object types and joint counts, and refine the selection by inspecting the models and renders.
\Cref{fig:bench_fidelity}b--c shows the subset's category and joint-count distributions.
\Cref{tab:corpus_artvip} lists all 50 objects using the same category grouping.
\begin{figure}[htbp]
    \centering
    \includegraphics[width=\linewidth]{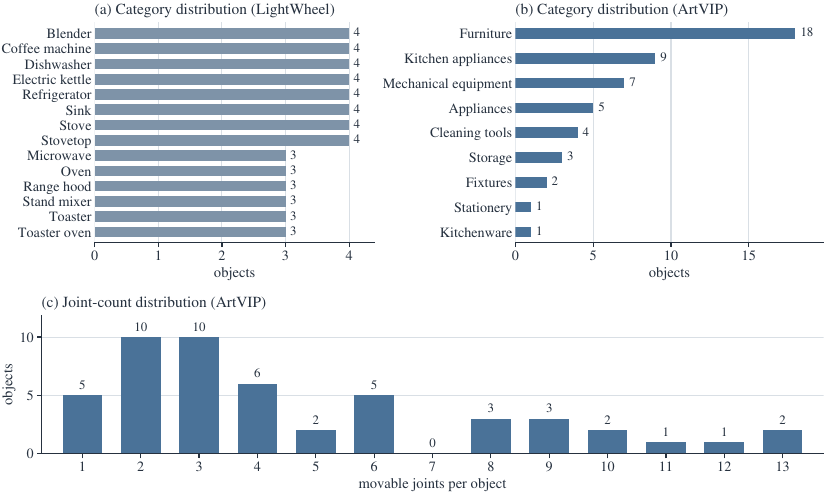}
    \caption{Composition of the spatial-fidelity evaluation subsets. \textbf{(a)} The LightWheel subset contains 50 objects across 14 categories, but all for kitchen-appliance categories. \textbf{(b)} The ArtVIP subset contains 50 objects across 9 large categories with 37 subcategories. \textbf{(c)} The ArtVIP objects have 1--13 movable joints each, with fixed joints excluded.}
    \label{fig:bench_fidelity}
\end{figure}

\begin{table}[htbp]
\centering
\footnotesize
\setlength{\tabcolsep}{4pt}
\renewcommand{\arraystretch}{1.12}
\caption{The 50 ArtVIP objects used for evaluation. Names follow the source asset identifiers.}
\label{tab:corpus_artvip}
\begin{tabular}{@{}>{\raggedright\arraybackslash}p{0.22\linewidth}r>{\raggedright\arraybackslash}p{0.67\linewidth}@{}}
\toprule
Category & Count & Object names \\
\midrule
Furniture & 18 & GULLABERG\_\allowbreak{}Shoe\_\allowbreak{}cabinet\_\allowbreak{}with\_\allowbreak{}2\_\allowbreak{}compartments\_\allowbreak{}grey\_\allowbreak{}88x30x122cm, KILBERGET\_\allowbreak{}chest\_\allowbreak{}of\_\allowbreak{}6\_\allowbreak{}drawers\_\allowbreak{}white\_\allowbreak{}anchor\_\allowbreak{}unlock\_\allowbreak{}function\_\allowbreak{}144x78cm, MUSKEN\_\allowbreak{}bedside\_\allowbreak{}table\_\allowbreak{} brown\_\allowbreak{}45x58cm, NORDKISA\_\allowbreak{}Dressing\_\allowbreak{}table\_\allowbreak{}bamboo\_\allowbreak{}76x47cm(2), RIDSPO\_\allowbreak{}Desk\_\allowbreak{}oak\_\allowbreak{}140X70cm, STOCKHOLM\_\allowbreak{}2025\_\allowbreak{}Sideboard\_\allowbreak{}oak\_\allowbreak{}veneer\_\allowbreak{}161x42x83cm, bookcase006, cabinet\_\allowbreak{}1, chest\_\allowbreak{}of\_\allowbreak{}drawers\_\allowbreak{}1, clock007, double\_\allowbreak{}door\_\allowbreak{}4, lab\_\allowbreak{}bench001, office\_\allowbreak{}chair\_\allowbreak{}3, poison\_\allowbreak{}cabinet001, table\_\allowbreak{}3, wardrobe\_\allowbreak{}1, wardrobe\_\allowbreak{}21, wardrobe\_\allowbreak{}22 \\
\addlinespace[3pt]
Kitchen appliances & 9 & capsule\_\allowbreak{}coffee\_\allowbreak{}maker\_\allowbreak{}3, dishwasher\_\allowbreak{}2, fridge\_\allowbreak{}1, l\_\allowbreak{}dishwasher\_\allowbreak{}1, microwave\_\allowbreak{}2, oven\_\allowbreak{}2, refrigerator\_\allowbreak{}6, rice\_\allowbreak{}cooker\_\allowbreak{}7, steamer \\
\addlinespace[3pt]
Mechanical equipment & 7 & dolly, elevator, incubator, medical\_\allowbreak{}cart001, medical\_\allowbreak{}monitor, microscope001, oscillator \\
\addlinespace[3pt]
Appliances & 5 & big\_\allowbreak{}electric\_\allowbreak{}fan\_\allowbreak{}9, desk\_\allowbreak{}lamp\_\allowbreak{}008, electric\_\allowbreak{}fan\_\allowbreak{}10, television003, washing\_\allowbreak{}machine\_\allowbreak{}2 \\
\addlinespace[3pt]
Cleaning tools & 4 & mop\_\allowbreak{}10, mop\_\allowbreak{}9, steping\_\allowbreak{}dustbin\_\allowbreak{}20, stepping\_\allowbreak{}dustbin\_\allowbreak{}6 \\
\addlinespace[3pt]
Storage & 3 & basket\_\allowbreak{}21, carton\_\allowbreak{}4, clamshell \_\allowbreak{}tool\_\allowbreak{}box\_\allowbreak{}2 \\
\addlinespace[3pt]
Fixtures & 2 & closestool\_\allowbreak{}3, faucet\_\allowbreak{}7 \\
\addlinespace[3pt]
Stationery & 1 & folder\_\allowbreak{}25 \\
\addlinespace[3pt]
Kitchenware & 1 & tongs\_\allowbreak{}10 \\
\bottomrule
\end{tabular}
\end{table}

\heading{Stability Evaluation Dataset}
We construct 50 object prompts around common destabilization mechanisms in everyday articulated objects, such as cantilevered loads, telescoping extensions, and tilting panels, as shown in \cref{fig:bench_stability}.
By varying both object categories and mechanisms, the set tests whether generated assets remain stable across diverse support geometries and joint configurations.

\begin{samepage}
\noindent
For example, one prompt used in the benchmark reads:
\begin{quote}
\small\itshape
A narrow hallway shoe cabinet with a slim upright carcass and three stacked tilt-out shoe bins. Each bin rotates outward and downward on its own horizontal revolute hinge at the bottom front edge of its opening, carrying the whole bin body clear of the carcass front. Hall-furniture scale, tall and shallow with a footprint barely deeper than a shoe.
\end{quote}
\end{samepage}
\Cref{fig:bench_stability} shows the distributions by the destabilization mechanism (left) and the category (right).
\begin{figure}[tb!]
    \centering
    \includegraphics[width=\linewidth]{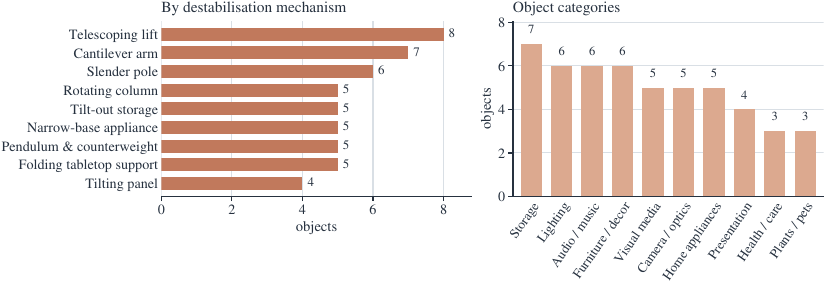}
    \caption{Composition of the stability dataset. \textbf{Left:} The 50 objects cover 9 destabilization mechanisms. \textbf{Right:} The objects span 10 categories defined by their primary use.}
    \label{fig:bench_stability}
\end{figure}

\heading{Mass Evaluation Dataset}
Following NeRF2Physics~\citep{zhai2024nerf2physics}, we use ABO-500, a collection of 500 objects from Amazon Berkeley Objects (ABO)~\citep{collins2022abo} with calibrated multi-view images and ground-truth masses from product listings.
We select 50 articulated objects from distinct product families to cover diverse articulation mechanisms and a ground-truth mass range of $0.086$--$205$\, kg.
For each object, we provide a brief prompt specifying its category and overall dimensions, together with a reference image.
\Cref{fig:bench_abo} shows the ground-truth mass distribution (left) and object categories (right), and \cref{tab:corpus_abo} lists the selected objects and their ground-truth masses.
\begin{figure}[htbp]
    \centering
    \includegraphics[width=\linewidth]{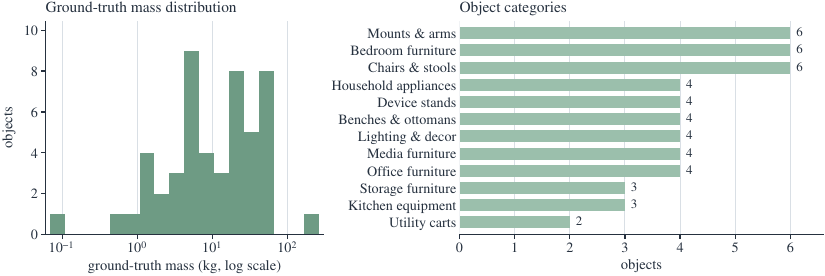}
    \caption{Composition of the mass-evaluation dataset. \textbf{Left:} The ground-truth masses range from $0.086$ to $205$\,kg. \textbf{Right:} The 50 objects span 12 categories grouped by primary use.}
    \label{fig:bench_abo}
\end{figure}

\begin{table}[htbp]
\centering
\scriptsize
\setlength{\tabcolsep}{5pt}
\renewcommand{\arraystretch}{1.12}
\begin{tabular}{@{}llr@{\hskip 16pt}llr@{}}
\toprule
ASIN & Category & Mass (kg) & ASIN & Category & Mass (kg) \\
\midrule
B006ZT4VA0 & tablet stand & 0.086 & B07TFQM3JZ & utility cart & 10.478 \\
B082L6NQZ7 & speaker wall mount & 0.617 & B084W2S628 & barstool & 10.977 \\
B075HXHKVF & jewelry box & 0.717 & B075X463J2 & ottoman & 12.247 \\
B00WRDS8H0 & laptop stand & 1.102 & B073GGGF7H & storage bench & 17.236 \\
B0742D9X4G & table lamp & 1.361 & B082L28P18 & file cabinet & 18.144 \\
B07MSYK43S & projector ceiling bracket & 1.383 & B007IHF36E & sideboard & 19.999 \\
B07MF1S1WL & wall sconce lamp & 1.397 & B07DMGM9T6 & kitchen cart & 21.319 \\
B07BL7H8N5 & frying pan with a lid & 1.896 & B07L48L3BH & kitchen cart & 23.501 \\
B07B8PXMRT & wall clock & 2.336 & B07H8JNCF3 & bedside table & 24 \\
B07WFNW48N & step stool & 3.452 & B07N6RT9SM & portable air conditioner & 25.301 \\
B00MIBN16O & monitor arm & 3.538 & B07TLDMQS2 & desk & 26 \\
B07XMV34PX & floor fan & 3.992 & B075X4WBNH & storage ottoman & 29.483 \\
B07DHNHZ2Y & dual monitor arm & 4.286 & B07H8P3K2R & TV stand & 32 \\
B07QVC3SLK & shop stool & 4.5 & B07TR659GF & TV stand & 32 \\
B07KR4BT31 & ceiling TV mount & 4.64 & B07GFFPKBG & TV stand & 33 \\
B07Q16JDH5 & keyboard stand & 4.74 & B07GFRP9RY & chest of drawers & 40 \\
B07T9NM5QR & DJ laptop stand & 4.931 & B07K6S2G5Z & corner desk & 47 \\
B07QWC7NYS & garage creeper seat & 5.198 & B07QBMQ737 & recliner & 48.081 \\
B07TX2D4WW & storage cube & 5.418 & B07H8T5XPT & TV stand & 50 \\
B07M9KPR1P & TV wall mount & 6.21 & B07K6RH9HS & desk & 51 \\
B07RKJCGXR & piano bench & 6.532 & B07Y2B738L & window air conditioner & 51.891 \\
B000S6N026 & paper shredder & 6.68 & B07H8SSY4F & sideboard & 56 \\
B07WBYQQ8F & garden utility wagon & 8.8 & B075YZPG5Y & dresser & 60.328 \\
B079VKDKC1 & bedside cabinet & 9.7 & B07GFRWJS7 & double bed frame with storage & 60.5 \\
B07TF7YLJC & desk chair & 10.4 & B07H8VCM68 & wardrobe & 205 \\
\bottomrule
\end{tabular}
\caption{The 50 ABO objects used for mass evaluation, ordered by ground-truth mass.}
\label{tab:corpus_abo}
\end{table}

\heading{Downstream Tasks Datasets}
We evaluate three articulated operation tasks from RoboTwin 2.0~\citep{chen2025robotwin2} and eight from RoboCasa365~\citep{nasiriany26robocasa365:},
using disjoint native training and test assets and generating additional assets matched to the training distribution.
For RoboTwin 2.0, each task contains 2--11 native assets;
we hold out 1--2 for testing and generate one asset per remaining training asset.
For RoboCasa365, we generate demonstrations on 10 native training assets and 10 generated assets, and hold out the remaining native training assets as test assets for each task.
We collect 50 demonstrations per asset using RoboTwin 2.0's task scripts with cuRobo~\citep{sundaralingam2023curobo} or RoboCasa365's MimicGen pipeline~\citep{mandlekar2023mimicgen}. Further details on the train-test split and demonstration collection can be found in \cref{sec:app_vla}.

\subsection{Human Evaluation}%
\label{sec:app_user}

We introduce an interactive user-study protocol in which participants assess generated assets through direct manipulation in MuJoCo alongside rendered views.
Making physical interaction part of human evaluation is essential for physically grounded assets: an object may look convincing yet fail to operate as expected.
Participants can move articulated parts and observe their responses to applied forces and contact, allowing functional failures to inform their judgments.

\heading{Assets and Comparisons}
The study covers 100 ground-truth assets, comprising the 50-object evaluation subsets from ArtVIP and LightWheel.
For each object, we compare \ours{} (GPT-6-Astra) with five representative baselines: Instruct-Particulate, Codex (GPT-6-Astra), Articraft (GPT-6-Astra), Claude (Fable-5.1), and agy (Gemini-3.8-Flash).
Participants receive the text prompt and reference images, inspect closed- and open-state renders from matched viewpoints, and interact with all six candidates in the simulator.
Method identities are hidden and candidate order is randomized.

\heading{Evaluation Criteria}
Participants select and rank the three best candidates by prompt alignment and overall quality, considering three complementary aspects:
\emph{visual quality}, whether appearance matches the reference image and description;
\emph{articulation}, whether movable parts are correctly separated and move about the intended axes and in the intended directions;
and \emph{physical interaction}, whether the asset behaves as its real-world counterpart would during manipulation.
These criteria guide a single overall ranking for each object.
For each method, we report top three selection rates, averaged equally across objects.

\heading{Study Interface}
\Cref{fig:user_study_ui} shows the web interface, which presents each object on one page with a 2D-render tab and a 3D-interaction tab that runs all candidates in MuJoCo in the browser.
Before submitting a ranking, participants must drag each candidate at least twice and spend at least three seconds in its 3D viewer.
\begin{figure}[t]
    \centering
    \setlength{\fboxsep}{0pt}%
    \setlength{\fboxrule}{0.4pt}%
    \begin{subfigure}{\linewidth}
        \centering
        \fcolorbox{black!25}{white}{\includegraphics[width=\dimexpr\linewidth-0.8pt\relax]{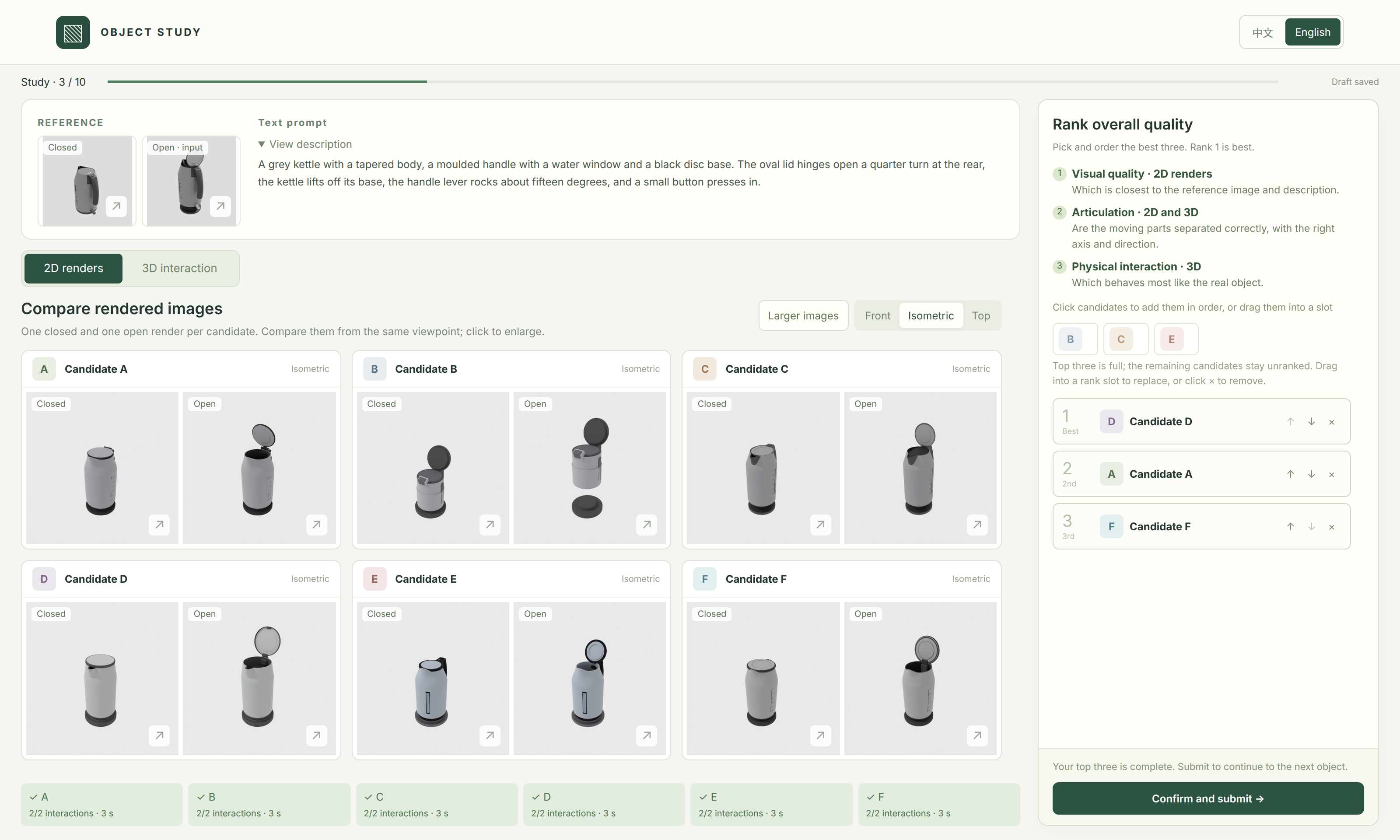}}
        \caption{\emph{2D renders} tab}
        \label{fig:user_study_ui_2d}
    \end{subfigure}
    \par\medskip
    \begin{subfigure}{0.495\linewidth}
        \centering
        \fcolorbox{black!25}{white}{\includegraphics[width=\dimexpr\linewidth-0.8pt\relax,trim={24bp 0bp 848bp 552bp},clip]{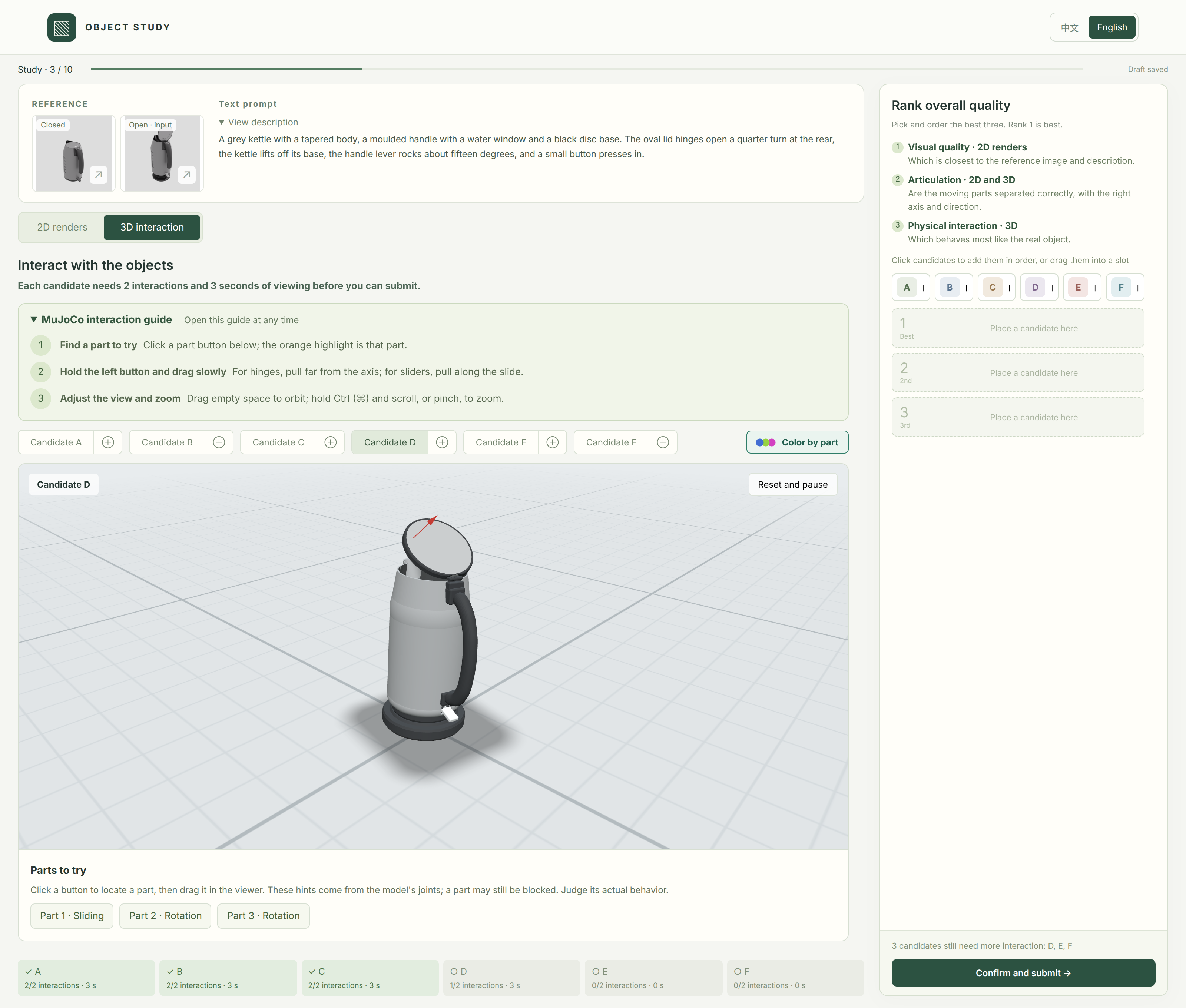}}
        \caption{\emph{3D interaction} tab}
        \label{fig:user_study_ui_3d}
    \end{subfigure}\hfill
    \begin{subfigure}{0.495\linewidth}
        \centering
        \fcolorbox{black!25}{white}{\includegraphics[width=\dimexpr\linewidth-0.8pt\relax,trim={24bp 0bp 848bp 552bp},clip]{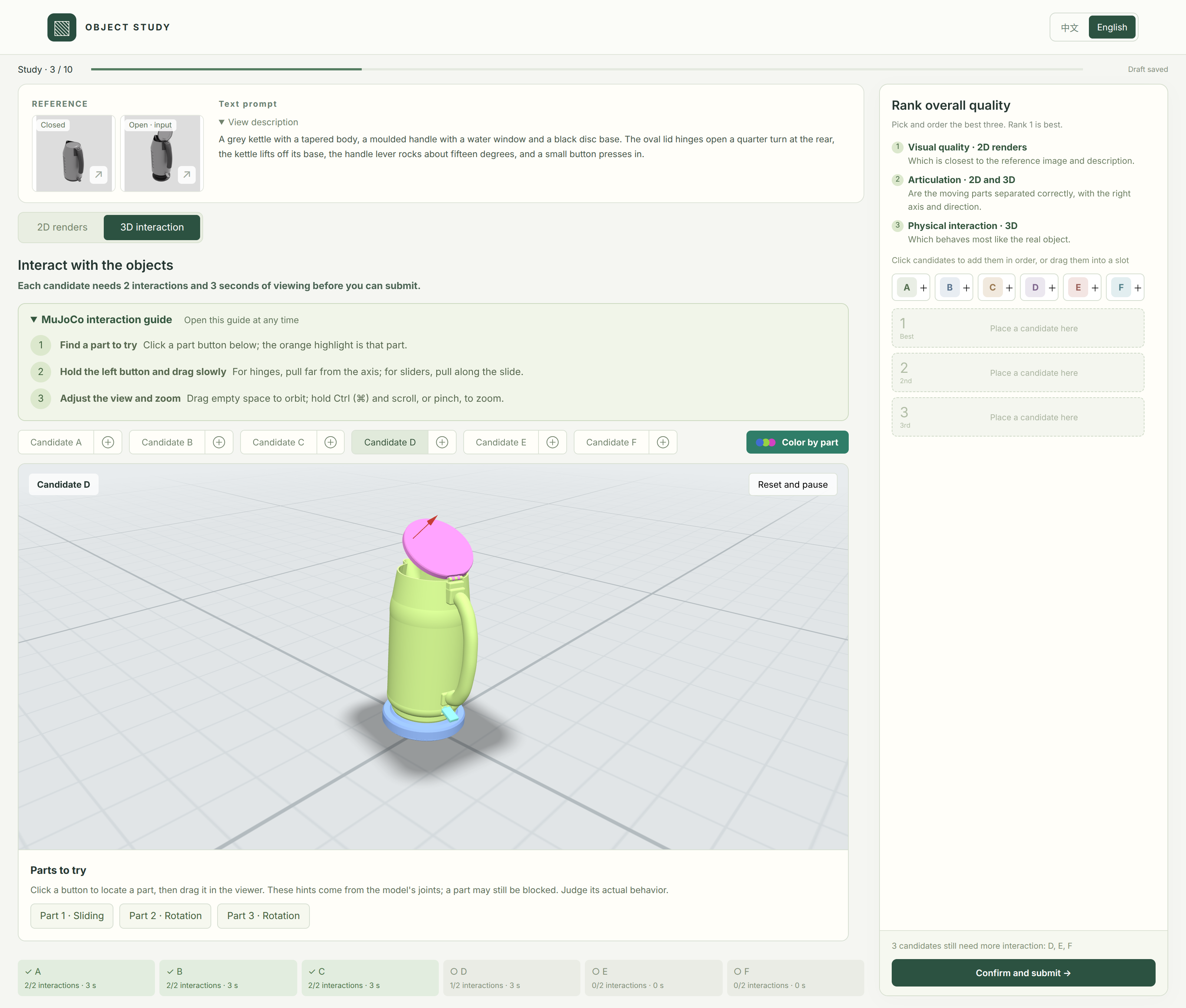}}
        \caption{\emph{3D interaction} tab with \emph{Color by part}}
        \label{fig:user_study_ui_partcolor}
    \end{subfigure}
    \caption{User study interface.
    \textbf{(a)}~Closed- and open-state renders of the six anonymized candidates from a shared viewpoint, next to the reference images and text prompt; the right panel lists the three criteria and collects the top-three ranking.
    \textbf{(b)}~The same page on the \emph{3D interaction} tab, cropped to the interaction area. Each candidate is loaded into an in-browser MuJoCo viewer, where the participant drags a part (red arrow: applied force); buttons below the viewer locate the movable parts, and the bottom bar tracks each candidate's interactions.
    \textbf{(c)}~The same moment with \emph{Color by part} enabled, which draws each rigid body in its own color to reveal the part decomposition.}
    \label{fig:user_study_ui}
\end{figure}

\heading{Participants and Results}
We recruit 30 non-expert college students from diverse academic backgrounds, each evaluating 10 randomly assigned objects.
The study yields 275 valid responses.
\Cref{fig:user_study_results} reports the selection rates for all six methods.

\FloatBarrier
\begingroup
\raggedbottom
\section{Additional Results}%
\label{sec:app_exp}

\subsection{Full Metric Tables}%
\label{sec:app_results}

\Cref{tab:spatial_full,tab:spatial_lightwheel} report complete spatial fidelity results on ArtVIP and LightWheel, including all evaluated model variants.
\Cref{tab:physics_full} reports collision rates on both datasets, together with stability, direct-material, and mass results.
\Cref{tab:material_transfer} separately reports material transfer onto the ground-truth geometry on ArtVIP.
All results follow the protocols in \cref{sec:evalution_protocol,sec:app_spatial,sec:app_material}.

\begin{table}[tb!]
\caption{\textbf{Complete spatial fidelity results on ArtVIP}. \textbf{Bold} and \underline{underline} indicate the best and second-best values per metric, including ties at the reported precision.}
\label{tab:spatial_full}
\centering
\small
\setlength{\tabcolsep}{3.5pt}
\renewcommand{\arraystretch}{1.1}
\resizebox{\linewidth}{!}{%
\begin{tabular}{@{}lccccccccccc@{}}
\toprule
\multirow{2}{*}{Method} & \multicolumn{5}{c}{\textbf{Appearance}} & \multicolumn{3}{c}{\textbf{Rest-Pose Geometry}} & \multicolumn{3}{c}{\textbf{Articulated Geometry}} \\
\cmidrule(lr){2-6} \cmidrule(lr){7-9} \cmidrule(lr){10-12}
& IoU$\uparrow$ & SSIM$\uparrow$ & PSNR$\uparrow$ & LPIPS$\downarrow$ & CLIP$\uparrow$ & oCD$\downarrow$ & pCD$\downarrow$ & pIoU$\uparrow$ & oCD$\downarrow$ & pCD$\downarrow$ & pIoU$\uparrow$ \\
\midrule
Instruct-Particulate & 0.6843 & 0.8581 & 17.12 & 0.2705 & 0.9167 & 0.0027 & 0.1143 & 0.2436 & \best{0.0010} & 0.1132 & 0.2732 \\
PhysX-Anything & 0.5259 & 0.8457 & 15.26 & 0.3899 & 0.8474 & 0.0098 & 0.2703 & 0.0965 & 0.0123 & 0.2989 & 0.0920 \\
URDF-Anything+ & 0.6294 & 0.8598 & 16.11 & 0.3819 & 0.8319 & 0.0062 & 0.2913 & 0.1262 & 0.0068 & 0.3034 & 0.1236 \\
Articulate-Anything & 0.6054 & 0.8197 & 14.44 & 0.3893 & 0.8444 & 0.0087 & 0.1923 & 0.1223 & 0.0097 & 0.2076 & 0.1254 \\
\midrule
Articraft (GPT-5.5) & 0.7195 & 0.8580 & 17.48 & 0.2781 & 0.9033 & 0.0026 & 0.0211 & 0.2795 & 0.0040 & 0.0235 & 0.2731 \\
Articraft (GPT-5.6-Sol) & 0.7297 & 0.8669 & 17.79 & 0.2658 & 0.9127 & 0.0024 & 0.0231 & 0.2755 & 0.0041 & 0.0282 & 0.2737 \\
Articraft (GPT-6-Astra) & \underline{0.7602} & 0.8754 & \underline{18.62} & \underline{0.2347} & \underline{0.9197} & 0.0022 & 0.0135 & 0.3244 & 0.0034 & \underline{0.0229} & \underline{0.3224} \\
Codex (GPT-5.5) & 0.6732 & 0.8543 & 16.92 & 0.3010 & 0.9003 & 0.0037 & 0.0259 & 0.2399 & 0.0049 & 0.0372 & 0.2421 \\
Codex (GPT-5.6-Sol) & 0.7126 & 0.8649 & 17.49 & 0.2826 & 0.9028 & 0.0040 & 0.0380 & 0.2617 & 0.0044 & 0.0413 & 0.2431 \\
Codex (GPT-6-Astra) & 0.7504 & 0.8700 & 18.16 & 0.2486 & 0.9169 & 0.0032 & 0.0206 & 0.3331 & 0.0039 & 0.0265 & 0.3095 \\
Claude (Fable 5.1) & 0.7316 & 0.8767 & 18.21 & 0.2619 & 0.9048 & 0.0015 & 0.0283 & 0.3309 & 0.0050 & 0.0407 & 0.3024 \\
Agy (Gemini-Flash-3.8) & 0.7542 & \best{0.8771} & 18.01 & 0.2562 & 0.9130 & \best{0.0010} & \underline{0.0129} & \underline{0.3553} & 0.0050 & 0.0235 & 0.3113 \\
\midrule
Ours (GPT-5.5 w/o phys.) & 0.7196 & 0.8669 & 17.64 & 0.2784 & 0.9100 & 0.0021 & 0.0162 & 0.2974 & 0.0052 & 0.0260 & 0.2825 \\
Ours (GPT-5.5) & 0.7374 & 0.8552 & 17.57 & 0.2826 & 0.9086 & 0.0016 & 0.0199 & 0.2956 & 0.0024 & 0.0248 & 0.2937 \\
Ours (GPT-5.6-Sol) & 0.7283 & 0.8701 & 17.63 & 0.2721 & 0.9150 & 0.0017 & 0.0179 & 0.2995 & 0.0026 & 0.0249 & 0.2991 \\
Ours (GPT-6-Astra) & \best{0.7675} & \underline{0.8769} & \best{18.65} & \best{0.2319} & \best{0.9239} & \underline{0.0011} & \best{0.0101} & \best{0.3587} & \underline{0.0021} & \best{0.0160} & \best{0.3417} \\
\bottomrule
\end{tabular}}
\end{table}

\begin{table}[tb!]
\caption{\textbf{Complete spatial fidelity results on LightWheel}. \textbf{Bold} and \underline{underline} indicate the best and second-best values per metric.}
\label{tab:spatial_lightwheel}
\centering
\small
\setlength{\tabcolsep}{3.5pt}
\renewcommand{\arraystretch}{1.1}
\resizebox{\linewidth}{!}{%
\begin{tabular}{@{}lccccccccccc@{}}
\toprule
\multirow{2}{*}{Method} & \multicolumn{5}{c}{\textbf{Appearance}} & \multicolumn{3}{c}{\textbf{Rest-Pose Geometry}} & \multicolumn{3}{c}{\textbf{Articulated Geometry}} \\
\cmidrule(lr){2-6} \cmidrule(lr){7-9} \cmidrule(lr){10-12}
& IoU$\uparrow$ & SSIM$\uparrow$ & PSNR$\uparrow$ & LPIPS$\downarrow$ & CLIP$\uparrow$ & oCD$\downarrow$ & pCD$\downarrow$ & pIoU$\uparrow$ & oCD$\downarrow$ & pCD$\downarrow$ & pIoU$\uparrow$ \\
\midrule
Instruct-Particulate & \best{0.8517} & \underline{0.9250} & \best{20.80} & \best{0.1085} & 0.9347 & 0.0039 & 0.0845 & \underline{0.2631} & 0.0033 & 0.0938 & \best{0.2890} \\
PhysX-Anything & 0.6225 & 0.9021 & 16.69 & 0.2158 & 0.8948 & 0.0082 & 0.1743 & 0.0986 & 0.0163 & 0.1967 & 0.0762 \\
URDF-Anything+ & 0.7126 & 0.9145 & 16.17 & 0.2455 & 0.8368 & 0.0063 & 0.3620 & 0.1200 & 0.0087 & 0.3728 & 0.1144 \\
Articulate-Anything & 0.6120 & 0.9066 & 16.45 & 0.2306 & 0.8596 & 0.0103 & 0.1627 & 0.1098 & 0.0149 & 0.1928 & 0.0924 \\
\midrule
Articraft (GPT-5.5) & 0.7891 & 0.9110 & 18.49 & 0.1534 & 0.9117 & 0.0019 & 0.0516 & 0.2050 & 0.0041 & 0.0635 & 0.1726 \\
Articraft (GPT-6-Astra) & 0.8301 & 0.9240 & \underline{20.46} & \underline{0.1154} & \best{0.9399} & 0.0018 & \best{0.0264} & 0.2511 & \underline{0.0032} & \underline{0.0357} & 0.2269 \\
Codex (GPT-5.5) & 0.7547 & 0.9054 & 18.01 & 0.1668 & 0.9052 & 0.0027 & 0.0657 & 0.1859 & 0.0047 & 0.0767 & 0.1648 \\
Codex (GPT-6-Astra) & 0.8124 & 0.9188 & 19.92 & 0.1248 & 0.9322 & \underline{0.0015} & 0.0329 & 0.2603 & 0.0045 & 0.0460 & 0.2282 \\
Claude (Fable 5.1) & 0.7771 & 0.9215 & 19.26 & 0.1525 & 0.9186 & 0.0023 & 0.0671 & 0.2097 & 0.0061 & 0.0815 & 0.1715 \\
Agy (Gemini-Flash-3.8) & 0.7826 & 0.9070 & 18.38 & 0.1504 & 0.9151 & 0.0022 & 0.0689 & 0.2290 & 0.0055 & 0.0813 & 0.1969 \\
\midrule
Ours (GPT-5.5 w/o phys.) & 0.7875 & 0.9153 & 18.55 & 0.1599 & 0.9095 & 0.0017 & 0.0437 & 0.2186 & 0.0045 & 0.0591 & 0.1974 \\
Ours (GPT-5.5) & 0.7933 & 0.9153 & 18.50 & 0.1562 & 0.9107 & 0.0018 & 0.0433 & 0.2127 & 0.0040 & 0.0583 & 0.1923 \\
Ours (GPT-6-Astra) & \underline{0.8320} & \best{0.9254} & 20.32 & 0.1174 & \underline{0.9355} & \best{0.0011} & \underline{0.0272} & \best{0.2901} & \best{0.0031} & \best{0.0320} & \underline{0.2512} \\
\bottomrule
\end{tabular}}
\end{table}

\heading{Spatial Fidelity}
With GPT-6-Astra, \ours{} leads on 8 of 11 ArtVIP metrics.
On LightWheel, it achieves competitive appearance scores and ranks first on 4 of the 6 geometry metrics, with the lowest object Chamfer distance in both states.
Instruct-Particulate, whose training mixture includes household assets from PartNet-Mobility and GRScenes~\citep{li2026instruct}, leads on 3 of the 5 LightWheel appearance metrics and open-state part IoU.
These results show that agentic construction can match or surpass learned articulation pipelines in spatial fidelity; its stronger performance on the more diverse ArtVIP benchmark further suggests broader generalization across object categories.

\begin{table}[tb!]
\caption{\textbf{Complete physical attribute results}. Collision rates are reported on ArtVIP and LightWheel; direct material PSNR uses ArtVIP, stability our stability dataset, and mass uses the ABO subset. Bold and underline mark the best and second-best values per metric; $-$ denotes unavailable results.}
\label{tab:physics_full}
\centering
\scriptsize
\setlength{\tabcolsep}{1.5pt}
\renewcommand{\arraystretch}{1.15}
\resizebox{\linewidth}{!}{%
\begin{tabular}{@{}l*{14}{c}@{}}
\toprule
\multirow{3}{*}{Method} & \multicolumn{6}{c}{\textbf{Collision (\%)}} & \multicolumn{2}{c}{\textbf{Stability}} & \multicolumn{3}{c}{\textbf{Material (dB)}} & \multicolumn{3}{c}{\textbf{Mass}} \\
\cmidrule(lr){2-7} \cmidrule(lr){8-9} \cmidrule(lr){10-12} \cmidrule(lr){13-15}
& \multicolumn{3}{c}{ArtVIP} & \multicolumn{3}{c}{LightWheel} & \multirow{2}{*}{Rest$\uparrow$} & \multirow{2}{*}{Open$\uparrow$} & \multirow{2}{*}{Base color$\uparrow$} & \multirow{2}{*}{Metallic$\uparrow$} & \multirow{2}{*}{Roughness$\uparrow$} & \multirow{2}{*}{MnRE$\uparrow$} & \multirow{2}{*}{ADE (kg)$\downarrow$} & \multirow{2}{*}{APE$\downarrow$} \\
\cmidrule(lr){2-4} \cmidrule(lr){5-7}
& Single$\downarrow$ & Double$\downarrow$ & Multi.$\downarrow$ & Single$\downarrow$ & Double$\downarrow$ & Multi.$\downarrow$ & & & & & & & & \\
\midrule
NeRF2Physics & $-$ & $-$ & $-$ & $-$ & $-$ & $-$ & $-$ & $-$ & $-$ & $-$ & $-$ & 0.601 & 10.78 & 0.509 \\
PUGS & $-$ & $-$ & $-$ & $-$ & $-$ & $-$ & $-$ & $-$ & $-$ & $-$ & $-$ & 0.412 & 44.50 & 3.016 \\
\midrule
Instruct-Particulate & 70.4 & 90.4 & 98.7 & 45.1 & 66.2 & 80.6 & 68\% & 42\% & \best{12.00} & \best{18.02} & 10.40 & $-$ & $-$ & $-$ \\
PhysX-Anything & 58.4 & 77.6 & 81.3 & 58.6 & 86.5 & 96.3 & 74\% & 22\% & 10.06 & $-$ & $-$ & 0.302 & 198.07 & 19.292 \\
URDF-Anything+ & 84.6 & 99.3 & 100.0 & 92.0 & 100.0 & 100.0 & 64\% & 48\% & $-$ & $-$ & $-$ & $-$ & $-$ & $-$ \\
Articulate-Anything & 31.2 & 39.2 & 32.8 & 24.1 & 38.7 & 62.5 & 64\% & 44\% & 7.92 & $-$ & $-$ & $-$ & $-$ & $-$ \\
\midrule
Articraft (GPT-5.5) & 44.5 & 57.5 & 79.4 & 43.4 & 66.0 & 83.1 & 82\% & 78\% & 10.75 & $-$ & $-$ & $-$ & $-$ & $-$ \\
Articraft (GPT-5.6-Sol) & 49.0 & 67.6 & 81.7 & $-$ & $-$ & $-$ & 82\% & 76\% & 11.06 & $-$ & $-$ & $-$ & $-$ & $-$ \\
Articraft (GPT-6-Astra) & 45.5 & 60.7 & 78.8 & 42.8 & 58.6 & 75.6 & 82\% & 82\% & 11.75 & $-$ & $-$ & $-$ & $-$ & $-$ \\
Codex (GPT-5.5) & 61.6 & 73.4 & 80.0 & 49.7 & 68.6 & 82.5 & $-$ & $-$ & 9.64 & $-$ & $-$ & $-$ & $-$ & $-$ \\
Codex (GPT-5.6-Sol) & 37.7 & 54.6 & 65.9 & $-$ & $-$ & $-$ & 80\% & 86\% & 10.66 & $-$ & $-$ & $-$ & $-$ & $-$ \\
Codex (GPT-6-Astra) & 32.4 & 47.0 & 63.7 & 25.2 & 38.0 & 55.6 & 86\% & 82\% & 9.98 & $-$ & $-$ & 0.763 & 6.63 & 0.303 \\
Claude (Fable 5.1) & 39.9 & 51.7 & 64.7 & 27.2 & 39.2 & 62.6 & 84\% & 86\% & 10.50 & $-$ & $-$ & 0.743 & 6.51 & 0.388 \\
Agy (Gemini-Flash-3.8) & 46.3 & 65.6 & 81.5 & 33.9 & 52.2 & 63.9 & 80\% & 78\% & 10.15 & $-$ & $-$ & \best{0.775} & \best{5.49} & \underline{0.296} \\
\midrule
Ours (GPT-5.5 w/o phys.) & 37.5 & 47.3 & 77.6 & 48.9 & 67.2 & 74.8 & 84\% & 82\% & 10.69 & 16.32 & 12.97 & 0.742 & 6.07 & 0.340 \\
Ours (GPT-5.5) & 3.5 & 6.4 & \best{12.3} & \best{2.1} & \best{3.9} & \best{7.8} & \underline{96\%} & \underline{94\%} & 11.48 & 16.28 & \underline{13.05} & 0.720 & 7.22 & 0.385 \\
Ours (GPT-5.6-Sol) & \underline{1.9} & \underline{4.8} & 16.4 & $-$ & $-$ & $-$ & \best{98\%} & \best{96\%} & 10.57 & 15.98 & \best{14.32} & 0.670 & 8.25 & 0.566 \\
Ours (GPT-6-Astra) & \best{1.7} & \best{3.7} & \underline{13.0} & \underline{3.2} & \underline{5.4} & \underline{9.7} & \best{98\%} & \best{96\%} & \underline{11.90} & \underline{16.38} & 12.36 & \underline{0.773} & \underline{5.86} & \best{0.260} \\
\bottomrule
\end{tabular}}
\end{table}

\heading{Physical Attributes}
As shown in \cref{tab:physics_full},
\ours{} (GPT-6-Astra) achieves lower collision rates than all baselines at all three sweep levels on both datasets, together with stability rates of 98\% at rest and 96\% when open.
It also achieves the lowest mass APE, although Agy has lower ADE and higher MnRE.
We also estimate mass by multiplying the densities of the agent-assigned materials by their geometric volumes.
This density-times-volume estimate yields an MnRE of 0.525, an ADE of 23.70\,kg, and an APE of 2.379, substantially underperforming our direct mass estimates.
This supports our design of reasoning directly about part mass to account for hollow and composite construction.
Compared with Instruct-Particulate, it achieves comparable base-color PSNR and higher roughness PSNR.

\begin{table}[tb!]
\caption{\textbf{Material transfer evaluation on ArtVIP}. PSNR (dB) uses predicted materials transferred to ground-truth geometry. \textbf{Bold} and \underline{underline} mark the best and second-best scores; $-$ denotes unavailable results due to the problem scope of methods.}
\label{tab:material_transfer}
\centering
\scriptsize
\setlength{\tabcolsep}{3pt}
\renewcommand{\arraystretch}{1.0}
\begin{tabular}{@{}lccc@{}}
\toprule
Method & Base color$\uparrow$ & Metallic$\uparrow$ & Roughness$\uparrow$ \\
\midrule
Instruct-Particulate & 10.07 & \best{18.14} & 10.54 \\
PhysX-Anything & 8.84 & $-$ & $-$ \\
Articulate-Anything & 7.55 & $-$ & $-$ \\
\midrule
Articraft (GPT-5.5) & 11.15 & $-$ & $-$ \\
Articraft (GPT-5.6-Sol) & 11.36 & $-$ & $-$ \\
Articraft (GPT-6-Astra) & \best{13.24} & $-$ & $-$ \\
Codex (GPT-5.5) & 10.32 & $-$ & $-$ \\
Codex (GPT-5.6-Sol) & 11.97 & $-$ & $-$ \\
Codex (GPT-6-Astra) & 10.72 & $-$ & $-$ \\
Claude (Fable 5.1) & 11.16 & $-$ & $-$ \\
Agy (Gemini-Flash-3.8) & 10.66 & $-$ & $-$ \\
\midrule
Ours (GPT-5.5 w/o phys.) & 10.75 & \underline{16.93} & \underline{12.80} \\
Ours (GPT-5.5) & 11.37 & 16.10 & 12.48 \\
Ours (GPT-5.6-Sol) & 11.29 & 16.26 & \best{14.39} \\
Ours (GPT-6-Astra) & \underline{12.08} & 16.40 & 12.12 \\
\bottomrule
\end{tabular}
\end{table}

\heading{Model Capacity}
For Articraft, Codex, and \ours{}, moving from GPT-5.5 to GPT-6-Astra improves every reported spatial fidelity metric on both datasets.
However, the GPT-6-Astra baselines still exhibit substantially higher collision rates than \ours{}.
Stronger models improve reconstruction quality, but do not remove the need for explicit physical validation.

\subsection{Ablation}%
\label{sec:ablation}

To assess the contributions of each component in \ours{},
we examine three components by removing the Design Agent, the Review Agent, or physical validation from \ours{} with GPT-5.6-Sol (\cref{tab:ablation}).
The geometry scores here are the average of the rest and open states.

\begin{table}[H]
\caption{\textbf{Component ablations with GPT-5.6-Sol}. Each variant removes the Design Agent, Review Agent, or physical validation. Appearance, geometry, collision, and mean part count are measured on ArtVIP; stability and mass use their respective datasets. Geometry scores average the rest and open states, and mass ADE is in kilograms. \textbf{Bold} and \underline{underline} mark the best and second-best scores per metric; part count is descriptive.}
\label{tab:ablation}
\centering
\scriptsize
\setlength{\tabcolsep}{1.2pt}
\renewcommand{\arraystretch}{1.15}
\resizebox{\linewidth}{!}{%
\begin{tabular}{@{}l*{14}{c}@{}}
\toprule
\multirow{2}{*}{Variant} & \multicolumn{2}{c}{\textbf{Appearance}} & \multicolumn{3}{c}{\textbf{Geometry}} & \textbf{Parts} & \multicolumn{3}{c}{\textbf{Collision (\%)}} & \multicolumn{2}{c}{\textbf{Stability}} & \multicolumn{3}{c}{\textbf{Mass}} \\
\cmidrule(lr){2-3} \cmidrule(lr){4-6} \cmidrule(lr){7-7} \cmidrule(lr){8-10} \cmidrule(lr){11-12} \cmidrule(lr){13-15}
& PSNR$\uparrow$ & CLIP$\uparrow$ & oCD$\downarrow$ & pCD$\downarrow$ & pIoU$\uparrow$ & $N_p$ & Single$\downarrow$ & Double$\downarrow$ & Multi.$\downarrow$ & Rest$\uparrow$ & Open$\uparrow$ & MnRE$\uparrow$ & ADE$\downarrow$ & APE$\downarrow$ \\
\midrule
w/o Design & 17.84 & \underline{0.9130} & 0.0035 & \best{0.0195} & 0.2788 & 5.82 & \best{1.0} & \best{3.5} & \underline{15.1} & \underline{96\%} & \best{96\%} & 0.630 & 9.57 & 0.710 \\
w/o Review & \best{17.95} & 0.9128 & \underline{0.0030} & 0.0211 & 0.2951 & 6.50 & 3.1 & 5.8 & \best{11.8} & \underline{96\%} & \underline{94\%} & 0.640 & 8.44 & 0.736 \\
w/o Physics & \underline{17.90} & 0.9091 & \underline{0.0030} & \underline{0.0199} & \underline{0.2952} & 6.66 & 35.7 & 54.1 & 73.2 & 90\% & 90\% & \underline{0.649} & \best{7.68} & \underline{0.628} \\
\midrule
Full & 17.63 & \best{0.9150} & \best{0.0022} & 0.0214 & \best{0.2993} & 6.78 & \underline{1.9} & \underline{4.8} & 16.4 & \best{98\%} & \best{96\%} & \best{0.670} & \underline{8.25} & \best{0.566} \\
\bottomrule
\end{tabular}}
\end{table}

\heading{Design Agent}
Without the Design Agent, part IoU falls from 0.2993 to 0.2788 and mass APE rises from 0.566 to 0.710.
The mean part count also falls from 6.78 to 5.82, alongside lower single- and double-joint collision rates.
This combination shows why collision rates must be interpreted together with structural fidelity: simpler assemblies can exhibit fewer collisions while reproducing the target less accurately.
These results suggest that explicit design encourages richer part structures while preserving the intended functionality.

\heading{Review Agent}
Removing the Review Agent reduces CLIP from 0.9150 to 0.9128 and part IoU from 0.2993 to 0.2951, while increasing mass APE from 0.566 to 0.736.
Its collision rates remain close to those of the full system, with a lower multi-joint rate.
The independent review thus helps assess whether the generated asset preserves the intended appearance, structure, and physical properties.

\heading{Physical Validation}
Without physical validation, single-joint collisions rise from 1.9\% to 35.7\% and multi-joint collisions from 16.4\% to 73.2\%, while open-state stability falls from 96\% to 90\%.
These findings reinforce the need for simulation feedback when constructing physically grounded assets.

\subsection{Additional Qualitative Examples}%
\label{sec:app_gallery}
In this section, we provide more qualitative comparisons for the ArtVIP and LightWheel benchmarks, as well as for stability and collision.
For Articraft, Codex, and our method, we use GPT-6-Astra to generate the assets. For Agy, we use Gemini-3.8-Flash, and for Claude we use Claude-Fable-5.1. For all LLMs, the thinking level is set to high.

\cref{fig:appendix_qualitative} provides more qualitative comparisons on the ArtVIP and LightWheel benchmarks, showing the reference image, the ground-truth asset, and the assets generated by the baselines and our method. The results show that our method reproduces the reference appearance more faithfully, recovers part decompositions that follow the ground-truth structure, and matches the ground-truth part motion better, whereas the baselines lose the reference colour, merge or fragment parts, and deviate from the ground-truth asset in the open state.

\begin{figure}[p]
    \centering
    \includegraphics[width=\linewidth,height=0.9\textheight,keepaspectratio]{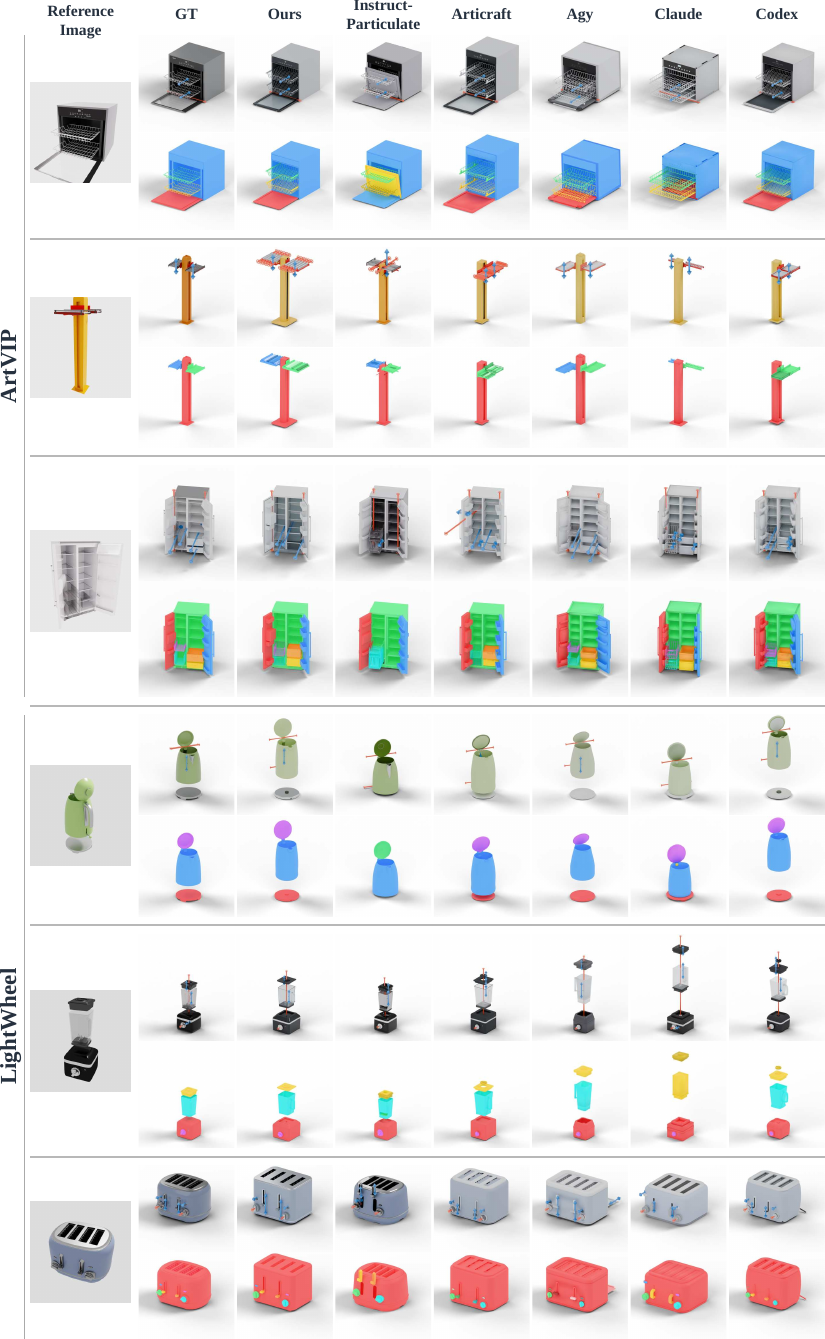}
    \caption{Additional qualitative comparisons on the ArtVIP and LightWheel benchmarks.}
    \label{fig:appendix_qualitative}
\end{figure}

\cref{fig:appendix_stability} visualizes more stability results under gravity in MuJoCo. The results show that our method can generate stable assets under gravity, while the baselines often produce unstable assets that fail the stability check.
\begin{figure}[p]
    \centering
    \includegraphics[width=\linewidth,height=0.9\textheight,keepaspectratio]{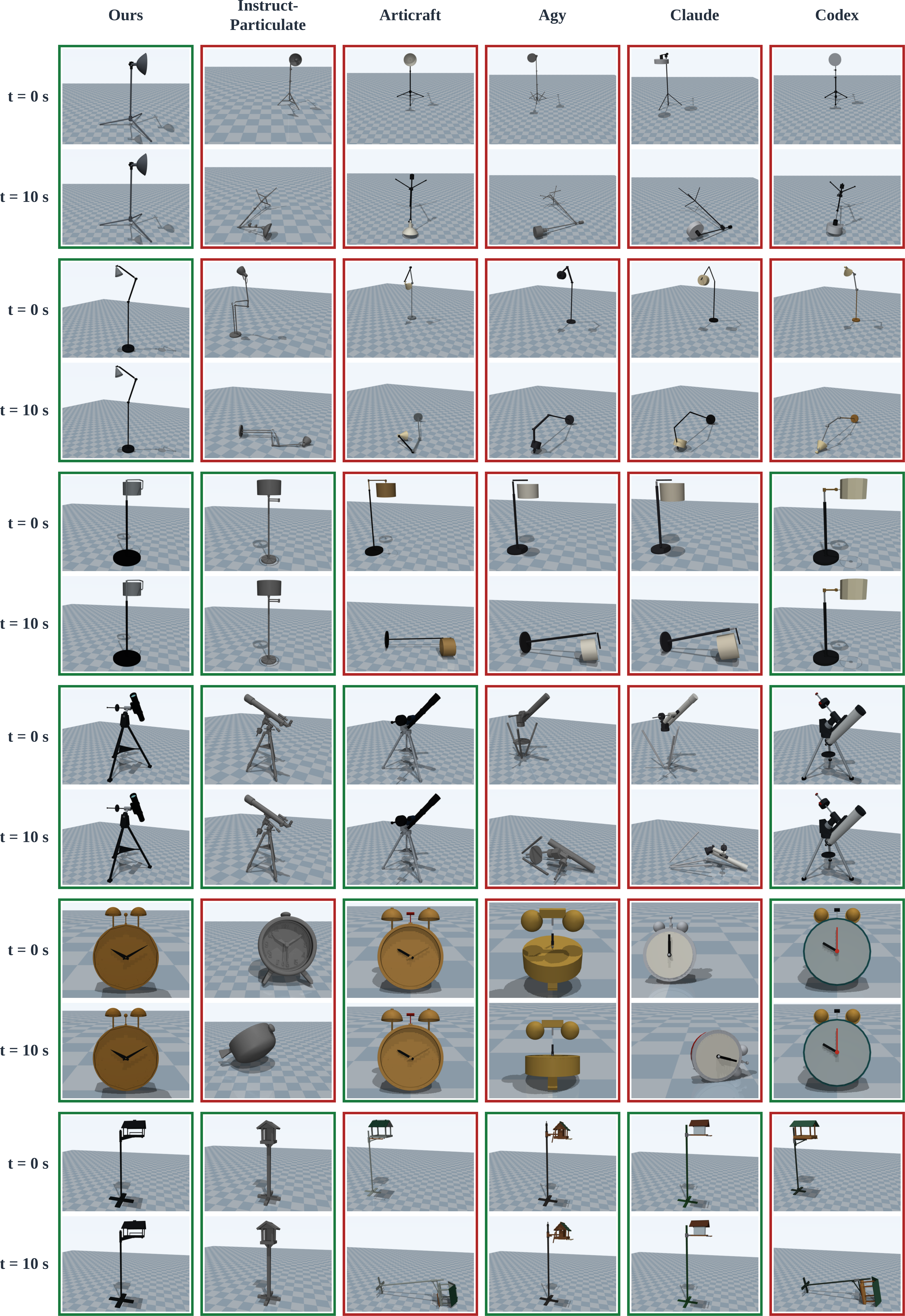}
    \caption{Additional visualizations of stability under gravity, each shown at rest and after ten seconds in MuJoCo. A green frame marks an asset that passes the stability check and a red frame one that does not.}
    \label{fig:appendix_stability}
\end{figure}

\cref{fig:appendix_collision} visualizes more collision results.  Our assets are collision-free while most baselines exhibit interpenetration, such as drawers, doors, and racks passing into the surrounding body.
\begin{figure}[p]
    \centering
    \includegraphics[width=\linewidth,height=0.9\textheight,keepaspectratio]{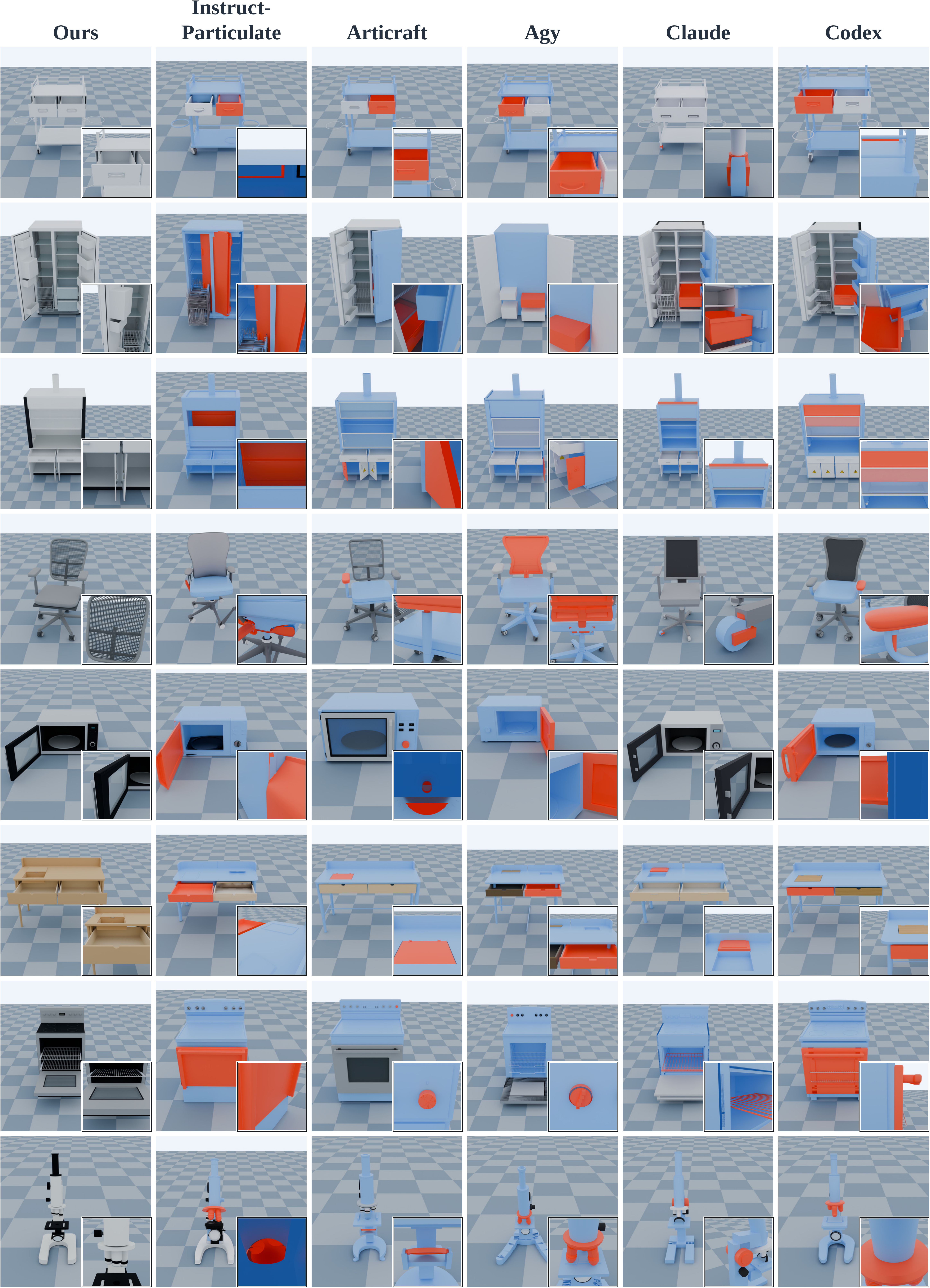}
    \caption{Additional collision comparisons. Red and blue highlight the most severely interpenetrating pair of parts, enlarged in the lower-right inset.}
    \label{fig:appendix_collision}
\end{figure}

\clearpage
\endgroup
\section{API and Time Cost}%
\label{sec:app_cost}
The Design Agent is allocated up to 12 turns for blueprint generation, and the Coding and Review Agents up to 150 turns each.
We set an API cost budget for each complete asset-generation run of \$10 for GPT-5.5 and GPT-5.6-Sol, and \$15 for GPT-6-Astra.
The repair loop between the Coding Agent and the Review Agent is bounded at five rounds and stops early after two consecutive rounds in which no rubric score rises.
To curb the rapid increase in API costs as conversation contexts grow, we trigger context compression at 110,000 tokens by capping and deduplicating tool outputs, dropping superseded renders, and summarizing earlier turns.
For each method--model configuration, we report the mean API cost and generation time per asset over all evaluated samples pooled from the available benchmarks.
\begin{table}[H]
\caption{\textbf{Average API cost and generation time per asset.}}
\label{tab:api_cost}
\centering
\footnotesize
\setlength{\tabcolsep}{5pt}
\renewcommand{\arraystretch}{1.1}
\begin{tabular}{@{}llrr@{}}
\toprule
Method & Model & Avg. cost (USD) & Avg. time (min) \\
\midrule
Ours (\ours{}) & GPT-5.6-Sol & 6.03 & 34.7 \\
Ours (\ours{}) & GPT-6-Astra & 5.79 & 34.3 \\
Ours (\ours{}) & GPT-5.5 & 7.35 & 33.9 \\
\midrule
Vanilla Codex & GPT-5.6-Sol & 0.33 & 9.9 \\
Vanilla Codex & GPT-6-Astra & 1.07 & 6.3 \\
Vanilla Codex & GPT-5.5 & 0.51 & 1.9 \\
\midrule
Articraft & GPT-5.6-Sol & 0.94 & 8.2 \\
Articraft & GPT-6-Astra & 0.95 & 5.4 \\
Articraft & GPT-5.5 & 1.32 & 6.0 \\
\midrule
Articulate-Anything & GPT-5.6-Sol & 0.46 & 6.8 \\
Claude Code CLI & claude-fable-5-1 & 0.54 & 1.5 \\
agy CLI & gemini-3.8-flash-high & 0.92 & 7.3 \\
\bottomrule
\end{tabular}
\end{table}

Although our method incurs higher API costs, it produces assets with fewer interpenetrations and greater stability, making them more usable in physical simulation (\cref{tab:physics_short}).
\Cref{fig:agent_cost_breakdown} shows the API cost and generation time breakdown of \ours{}, with simulation time reported separately.
\begin{figure}[H]
    \centering
    \includegraphics[width=\linewidth]{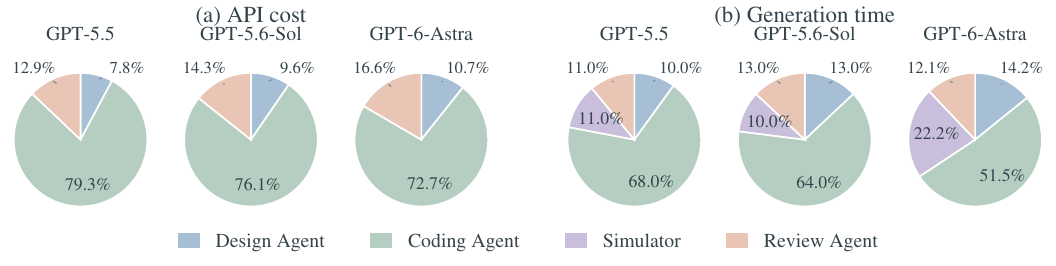}
    \caption{\textbf{API cost and generation time breakdown of \ours{}.} The left three pies show API cost shares of the Design, Coding, and Review Agents; the right three show generation time shares, including the simulator.}
    \label{fig:agent_cost_breakdown}
\end{figure}

\section{Robotic Experiments}
\label{sec:app_vla}

We evaluate whether our generated articulated assets provide usable training data for robotic manipulation. Specifically, we assess three properties: (1) physical validity and task feasibility, (2) compatibility with benchmark demonstration-generation pipelines, and (3) downstream policy generalization to unseen native assets.

\heading{Experimental Protocol}
For each task, we partition native benchmark assets into disjoint training and test sets. We generate the same number of additional assets as the native training set using both \ours{} and Articraft, where applicable, with a minimum number of 6. All generated assets are then checked for readiness through the 2 round quality tests and used exclusively for demonstration collection and policy training, while policy evaluation is performed on held-out native assets.

As shown in \cref{tab:vla} (b) and discussed in \cref{sec:evalution_protocol}, asset qualification consists of two rounds. \textbf{Round 1} verifies that an asset remains physically stable in its intended initial configuration and can reach the benchmark-defined success state under valid interactions. \textbf{Round 2} evaluates whether successful demonstrations can be collected using the benchmark's data-generation pipeline. We initially perform 50 collection trials per asset and retain assets yielding at least one successful demonstration. For retained assets, we allow up to 3,000 trials to collect 50 successful demonstrations. Assets failing qualification are excluded from policy training.

To isolate the effect of asset diversity, we use a clean tabletop configuration for RoboTwin and a fixed kitchen scene for RoboCasa. We compare policies trained on native assets (\textbf{Train}), generated assets (\textbf{\ours{}}), and their combination (\textbf{Train + \ours{}}), with Articraft included as a baseline where feasible.

\heading{Policy Training}
We fine-tune pretrained $\pi_{0.5}$~\citep{intelligence2025pi05} policies using LoRA~\citep{hu2021lora}, with rank 16 for the PaliGemma 2B backbone and rank 32 for the 300M action expert. All policies are trained until convergence with batch size 64 using AdamW ($\beta_1=0.9$, $\beta_2=0.95$, $\epsilon=10^{-8}$, weight decay $10^{-10}$, gradient clipping 1.0). The learning rate follows cosine decay with 1,000 warmup steps, a peak of $5\times10^{-5}$, and a configured 30,000-step decay horizon to $5\times10^{-6}$. EMA is disabled, and the vision encoder remains trainable.

\subsection{RoboTwin 2.0}
\label{sec:app_robotwin}

\heading{Setup}
We evaluate three articulated-object manipulation tasks: \emph{Open Microwave}, \emph{Open Laptop}, and \emph{Turn Switch}. Demonstrations are collected using the benchmark-provided scripts and cuRobo~\citep{sundaralingam2023curobo}. Contact points are annotated on assets for compatibility with the scripted manipulation pipeline. The asset splits are provided in \cref{tab:robotwin_asset_split}.

\begin{table}[tb!]
\centering
\setlength{\tabcolsep}{5pt}
\caption{
\textbf{Asset split in each RoboTwin 2.0 task.}
Native assets are split into disjoint training and held-out test sets.
\ours{} and Articraft denote generated assets that are physically stable and can be used for downstream evaluation and policy training.
}
\label{tab:robotwin_asset_split}
\begin{tabular}{lccccc}
\toprule
\textbf{Task}
& \textbf{Train}
& \textbf{Test}
& \textbf{\ours{}}
& \textbf{Train + \ours{}}
& \textbf{Articraft} \\
\midrule
 OpenMicrowave
  & 1 & 1 & 6 & 7 & 3\\
  OpenLaptop
  & 9 & 2 & 9 & 18 & 4\\
  TurnSwitch
  & 6 & 2 & 6 & 12 & 6\\
  \midrule
  \textbf{Total}
  & \textbf{16} & \textbf{5} & \textbf{21} & \textbf{37} & \textbf{13}\\
\bottomrule
\end{tabular}
\end{table}

\begin{table}[t]
\caption{\textbf{Round 2 qualification test results on RoboTwin 2.0.} Number of assets that pass the qualification test and are successful in generating demonstrations on the RoboTwin 2.0 tasks, for the native assets, the assets generated by \ours{}, and the assets generated by Articraft.}
\label{tab:robotwin_datagen}
\centering
\small
\setlength{\tabcolsep}{6pt}
\begin{tabular}{lccc}
\toprule
Task & Train & \ours{} & Articraft \\
\midrule
OpenMicrowave & 1/1 & 6/6 & 3/6 \\
OpenLaptop      & 9/9 & 9/9 & 4/9 \\
TurnSwitch    & 6/6 & 6/6 & 6/6 \\
\bottomrule
\end{tabular}
\par\vspace{2pt}
\footnotesize\textit{Entries denote \# successful assets / \# total assets.}
\end{table}

\heading{Results}
\cref{tab:robotwin_sr} reports policy success rates on held-out native objects. Augmenting native training assets with \ours{} improves average success rates over training with native assets alone, indicating that our generated assets introduce useful object-level variation rather than merely increasing performance on generated instances.

\begin{table}[tb!]
\caption{
\textbf{Task execution success rate (\%) of $\pi_{0.5}$ on RoboTwin 2.0.}
All policies are evaluated on held-out native test assets unseen during training.
Native, \textbf{Ours}, Native + \textbf{Ours}, and Articraft denote policies fine-tuned on
demonstrations collected from the corresponding asset sets.
Higher is better.
}
\label{tab:robotwin_sr}
\centering
\small
\setlength{\tabcolsep}{6pt}

\begin{tabular}{lcccc}
\toprule
\textbf{Task}
& \textbf{Train}
& \textbf{Train + \ours{}}
& \textbf{\ours{}}
& \textbf{Articraft}\\
\midrule

Open Microwave
& 64\% & 78\% & 40\% & 46\% \\

Open Laptop
& 98\% & 97\% & 97\% & 86\% \\

Turn Switch
& 78\% & 76\% & 53\% & 49\% \\
\midrule
\textbf{Average}
& \textbf{80\%}
& \textbf{84\%}
& \textbf{63\%}
& \textbf{60\%}\\
\bottomrule
\end{tabular}

\end{table}

\heading{Failure Analysis}
As shown in \cref{tab:robotwin_datagen}, Articraft exhibits collision-geometry failures that prevent successful manipulation despite the assets being loadable in simulation. In particular, 5 of 9 generated laptop assets and 3 of 6 microwave assets contain incorrect collision meshes that cause grasp failures and are excluded from the demonstration collection. These cases show that visually plausible articulated assets are not necessarily usable for robotic interaction: collision geometry must also permit physically valid contact and manipulation.

\subsection{RoboCasa365}
\label{sec:app_robocasa}

\heading{Setup}
We evaluate eight articulated-object manipulation tasks in RoboCasa365, using its MimicGen-based demonstration-generation workflow~\citep{mandlekar2023mimicgen}. Unlike RoboTwin's scripted motion-planning pipeline, MimicGen transfers manipulation trajectories from reference demonstrations to new object configurations. The evaluated asset splits are summarized in \cref{tab:robocasa_asset_split}.

\begin{table}[tb!]
\centering
\setlength{\tabcolsep}{4pt}
\caption{
\textbf{Asset split in each RoboCasa365 task.}
Native assets are split into disjoint training and held-out test sets.
\ours{} and Articraft denote generated assets physically able to be used for downstream evaluation and policy training.
Each training asset contributes 50 demonstration episodes.
}
\label{tab:robocasa_asset_split}
\begin{tabular}{lccccc}
\toprule
\textbf{Task}
& \textbf{Train}
& \textbf{Test}
& \textbf{\ours{}}
& \textbf{Train + \ours{}}
& \textbf{Articraft} \\
\midrule
OpenElectricKettleLid
& 8 
& 9 
& 10
& 18
& 0\\

OpenFridgeDrawer
& 5
& 11
& 1
& 6
& 0\\

OpenStandMixerHead
& 9
& 12
& 8
& 17 
& 0\\

OpenToasterOvenDoor
& 6
& 5
& 3
& 9
& 1\\

SlideDishwasherRack
& 10
& 4
& 1
& 11
& 0\\

SlideOvenRack
& 3
& 9
& 2
& 5 
& 0 \\

TurnOnStove
& 10
& 1
& 5
& 15 
& 0\\

TurnSinkSpout
& 10
& 12
& 5
& 15 
& 5\\

\midrule
\textbf{Total}
& \textbf{61}
& \textbf{63}
& \textbf{35}
& \textbf{96} 
& \textbf{6} \\
\bottomrule
\end{tabular}

\vspace{2pt}
{\footnotesize \textit{Entries denote the number of assets.}}
\end{table}

\heading{Asset Qualification and Demonstration Generation}
RoboCasa reveals a substantial difference between physical feasibility and compatibility with demonstration generation. Among 80 assets per group, 78 native assets, 76 assets generated by \ours{}, and 31 Articraft assets pass Round 1. After Round 2, these counts drop to 61, 35, and 6, respectively.

These results reveal two distinct bottlenecks. First, Articraft frequently fails physical qualification, with only 31 of 80 assets satisfying the stability and task-feasibility requirements. Second, even physically qualified assets do not necessarily support successful demonstration collection: 35 of our 76 Round-1-qualified assets and 61 of the 78 qualified native assets pass Round 2. Thus, demonstration-generation compatibility introduces additional constraints beyond physical plausibility.

\cref{tab:robocasa_datagen} further reports per-task demonstration-generation success rates. Articraft achieves zero successful demonstrations in six of eight task groups, with observed failures including physical instability, incorrect collision geometry, and mesh interpenetration. The results indicate that physically grounded geometry and articulation are essential for generating usable robotic training data.

\begin{table}[t]
\caption{\textbf{Round 2 qualification test results on RoboCasa365.} Number of assets that pass the qualification test and are successful in generating demonstrations on the RoboCasa365 tasks, for the native assets, the assets generated by \ours{}, and the assets generated by Articraft.}
\label{tab:robocasa_datagen}
\centering
\small
\setlength{\tabcolsep}{6pt}
\begin{tabular}{lccc}
\toprule
Task & Train & \ours{} & Articraft \\
\midrule
OpenElectricKettleLid & 8/10 & 10/10 & 0/10 \\
OpenFridgeDrawer      & 5/10 & 1/10 & 0/10 \\
OpenStandMixerHead    & 9/10 & 8/10 & 0/10 \\
OpenToasterOvenDoor   & 6/10 & 3/10 & 1/10 \\
SlideDishwasherRack   & 10/10 & 1/10 & 0/10 \\
SlideOvenRack         & 3/10 & 2/10 & 0/10 \\
TurnOnStove           & 10/10 & 5/10 & 0/10 \\
TurnSinkSpout         & 10/10 & 5/10 & 5/10 \\
\bottomrule
\end{tabular}
\par\vspace{2pt}
\footnotesize\textit{Entries denote \# successful assets / \# total assets.}
\end{table}

\heading{Downstream Policy Evaluation}
\cref{tab:robocasa_sr} reports policy success rates on held-out native assets under different training-asset compositions. These evaluations measure whether demonstrations collected from generated objects provide transferable training signals rather than performance gains restricted to generated instances.

\begin{table}[tb!]
\centering
\setlength{\tabcolsep}{5pt}
\caption{
\textbf{Task execution success rate (\%) of $\pi_{0.5}$ on RoboCasa365.}
All policies are evaluated on the held-out native test assets listed in
\cref{tab:robocasa_asset_split}, with 50 evaluation rollouts per test asset.
Native, Native + \ours{}, \ours{}, and Articraft denote policies fine-tuned on demonstrations
collected from the corresponding asset sets. For tasks with no qualified training assets, we assign a success rate of 0\% when computing the benchmark average. Higher is better.
}
\label{tab:robocasa_sr}
\begin{tabular}{lcccc}
\toprule
\textbf{Task}
& \textbf{Train}
& \textbf{Train + \ours{}}
& \textbf{\ours{}}
& \textbf{Articraft}\\
\midrule

OpenElectricKettleLid
& 11\% & 12\% & 14\% & 0\% \\

OpenFridgeDrawer
& 18\% & 23\% & 4\% & 0\% \\

OpenStandMixerHead
& 17\% & 29\% & 12\% & 0\% \\

OpenToasterOvenDoor
& 30\% & 35\% & 21\% & 23\% \\

SlideDishwasherRack
& 6\% & 13\% & 72\% & 0\% \\

SlideOvenRack
& 87\% & 84\% & 59\% & 0\% \\

TurnOnStove
& 92\% & 98\% & 8\% & 0\% \\

TurnSinkSpout
& 51\% & 49\% & 28\% & 19\% \\

\midrule
\textbf{Average}
& \textbf{39\%}
& \textbf{43\%}
& \textbf{27\%}
& \textbf{5\%} \\

\bottomrule
\end{tabular}
\end{table}

\heading{Failure Analysis}
We identify two distinct sources of demonstration-generation failure.

\textbf{Physical and collision failures.}
Some generated assets cannot be manipulated successfully because of unstable configurations, incorrect collision geometry, or mesh interpenetration. Such failures violate the requirements for physically executable manipulation and prevent reliable demonstration collection.

\textbf{Trajectory-transfer failures.}
Other assets are physically feasible but incompatible with the reference interactions used by MimicGen. For example, in \emph{OpenElectricKettleLid}, a generated kettle has a hinged button rather than the sliding button present in the source demonstrations. Although the generated mechanism is physically plausible, the transferred manipulation trajectory does not successfully actuate it (\cref{fig:mimicgen_lim}).

\begin{figure}[p]
    \centering
    \includegraphics[width=0.5\linewidth,height=0.9\textheight,keepaspectratio]{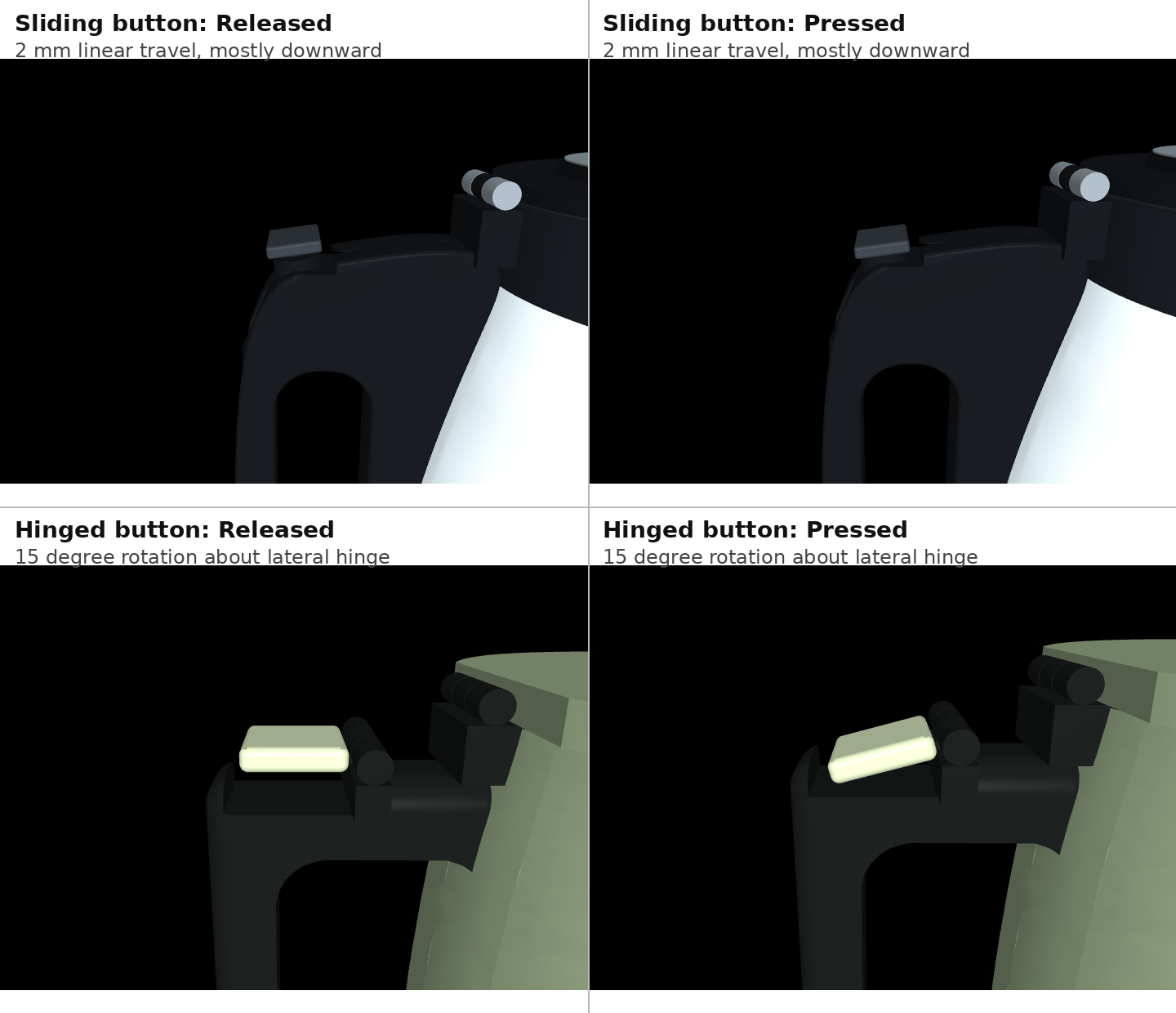}
    \caption{Close-up comparison of electric kettle buttons. \textbf{Top:} The native asset uses a sliding button in the human demonstration. \textbf{Bottom:} The generated asset uses a hinged button, which is plausible but prevents successful demonstration transfer with MimicGen.}
    \label{fig:mimicgen_lim}
\end{figure}

\FloatBarrier
\section{Agent Trace Examples}%
\label{sec:app_trace}
We show outputs, tool calls, and images from two real agent traces; \texttt{[...]} marks omissions.
Turns count recorded responses within each agent across rounds, with separate intake and blueprint counters for Design.

\begingroup
\raggedbottom
\renewtcolorbox{tracestep}[4]{%
  enhanced, breakable, sharp corners,
  colback=#1!6!white, colframe=#1!35!white, boxrule=0.3pt,
  borderline west={3.2pt}{0pt}{#1},
  left=5pt, right=5pt, top=2pt, bottom=2pt,
  toptitle=1.5pt, bottomtitle=1.5pt, lefttitle=5pt, righttitle=5pt,
  colbacktitle=#1, coltitle=white, fonttitle=\bfseries\scriptsize,
  title={#2\hspace{0.9em}\normalfont#3\hfill#4},
  title after break={#2\hspace{0.9em}\normalfont#3 (continued)},
  before upper={\def\tracelane{#1}},
  after skip=1.5pt, before skip=1.5pt,
}
\newcommand{\traceprompt}[1]{%
  \par\noindent{\small\textbf{Prompt.} \emph{``#1''}}\par\vspace{3pt}}
\newcommand{\tracecompactrow}[4][]{%
  \begin{minipage}[c]{0.705\linewidth}\scriptsize\raggedright
  #2
  \end{minipage}\hfill
  \begin{minipage}[c]{0.265\linewidth}\centering
  \includegraphics[width=\linewidth,height=2.05cm,keepaspectratio,#1]{#3}\par
  {\fontsize{6.5}{7.5}\selectfont #4\par}
  \end{minipage}\par}
\newcommand{\traceexcerpt}[1]{\par\smallskip\emph{``#1''}\par}
\newcommand{\tracephase}[3]{\par{\scriptsize
  \textbf{\color{\tracelane}Turns #1}\enspace\textbf{#2.} #3\par}}
\newcommand{\tracepic}[4][]{%
  \begin{minipage}[t]{#2\linewidth}\centering
  \includegraphics[width=\linewidth,height=1.75cm,keepaspectratio,#1]{#3}\par
  {\fontsize{6.5}{7.5}\selectfont #4\par}
  \end{minipage}}
\newcommand{\traceimages}[1]{\par\vspace{3pt}\noindent\hfill #1\hfill\mbox{}\par}

\captionsetup{font=small,skip=5pt}

\subsection{Stand Mixer: Control Accessibility}
\label{sec:app_trace_mixer}
\begingroup
\setlength{\parindent}{0pt}
{\small\textbf{Prompt.}
\emph{``A slate blue stand mixer with a heavy rounded head, a matching pedestal and a steel mixing bowl. The head tips back a quarter turn, the bowl lifts off its seat, the whisk spins on the head, the side speed dial turns, and a release button presses in.''}\par}

\begin{tracestep}{agdesign}{Design Agent}{Blueprint and design board}{\tchip{agdesign!75}{design specified}}
\scriptsize
\textbf{Intake Turn 1.} Identifies the appliance, visible motions, and user interactions from the input image.\\
\textbf{Blueprint Turns 1--4.} Authors the blueprint (1), checks its consistency (2), corrects the accessory-cap location (3), and verifies the revision (4). A design board is generated from the resulting blueprint.
\par\smallskip
\textbf{Selected outputs (abbreviated).}\\
\textbf{Parts:} base, mixing head, whisk, speed selector, mixing bowl.\\
\textbf{Geometry/materials:} a rounded $0.29\times0.46\times0.31$\,m slate-blue base; a hollow stainless-steel bowl with an integral handle; an open wire whisk; [...].\\
\textbf{Structure/joints:} base $\rightarrow$ head, revolute about $+X$, travel $1.48$\,rad; head $\rightarrow$ whisk, continuous rotation; selector rotation $2.36$\,rad and bowl lift $0.22$\,m.\\
\textbf{Dependencies/tasks:} raise the head before lifting out the bowl; turn the selector to choose a setting. Electrical speed selection imposes no mechanical angle coupling to the whisk.\\
\textbf{Functional constraint:} the selector tab must be reachable with the bowl seated.\\
\textbf{Design board:} orthographic, articulated, exploded, and mechanism views.
\end{tracestep}

\begin{tracestep}{agcode}{Coding Agent}{First build}{\tchip{agcode}{construction and checks}}
\tracephase{1--13}{Construction}{uses \texttt{read\_file} and \texttt{lookup\_cadquery\_api}, then \texttt{write\_file}, \texttt{replace}, and \texttt{compile\_model} to build the five parts and their joints. \texttt{probe\_model} and \texttt{request\_render} check dimensions and the head connection.}
\tracephase{14--33}{Motion repair}{examines motion feedback, probes intermediate poses, and revises the head housing and bearing geometry. Further compilation and rendering precede handoff.}
\traceimages{%
  \begin{minipage}[t]{0.28\linewidth}\centering
  \includegraphics[width=\linewidth,height=1.75cm,keepaspectratio,trim={285bp 220bp 290bp 235bp},clip]{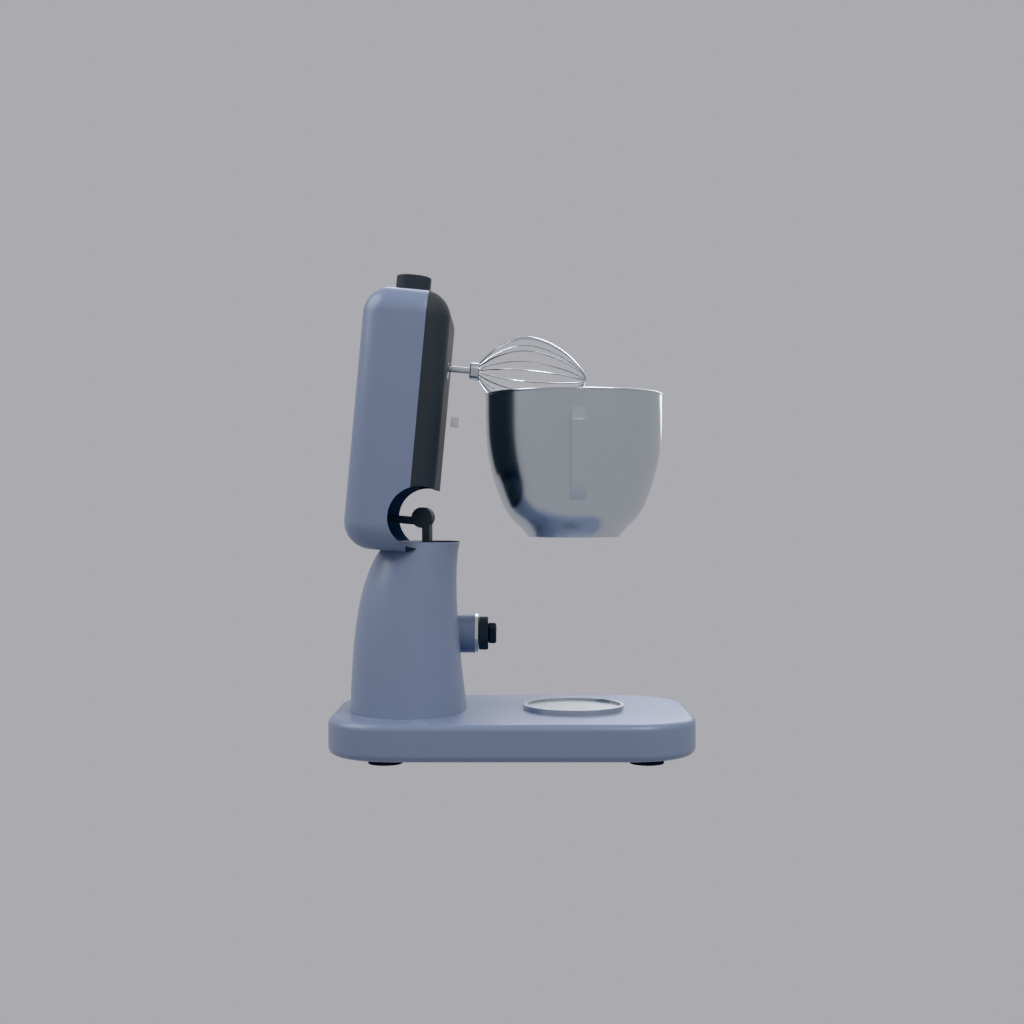}\par
  {\fontsize{6.5}{7.5}\selectfont Early build: head raised, bowl removed\par}
  \end{minipage}\hspace{8pt}%
  \begin{minipage}[t]{0.28\linewidth}\centering
  \includegraphics[width=\linewidth,height=1.75cm,keepaspectratio,trim={285bp 220bp 290bp 235bp},clip]{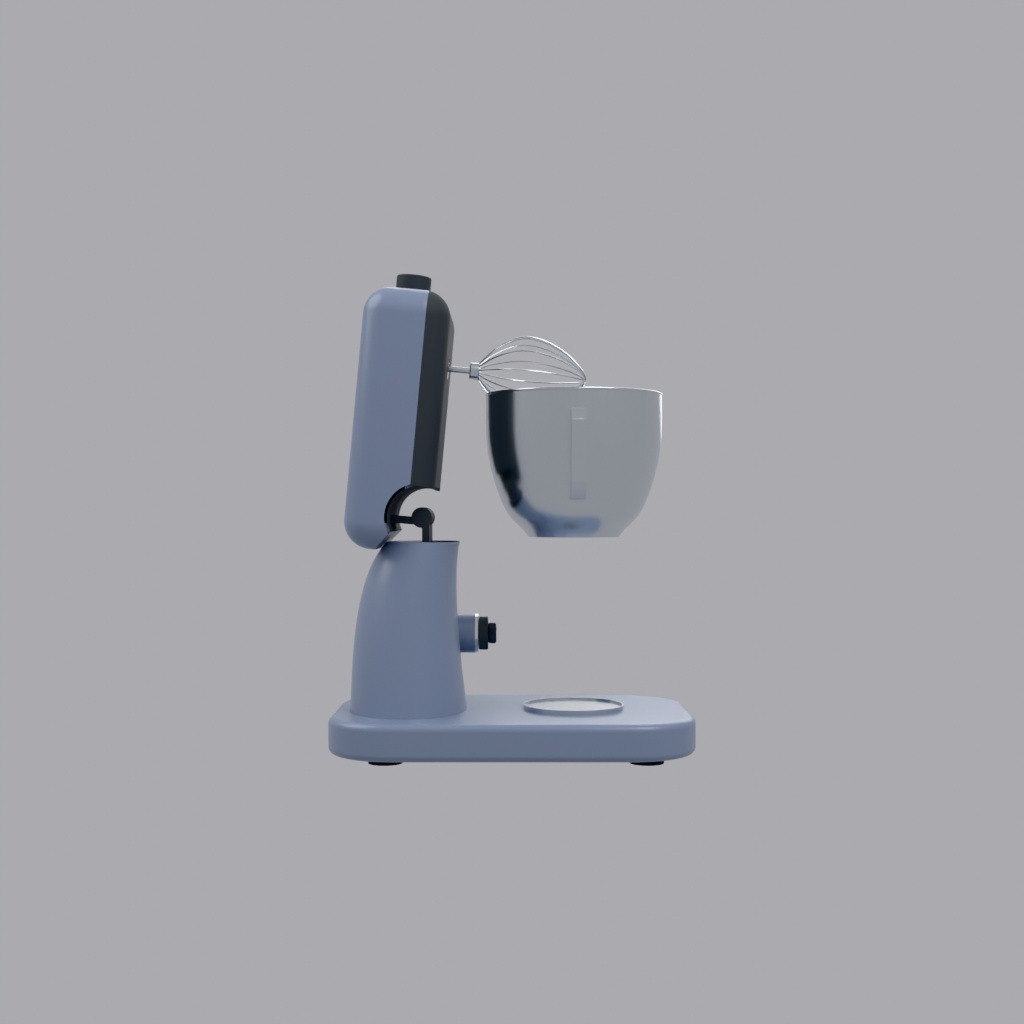}\par
  {\fontsize{6.5}{7.5}\selectfont After motion-clearance revisions\par}
  \end{minipage}\hspace{8pt}%
  \begin{minipage}[t]{0.25\linewidth}\centering
  \includegraphics[width=\linewidth,height=1.75cm,keepaspectratio,]{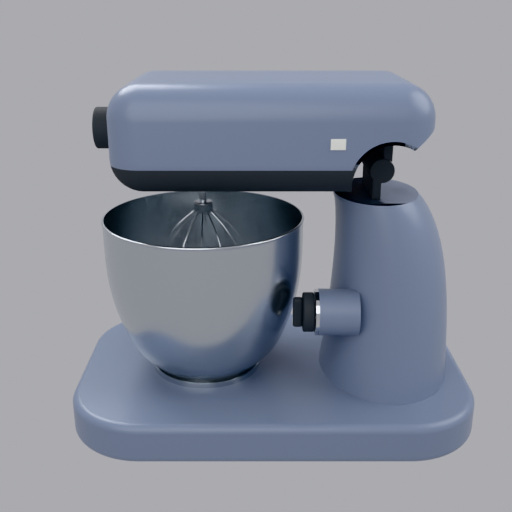}\par
  {\fontsize{6.5}{7.5}\selectfont Candidate sent for review\par}
  \end{minipage}}
\end{tracestep}

\begin{tracestep}{agsim}{Physical validation}{First candidate}{\tchip{aggood}{checks passed}}
\scriptsize
MuJoCo reports passing collision and gravity-stability checks for the submitted candidate. The report supplies the tested poses and measurements to the Review Agent; it does not assess finger access to the selector.
\end{tracestep}

\begin{tracestep}{agreview}{Review Agent}{First review}{\tchip{agbad}{functional repair requested}}
\tracephase{1--3}{Evidence and tests}{reads the program, task bindings, and physics report; uses \texttt{write\_test} and \texttt{run\_tests} to check articulation, clearances, and bowl removal. All 24 checks pass.}
\tracephase{4--9}{Focused inspection}{uses \texttt{crop\_region}, \texttt{request\_render}, and \texttt{probe\_model} to inspect the neck and selector. The tab is trapped behind the seated bowl, with only $4.499$\,mm clearance in the rest pose.}
\tracephase{10}{Feedback}{requests a selector location that satisfies the access requirement, together with a more enclosed rear hinge housing.}
\scriptsize
\traceexcerpt{Revise the stationary socket/bezel location and matching selector joint together so the control is beside the bowl rather than trapped behind its wall. [...] do not solve access by requiring bowl removal.}
\traceimages{%
  \begin{minipage}[t]{0.25\linewidth}\centering
  \includegraphics[width=\linewidth,height=1.75cm,keepaspectratio,]{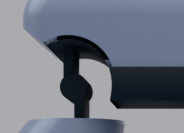}\par
  {\fontsize{6.5}{7.5}\selectfont Original review crop: exposed hinge\par}
  \end{minipage}\hspace{8pt}%
  \begin{minipage}[t]{0.25\linewidth}\centering
  \includegraphics[width=\linewidth,height=1.75cm,keepaspectratio,]{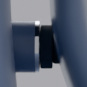}\par
  {\fontsize{6.5}{7.5}\selectfont Original review crop: selector\par}
  \end{minipage}\hspace{8pt}%
  \begin{minipage}[t]{0.25\linewidth}\centering
  \includegraphics[width=\linewidth,height=1.75cm,keepaspectratio,]{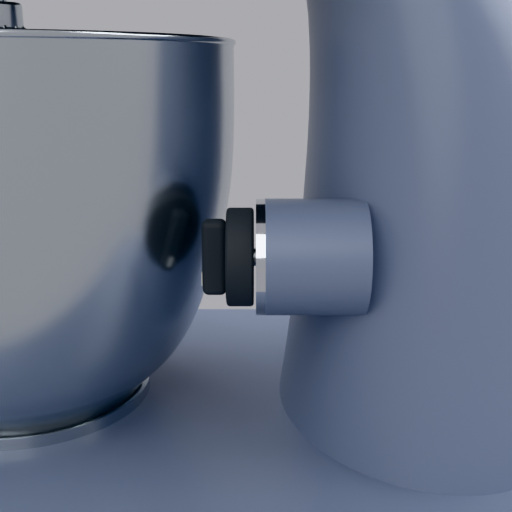}\par
  {\fontsize{6.5}{7.5}\selectfont Matched detail: $4.499$\,mm gap\par}
  \end{minipage}}
\end{tracestep}

\begin{tracestep}{agcode}{Coding Agent}{Second build}{\tchip{agcode}{control and housing revised}}
\tracephase{34--36}{Revision}{reads the review, probes the affected parts, renders the hinge, and uses \texttt{replace} to move the selector/socket assembly to an outer pedestal shoulder and add a hinge cover.}
\tracephase{37--41}{Verification}{probes the motion and selector clearance, recompiles, and inspects full and focused renders before resubmission.}
\scriptsize
\textbf{Preserved:} bowl size, all part masses, and all joint ranges. The selector's placement is changed to meet the blueprint's higher-level access requirement.
\traceimages{%
  \begin{minipage}[t]{0.28\linewidth}\centering
  \includegraphics[width=\linewidth,height=1.75cm,keepaspectratio,trim={285bp 220bp 290bp 235bp},clip]{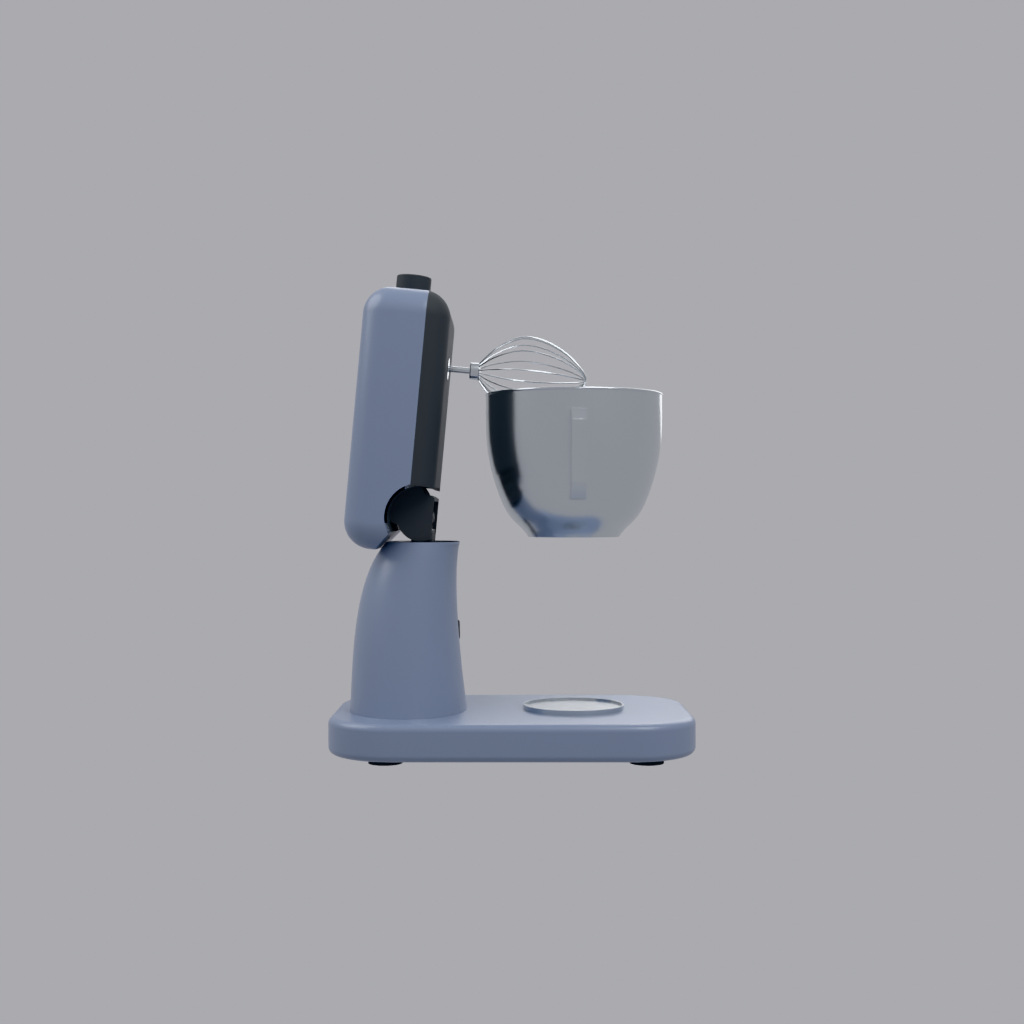}\par
  {\fontsize{6.5}{7.5}\selectfont Revised raised-head configuration\par}
  \end{minipage}\hspace{8pt}%
  \begin{minipage}[t]{0.28\linewidth}\centering
  \includegraphics[width=\linewidth,height=1.75cm,keepaspectratio,]{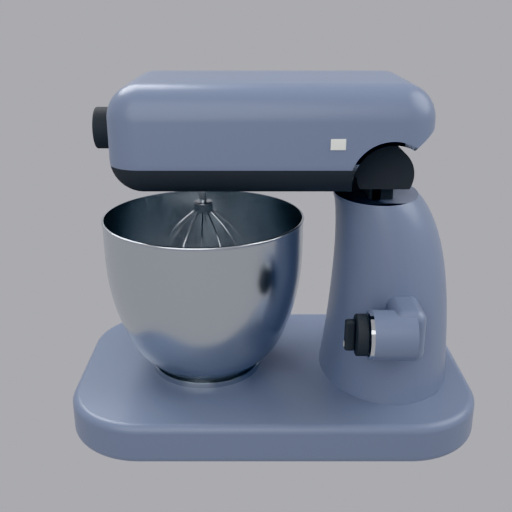}\par
  {\fontsize{6.5}{7.5}\selectfont Selector relocated to the side\par}
  \end{minipage}}
\end{tracestep}

\begin{tracestep}{agsim}{Physical validation}{Revised candidate}{\tchip{aggood}{checks passed}}
\scriptsize
The revised candidate continues to pass collision and stability checks. The updated physics report is supplied for final review.
\end{tracestep}

\begin{tracestep}{agreview}{Review Agent}{Final review}{\tchip{aggood}{repair verified}}
\tracephase{11--12}{Independent verification}{reads the revised program and physics evidence, crops the selector and hinge, and probes 17 selector positions. The existing test suite remains $24/24$.}
\tracephase{13}{Decision}{approves the revision with an overall score of 8 and no listed issues, compared with 6 in the first review.}
\scriptsize
\textbf{Minimum tab-to-bowl distance over 17 poses:} $72.378$\,mm. This geometric measurement and the renders support the access assessment; no hand is simulated.
\traceimages{%
  \begin{minipage}[t]{0.25\linewidth}\centering
  \includegraphics[width=\linewidth,height=1.75cm,keepaspectratio,]{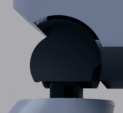}\par
  {\fontsize{6.5}{7.5}\selectfont Original review crop: covered hinge\par}
  \end{minipage}\hspace{8pt}%
  \begin{minipage}[t]{0.25\linewidth}\centering
  \includegraphics[width=\linewidth,height=1.75cm,keepaspectratio,]{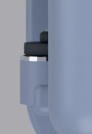}\par
  {\fontsize{6.5}{7.5}\selectfont Original review crop: side approach\par}
  \end{minipage}\hspace{8pt}%
  \begin{minipage}[t]{0.25\linewidth}\centering
  \includegraphics[width=\linewidth,height=1.75cm,keepaspectratio,]{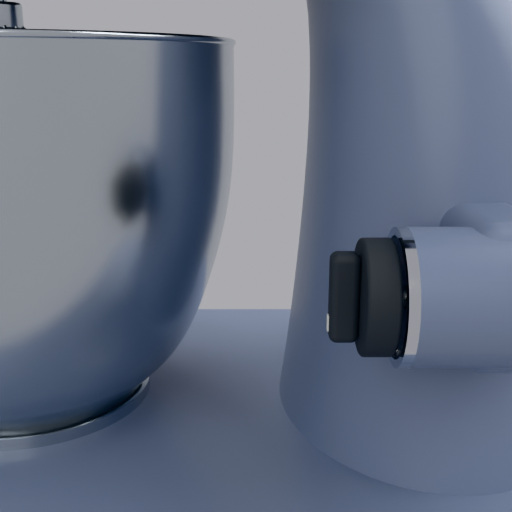}\par
  {\fontsize{6.5}{7.5}\selectfont Matched detail after repair\par}
  \end{minipage}}
\end{tracestep}

\par\vspace{4pt}
\captionof{figure}{\textbf{Stand-mixer construction and review trace.}
Inspection identifies restricted control access despite passing physical checks; the Coding Agent relocates the selector while preserving the bowl and joint ranges.
Small panels include original intermediate renders/crops and matched detail views of the archived candidates.}
\label{fig:agent_trace_mixer}
\endgroup

\subsection{Shoe Cabinet: Drawer Connections}
\label{sec:app_trace_shoe}
\begingroup
\setlength{\parindent}{0pt}
\traceprompt{Reconstruct this articulated object. A slim grey shoe cabinet with two small drawers across the top and two tall shoe compartments below, all fitted with round knobs. The drawers pull straight out, while each shoe compartment tilts forward from its lower edge like a hopper. [...]}

\begin{tracestep}{agdesign}{Design Agent}{Blueprint and design board}{\tchip{agdesign!75}{design specified}}
\scriptsize
\textbf{Intake Turn 1.} Describes the cabinet and its four independent storage motions.\\
\textbf{Blueprint Turns 1--2.} Writes the specification (1), then confirms completion (2). A design board is generated from its geometry and mechanism descriptions.
\par\smallskip
\textbf{Selected outputs (abbreviated).}\\
\textbf{Parts:} one structural carcass, two drawers, two tilt-out shoe compartments.\\
\textbf{Geometry/materials:} an open-front $0.82\times0.31\times1.40$\,m frame in light warm-grey painted wood; hollow drawer trays with $10$--$12$\,mm walls; shoe compartments with curved black ABS liners; [...].\\
\textbf{Structure/joints:} each drawer connects to the carcass by a $+Y$ prismatic joint with $0.20$\,m travel; each compartment has a lower-edge revolute joint about $-X$ with $0.96$\,rad travel. No mechanical couplings are specified.\\
\textbf{Tasks:} pull either drawer knob or tilt either compartment using its front knob.\\
\textbf{Functional constraints:} runners remain engaged at full drawer extension; all four storage members can be open together without interpenetration.\\
\textbf{Design board:} orthographic, articulated, exploded, and connection views.
\end{tracestep}

\begin{tracestep}{agcode}{Coding Agent}{First build}{\tchip{agcode}{construction and checks}}
\tracephase{1--10}{Construction}{uses \texttt{read\_file}, \texttt{find\_examples}, and \texttt{lookup\_cadquery\_api} to inspect slide/hinge interfaces; writes the asset and probes its dimensions before compilation.}
\tracephase{11--23}{Geometry and appearance}{renders open and closed states, probes connectivity and overlaps, revises the hinges and facade, and writes the task bindings.}
\tracephase{24--31}{Repair and handoff}{responds to disconnected drawer geometry using \texttt{replace}, \texttt{probe\_model}, and \texttt{compile\_model}; checks a new front render and submits the candidate.}
\traceimages{%
  \begin{minipage}[t]{0.27\linewidth}\centering
  \includegraphics[width=\linewidth,height=1.75cm,keepaspectratio,trim={325bp 240bp 340bp 240bp},clip]{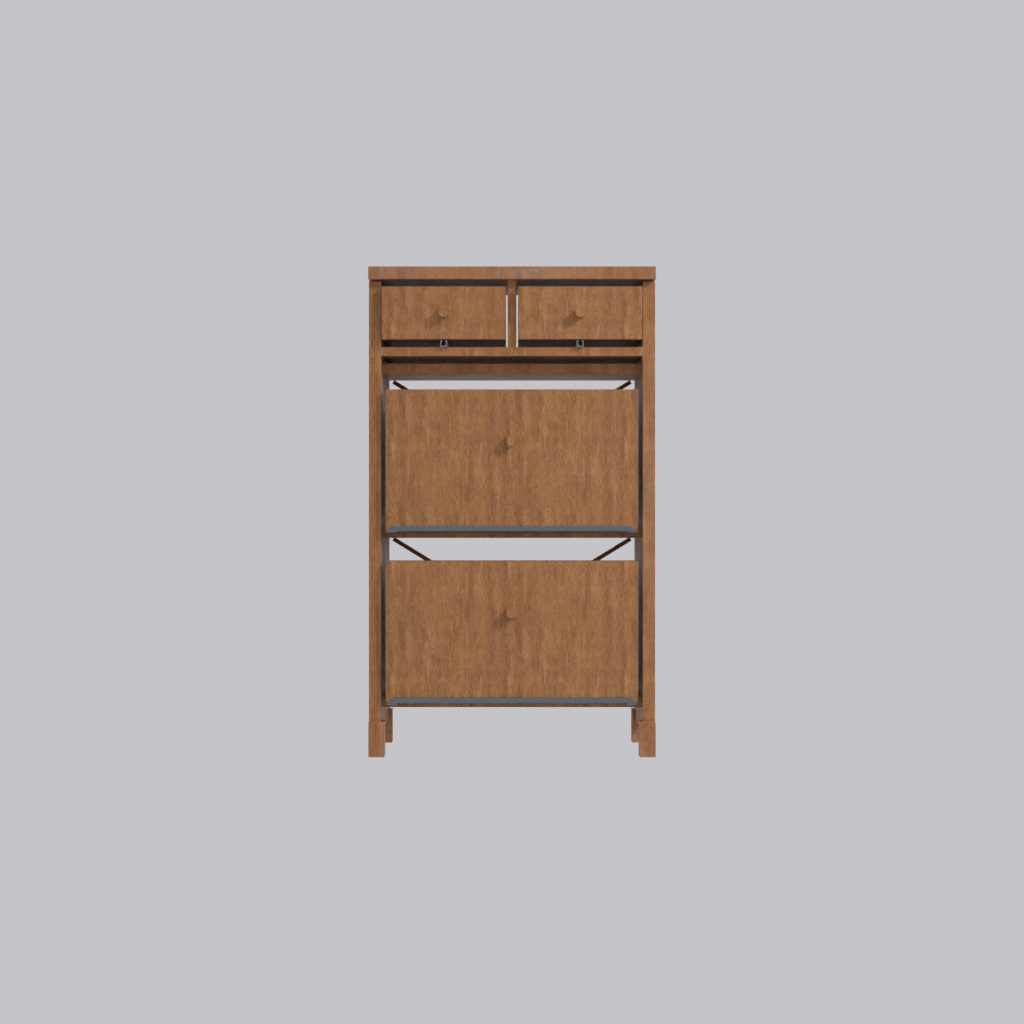}\par
  {\fontsize{6.5}{7.5}\selectfont Intermediate front render\par}
  \end{minipage}\hspace{8pt}%
  \begin{minipage}[t]{0.27\linewidth}\centering
  \includegraphics[width=\linewidth,height=1.75cm,keepaspectratio,trim={285bp 220bp 310bp 190bp},clip]{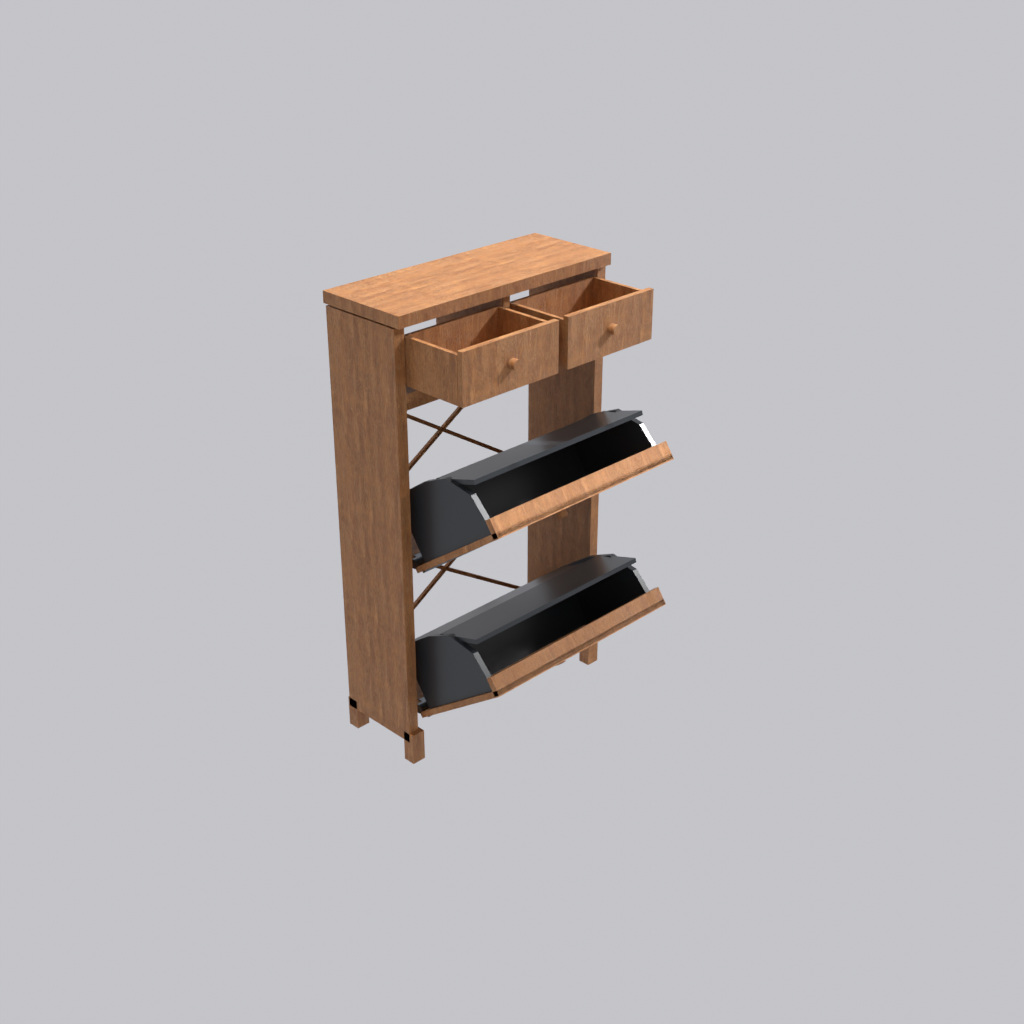}\par
  {\fontsize{6.5}{7.5}\selectfont Intermediate open-state render\par}
  \end{minipage}}
\end{tracestep}

\begin{tracestep}{agsim}{Physical validation}{First candidate}{\tchip{aggood}{checks passed}}
\scriptsize
The submitted candidate passes MuJoCo collision and gravity-stability checks. This does not establish that the drawer runners remain mechanically engaged at the ends of their travel.
\end{tracestep}

\begin{tracestep}{agreview}{Review Agent}{First review}{\tchip{agbad}{connection repair requested}}
\tracephase{1--4}{Independent tests}{reads the blueprint, asset program, and task bindings; uses \texttt{probe\_model}, \texttt{write\_test}, and \texttt{run\_tests} to check scale, motion, retention, and clearances. Only $7/9$ tests pass.}
\tracephase{5--7}{Focused inspection}{crops the facade, drawers, and compartment interiors, then probes their gaps and connections. Both drawer runners lie $15$\,mm beyond their guide channels at full extension; facade gaps are $21$--$65$\,mm.}
\tracephase{8}{Feedback}{requests retained runner engagement, tighter facade gaps, smoother liners, and the specified grey finish.}
\traceimages{%
  \begin{minipage}[t]{0.27\linewidth}\centering
  \includegraphics[width=\linewidth,height=1.75cm,keepaspectratio,]{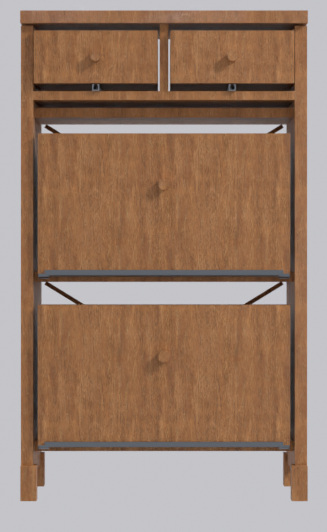}\par
  {\fontsize{6.5}{7.5}\selectfont Original review crop: loose facade\par}
  \end{minipage}\hspace{8pt}%
  \begin{minipage}[t]{0.31\linewidth}\centering
  \includegraphics[width=\linewidth,height=1.75cm,keepaspectratio,]{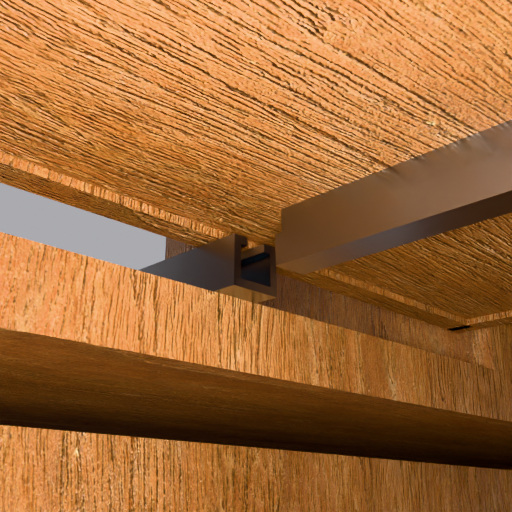}\par
  {\fontsize{6.5}{7.5}\selectfont Under-drawer detail: detached runner\par}
  \end{minipage}}
\end{tracestep}

\begin{tracestep}{agcode}{Coding Agent}{Second build}{\tchip{agcode}{runners and facade revised}}
\tracephase{32--34}{Revision}{queries curve-construction APIs, probes runner/channel bounds, and replaces the runner and liner geometry. Runner length increases from $80$ to $140$\,mm; the exterior finish changes to light grey.}
\tracephase{35--40}{Checks}{probes the revised structure, adjusts facade rails, recompiles, and renders the result. All part masses and joint ranges are preserved.}
\scriptsize
\textbf{Selected change:} \texttt{runner\_length=0.080} $\rightarrow$ \texttt{runner\_length=0.140}.
\traceimages{%
  \begin{minipage}[t]{0.28\linewidth}\centering
  \includegraphics[width=\linewidth,height=1.75cm,keepaspectratio,trim={290bp 225bp 315bp 195bp},clip]{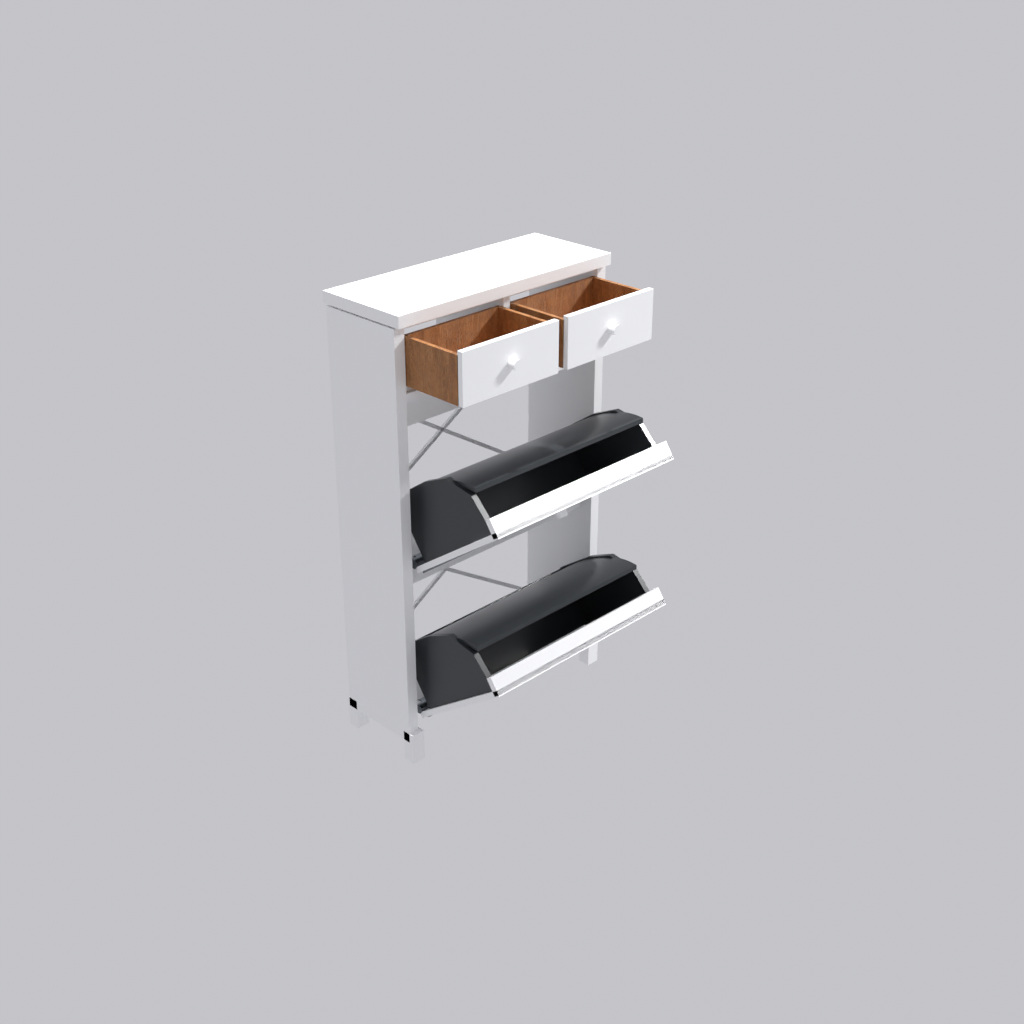}\par
  {\fontsize{6.5}{7.5}\selectfont Revised runner length and finish\par}
  \end{minipage}\hspace{8pt}%
  \begin{minipage}[t]{0.28\linewidth}\centering
  \includegraphics[width=\linewidth,height=1.75cm,keepaspectratio,trim={315bp 215bp 325bp 215bp},clip]{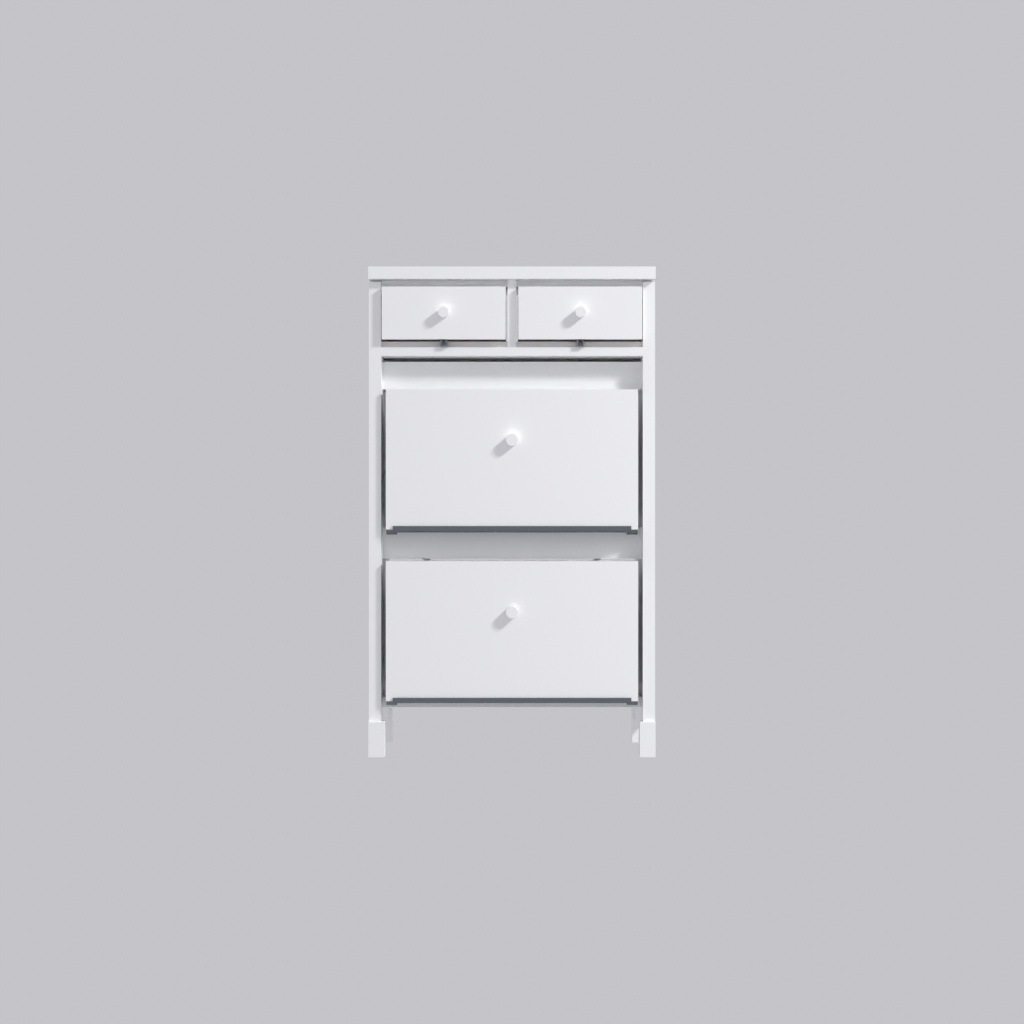}\par
  {\fontsize{6.5}{7.5}\selectfont Tighter facade and grey finish\par}
  \end{minipage}}
\end{tracestep}

\begin{tracestep}{agsim}{Physical validation}{Repair feedback}{\tchip{agwarn}{new interference detected}}
\scriptsize
The motion check reports interference between the revised drawer assembly and the carcass, including approximately $26$\,mm penetration along the slide direction. The finding is returned to the Coding Agent for another correction before final review.
\end{tracestep}

\begin{tracestep}{agcode}{Coding Agent}{Follow-up correction}{\tchip{agcode}{runner paths cleared}}
\tracephase{41--46}{Repair and revalidation}{uses \texttt{replace} to split the fixed front rail around the two runner paths, verifies the gaps with \texttt{probe\_model}, recompiles, and checks the front render. The final candidate passes collision and stability checks.}
\scriptsize
The drawer travel remains $0.20$\,m; the repair changes surrounding geometry rather than shortening the motion.
\end{tracestep}

\begin{tracestep}{agreview}{Review Agent}{Final review}{\tchip{aggood}{engagement verified}}
\tracephase{9--11}{Verification}{reads the revised program and tests, independently probes both runners at full extension, and crops the updated facade and interiors. The test suite now passes $9/9$.}
\tracephase{12}{Decision}{approves the revision with an overall score of 8, up from 5, while noting a minor drawer-interior finish discrepancy.}
\scriptsize
\textbf{Full-extension engagement:} $15$\,mm longitudinally, with $0.7$\,mm surface clearance and no collision. All joint ranges remain unchanged.
\traceimages{%
  \begin{minipage}[t]{0.27\linewidth}\centering
  \includegraphics[width=\linewidth,height=1.75cm,keepaspectratio,]{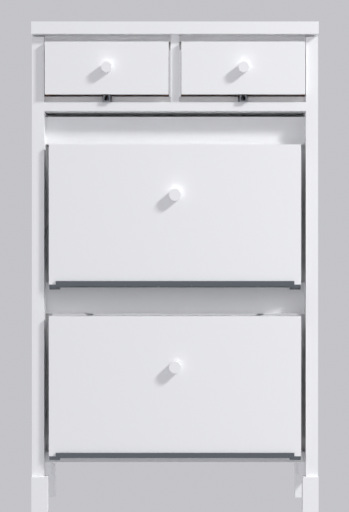}\par
  {\fontsize{6.5}{7.5}\selectfont Original review crop: revised facade\par}
  \end{minipage}\hspace{8pt}%
  \begin{minipage}[t]{0.31\linewidth}\centering
  \includegraphics[width=\linewidth,height=1.75cm,keepaspectratio,]{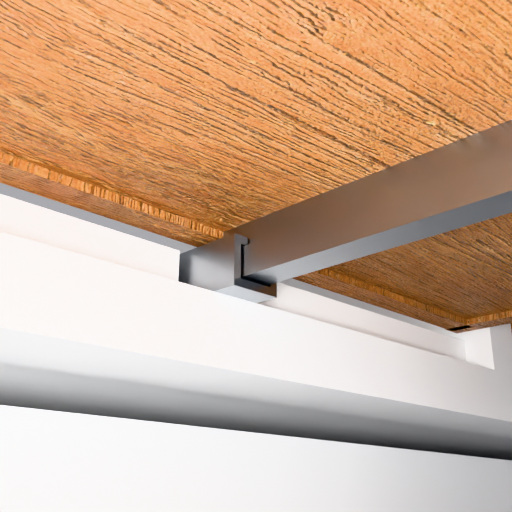}\par
  {\fontsize{6.5}{7.5}\selectfont Under-drawer detail: retained engagement\par}
  \end{minipage}}
\end{tracestep}

\par\vspace{4pt}
\captionof{figure}{\textbf{Shoe-cabinet construction and review trace.}
Longer runners and revised front rails restore engagement without shortening travel.
Under-drawer views use matched cameras and supplementary lighting on the unmodified archived models.}
\label{fig:agent_trace_shoe}
\endgroup

\endgroup

\fi
\end{document}